\documentclass{article} 

\usepackage{amsmath}
\usepackage{amssymb}
\usepackage{amsfonts}
\usepackage{graphicx}
\usepackage{url}
\usepackage{ccaption}
\usepackage{booktabs}  
\usepackage{float}

\usepackage{enumitem}

\usepackage[caption=false]{subfig}

\usepackage[utf8]{inputenc} 
\usepackage[T1]{fontenc}    
\usepackage{hyperref}       
\usepackage{url}            
\usepackage{booktabs}       
\usepackage{amsfonts}       
\usepackage{nicefrac}       
\usepackage{microtype}      
\usepackage[svgnames]{xcolor}
\usepackage{xcolor,colortbl} 
\providecommand{\scal}[2]{\left\langle{#1},{#2}\right\rangle}    

\DeclareMathOperator{\R}{\mathbb{R}}

\providecommand{\scal}[2]{\left\langle{#1},{#2}\right\rangle}
\newcommand{\be}{\begin{equation}}
\newcommand{\ee}{\end{equation}}
\newcommand{\bt}{\begin{theorem}}
\newcommand{\et}{\end{theorem}}
\newcommand{\bd}{\begin{definition}}
\newcommand{\ed}{\end{definition}}
\newcommand{\br}{\begin{remark}}
\newcommand{\er}{\end{remark}}

\newtheorem{theorem}{Theorem}
\newtheorem{definition}{Definition}

\newtheorem{proposition}{Proposition}
\newtheorem{remark}{Remark}

\newtheorem{corollary}{Corollary}

\title{Landscape of the Empirical Risk of Overparametrized Deep Networks} 

\author{ 
  David S.~Hippocampus\thanks{Use footnote for providing further
    information about author (webpage, alternative
    address)---\emph{not} for acknowledging funding agencies.} \\
  Department of Computer Science\\
  Cranberry-Lemon University\\
  Pittsburgh, PA 15213 \\
  \texttt{hippo@cs.cranberry-lemon.edu} \\
}

\usepackage[margin= 0.8in,top=12mm]{geometry}

\newcommand*{\titleAT}{\begingroup
  \newlength{\drop}
  \drop=0.05\textheight
  \begin{center}
  \includegraphics[scale=0.4]{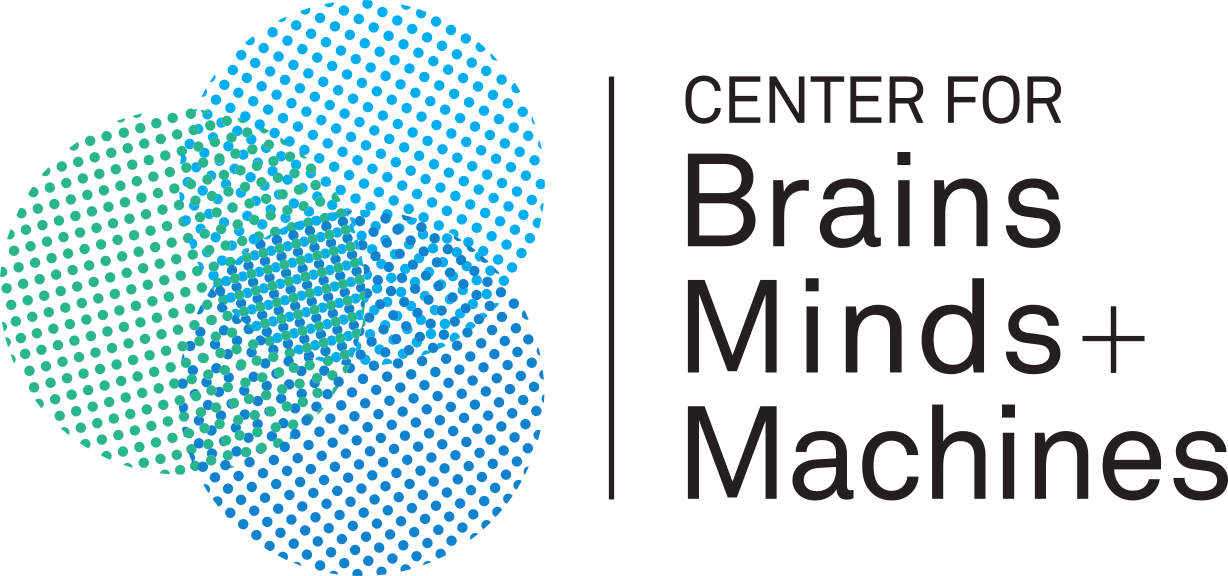} 
  \end{center} 
  \vspace{2pt}\vspace{-\baselineskip}

  \vspace{\drop}
  \textbf{\large{CBMM Memo No. \memonumber}}   \hfill    \textbf{\large{\memodate}} 

  \vspace{\drop}
  \begin{center}
    \textbf{\huge{\memotitle}}\\
    \vspace{0.4\drop}
    \textbf{\Large{by}}\\
    \vspace{0.4\drop}
    \large{\memoauthors}
  \end{center}
  \vspace{\drop} 
  \textbf{\large{\noindent Abstract}:} {\memoabstract}

\vspace{\fill}
  \rule{\textwidth}{0.4pt}\par

  \begin{minipage}{.15\linewidth}
    \includegraphics[scale=0.1]{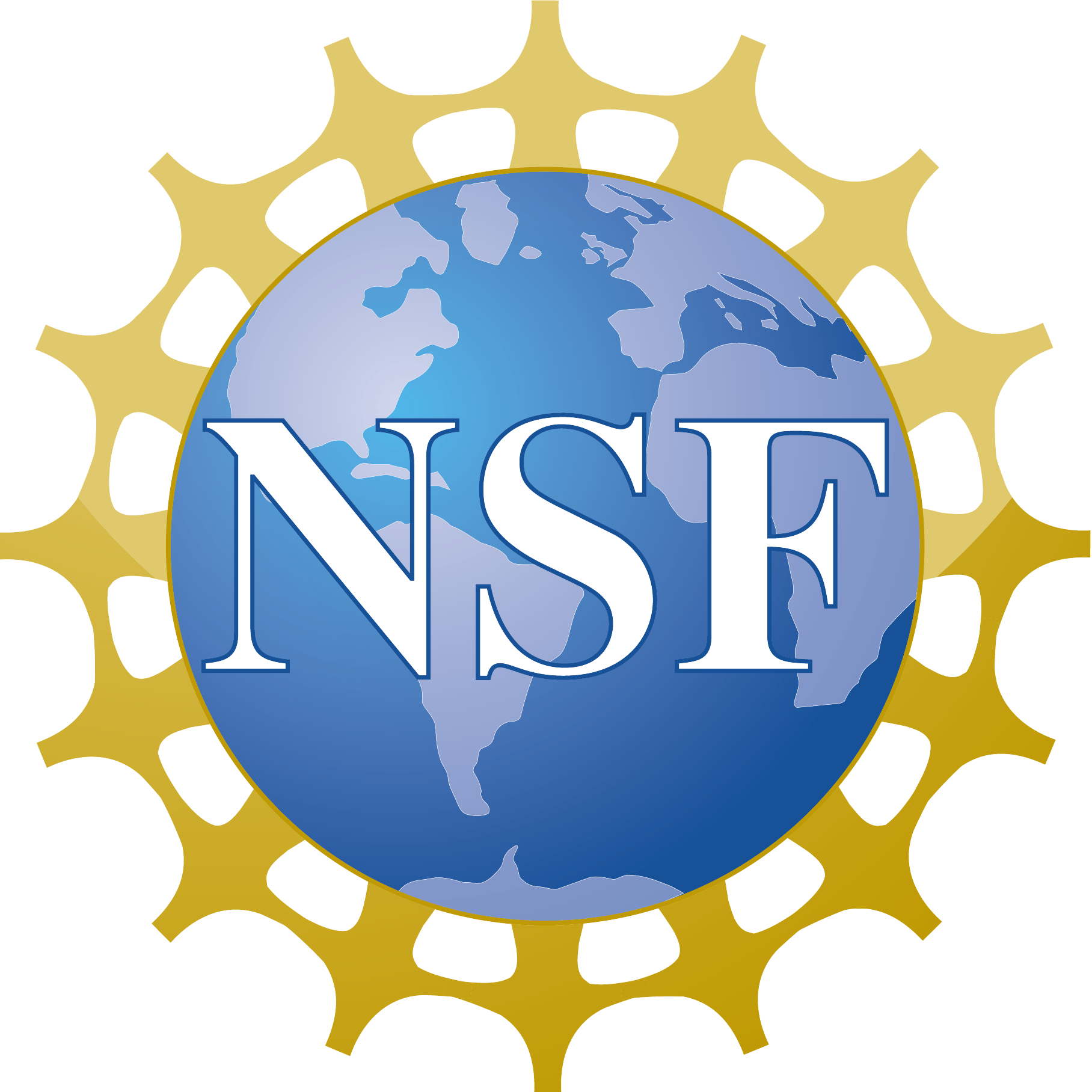}
  \end{minipage}
  \begin{minipage}{.84\linewidth}
    \textbf{\large{This work was supported by the Center for Brains, Minds and Machines (CBMM), funded by NSF STC award  CCF - 1231216.}}
  \end{minipage}
  \endgroup}

\begin{document}

\def\memonumber{066}
\def\memodate{\today}
\def\memotitle {Theory of Deep Learning II: Landscape of the Empirical Risk in Deep Learning}
\def\memoauthors{ Qianli Liao  and Tomaso Poggio  \\ 
Center for Brains, Minds, and Machines, McGovern Institute for Brain Research, \\
Massachusetts Institute of Technology, Cambridge, MA, 02139. 
}

\normalsize
\def\memoabstract{
   
   Previous theoretical work on deep learning and neural network
  optimization tend to focus on avoiding saddle points and local
  minima. However, the practical observation is that, at least in the
  case of the most successful Deep Convolutional Neural Networks
  (DCNNs), practitioners can always increase the network size to fit
  the training data (an extreme example would be
  \cite{zhang2016understanding}). The most successful DCNNs such as
  VGG and ResNets are best used with a degree of
  ``overparametrization''. In this work, we characterize with a mix of
  theory and experiments, the landscape of the empirical risk of
  overparametrized DCNNs. We first prove in the regression framework
  the existence of a large number of degenerate global minimizers with
  zero empirical error (modulo inconsistent equations). The argument
  that relies on the use of Bezout theorem is rigorous when the RELUs
  are replaced by a polynomial nonlinearity (which empirically works
  as well). As described in our Theory III \cite{zhang2017theory}
  paper, the same minimizers are degenerate and thus very likely to be
  found by SGD that will furthermore select with higher probability
  the most robust zero-minimizer. We further experimentally explored
  and visualized the landscape of empirical risk of a DCNN on CIFAR-10
  during the entire training process and especially the global minima.
  Finally, based on our theoretical and experimental results, we
  propose an intuitive model of the landscape of DCNN's empirical loss
  surface, which might not be as complicated as people commonly
  believe.

} 

\titleAT

\newpage

\tableofcontents

\newpage

\section{Introduction}

There are at least three main parts in a theory of Deep Neural Networks.  The
first part is about approximation -- how and when can deep neural networks
avoid the curse of dimensionality' \cite{poggio2016and}?  The second part is about the
landscape of the minima of the empirical risk: what can we say in
general about
global and local minima? The third part is about generalization: why can SGD
(Stochastic Gradient Descent) generalize so well despite standard
overparametrization of the deep neural networks \cite{zhang2017theory}?   
In this paper we focus on the second part: the landscape of the
empirical risk.   
 
\begin{figure}\centering 
  \includegraphics[width=\textwidth]{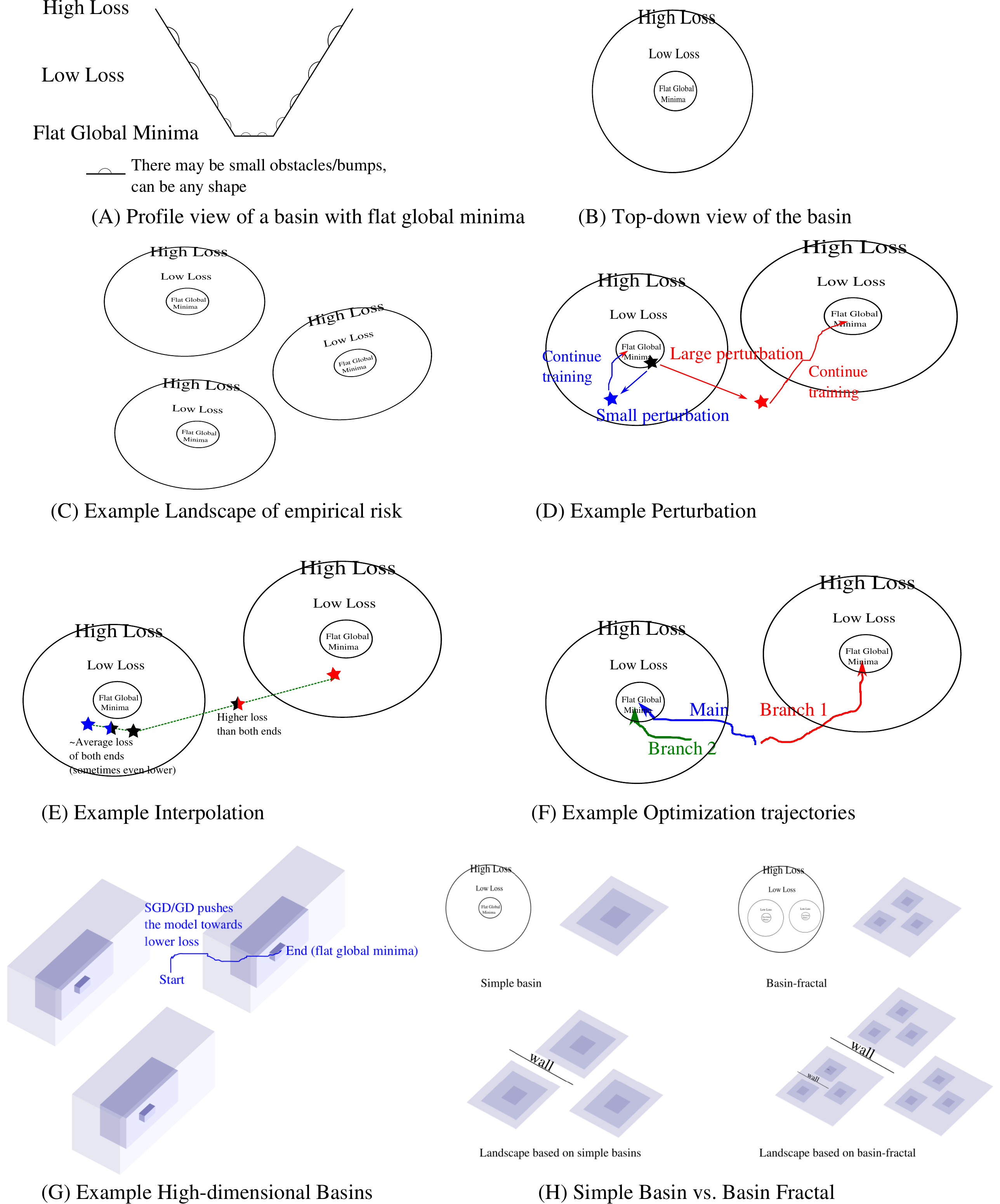}  
\caption{The Landscape of empirical risk of overparametrized DCNN may
  be simply a collection of (perhaps slightly rugged) basins. (A) the
  profile view of a basin (B) the top-down view of a basin (C) example
  landscape of empirical risk (D) example perturbation: a small
  perturbation does not move the model out of its current basin, so
  re-training converges back to the bottom of the same basin. If the
  perturbation is large, re-training converges to another basin. (E)
  Example Interpolation: averaging two models within a basin tend to
  give a error that is the average of the two models (or less). 
  Averaging two models between basins tend to give an error that is
  higher than both models. (F) Example optimization trajectories that 
  correspond to Figure \ref{fig:branch_layer_2_all_perturb_0.25} (G), (H) see Section \ref{sec:intuitive}. }   
\label{fig:landscape_model}
\end{figure}

Our \textbf{main results}: we characterize the \textbf{landscape of the empirical risk} from three perspectives: 

\begin{itemize}[leftmargin=*] 
\item \textbf{Theoretical Analyses (Section \ref{sec:theoretical}):} We
  study the nonlinear system of equations corresponding to critical
  points of the gradient of the loss (for the $L_2$ loss function) and
  to zero minimizers, corresponding to interpolating solutions. In the
  equations the functions representing the network's output contain
  the RELU nonlinearity. We consider an $\epsilon$- approximation of
  it in the sup norm using a polynomial approximation or the
  corresponding Legendre expansion. We can then invoke Bezout theorem
  to conclude that there are a {\it very large number of zero-error
    minima}, and that {\it the zero-error minima are highly
    degenerate}, whereas the local non-zero minima, if they exist, may
  not be degenerate. In the case of classification, zero error implies
  the existence of a margin, that is a flat region in all dimensions around
  zero error.
  \item \textbf{Visualizations and Experimental Explorations (Section \ref{sec:vis}):} The 
    theoretical results indicate that there are degenerate global
    minima in the loss surface of DCNN. However, it is unclear how the
    rest of the landscape look like. To gain more knowledge about
    this, we visualize the landscape of the entire training process
    using \textbf{Multidimensional Scaling}. We also probe locally the
    landscape at different locations by perturbation and interpolation
    experiments.  
  \item \textbf{A simple model of the landscape (Section
    \ref{sec:intuitive}). } Summarizing our theoretical and
    experimental results, we propose a simple baseline model for the
    landscape of empirical risk, as shown in Figure
    \ref{fig:landscape_model}. We conjecture that the loss surface of
    DCNN is not as complicated as commonly believed. At least in the
    case of overparametrized DCNNs, the loss surface might be simply a
    collection of (high-dimensional) basins, which have some of the
    following interesting properties: 1. Every basin reaches a flat
    global minima. 2. The basins may be rugged such that any
    perturbation or noise leads to a slightly different convergence
    path. 3. Despite being perhaps locally rugged, the basin has a
    relatively regular overall landscape such that the average of two
    model within a basin gives a model whose error is roughly the
    average of (or even lower than) the errors of original two models.
    4. Interpolation between basins, on the other hand, may
    significantly raise the error. 5. There might be some good
    properties in each basin such that there is no local minima --- we
    do not encounter any local minima in CIFAR, even when training
    with batch gradient descent without noise.
\end{itemize}

\section{Framework}

  
We assume a deep network of the convolutional type and
 overparametrization, that is more weights than data points,
since this is how successful deep networks have been used. 
Under these conditions, we will show that imposing zero empirical
error provides a system of equations (at the zeros) that have a large
number of degenerate solutions in the weights. The equations are
polynomial in the weights, with coefficients reflecting components of
the data vectors (one vector per data point). The system of equations
is underdetermined (more unknowns than equations, e.g. data points)
because of the assumed overparametrization. Because the global minima
are degenerate, that is flat in many of the dimensions, they are more
likely to be found by SGD than local minima which are less degenerate.
Although generalization is not the focus of this work, such flat minima
are likely to generalize better than sharp minima, which might be the
reason why overparametrized deep networks do not seem to overfit.


\section{Landscape of the Empirical Risk: Theoretical Analyses}
\label{landscape}
\label{sec:theoretical}


The following theoretical analysis of the landscape of the empirical risk is based
on a few assumptions: (1) We assume that the network is overparametrized, typically using several
  times more parameters (the weights of the network) than data points.
  In practice, even with data augmentation (in most of the experiments
  in this paper we do not use data augmentation), one can always make the
  model larger to achieve overparametrization without sacrificing
  either the training or generalization performance. 
(2) This section assumes a regression framework. We study how many
  solutions in weights lead to perfect prediction of training labels.
  In classification settings, such solutions is a subset of all solutions.   


Among the critical points of the gradient of the empirical loss, we
consider first the zeros of loss function given by the set of
equations where $N$ is the number of training examples $f(x_i) - y_i= 0\,\,\,\,for\,\,i=1,\cdots,N$


The function $f$ realized by a deep neural network is polynomial if
each of RELU units is replaced by a univariate polynomial. Of course
each RELU can be approximated within any desired $\epsilon$ in the sup
norm by a polynomial.  In the well-determined case (as many unknown
weights as equations, that is data points), Bezout theorem provides an
upper bound on the number of solutions. {\it The number of distinct
  zeros} (counting points at infinity, using projective space,
assigning an appropriate multiplicity to each intersection point, and
excluding degenerate cases) would be {\it equal to $Z$ - the product
  of the degrees of each of the equations}.  Since the system of
equations is usually underdetermined -- as many equations as data
points but more unknowns (the weights) -- we expect an infinite number
of global minima, under the form of $Z$ {\it regions} of zero
empirical error. If the equations are inconsistent there are still
many global minima of the squared error that are solutions of systems
of equations with a similar form.

We assume for simplicity that the equations have a particular compositional form (see
\cite{Mhaskaretal2016}).  The degree of each approximating equation
$\ell^d(\epsilon)$ is determined by the desired accuracy $\epsilon$
for approximating the ReLU activation by a univariate polynomial $P$
of degree $\ell(\epsilon)$ and by the number of layers $d$.

\begin{figure}\centering 
  \centering  
\makebox[0pt]{  \includegraphics[width=1.2\textwidth]{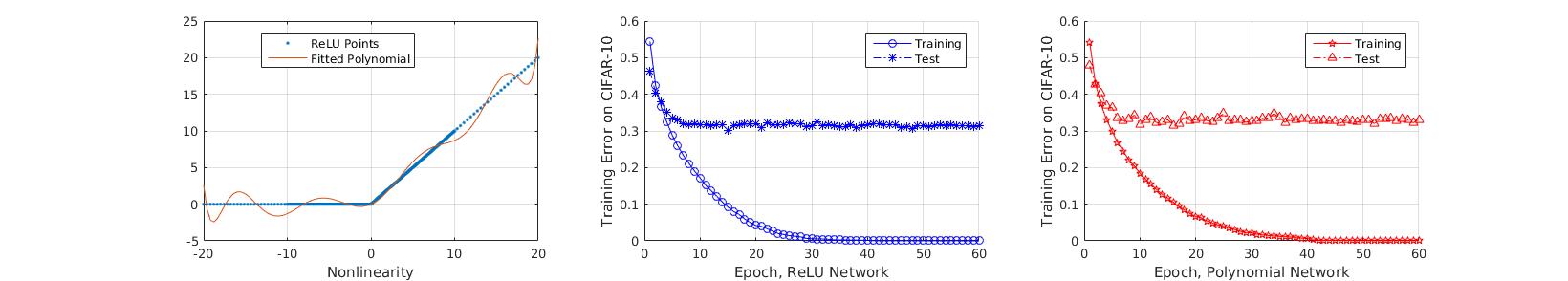}  }
\caption{One can convert a deep network into a polynomial function by
  using polynomial nonlinearity. As long as the nonlinearity
  approximates ReLU well (especially near 0), the ``polynomial net''
  performs similarly to a ReLU net. Our theory applies rigorously to a ``polynomial net''.   }
\label{fig:relu_vs_polynomial}   
\end{figure}

The argument based on RELUs approximation for estimating the number of
zeros is a qualitative one since good approximation of the $f(x_i)$
does not imply by itself good approximation -- via Bezout theorem -- of the number of its
zeros. Notice, however, that we could abandon completely the approximation
approach and just consider the number of zeros when the RELUs are
replaced by a low order univariate polynomial. The argument then would not
strctly apply to RELU networks but would still carry weight because the two
types of networks -- with polynomial activation and with RELUs --
behave empirically (see Figure \ref{fig:relu_vs_polynomial}) in a similar way.  
 
In the Supporting Material we provide a simple example of a network
with associated equations for the exact zeros. They are  a system of underconstrained polynomial
equations of degree $l^d$. In general, there are as many constraints
as data points $i=1,\cdots,N$ for a much larger number $K$ of unknown
weights $W, w, \cdots$.  There are no solutions if the system is
inconsistent -- which happens if and only if $0 = 1$ is a linear
combination (with polynomial coefficients) of the equations (this is
Hilbert's Nullstellensatz).  Otherwise, it has infinitely many complex
solutions: the set of all solutions is an algebraic set of dimension
at least $K-N$. If the underdetermined system is chosen at random the
dimension is equal to $K-N$ with probability one.


Even in the non-degenerate case (as many data as parameters), Bezout
theorem suggests that there are many solutions.  With $d$ layers the
degree of the polynomial equations is $\ell^d$. With $N$ datapoints
the Bezout upper bound in the zeros of the weights is
$\ell^{Nd}$. Even if the number of real zero -- corresponding to zero
empirical error -- is much smaller (Smale and Shub estimate
\cite{ShubSmale94} $l^{\frac{Nd}{2}}$), the number is still enormous:
for a CiFAR situation this may be as high as $2^{10^5}$.

As mentioned, in several cases we expect absolute zeros to exist with
zero empirical error. If the equations are inconsistent it seems
likely that global minima with similar properties exist.


It is interesting now to consider the critical points of the gradient.
The critical points of the gradient are
$\nabla_w\sum_{i=1}^NV(f(x_i), y_i)=0$, \noindent which gives $K$
equations: $\sum_{i=1}^N \nabla_wV(f(x_i), y_i) \nabla_w f(x_i)=0$,
where $V(. , .)$ is the loss function.      




Approximating within $\epsilon$ in the sup norm each ReLU in $f(x_i)$
with a fixed polynomial $P(z)$ yields again a system of $K$ polynomial
equations in the weights of higher order than in the case of
zero-minimizers. They are of course satisfied by the degenerate zeros
of the empirical error but also by additional non-degenerate (in the
general case) solutions.

Thus, we have \textbf{Proposition 1:} \textit{There are a very large number of zero-error minima which are highly degenerate unlike the local  non-zero minima.}  



\section{The Landscape of the Empirical Risk: Visualizing and Analysing the Loss Surface During the Entire Training Process (on CIFAR-10)}
\label{sec:vis}

\subsection{Experimental Settings}
In the empirical work described below we analyze a classification
problem with cross entropy loss. Our theoretical analyses with the
regression framework provide a \textit{lower bound} of the number of
solutions of the classification problem.

Unless mentioned otherwise, we trained a 6-layer (with the 1st layer
being the input) Deep Convolutional Neural Network (DCNN) on
CIFAR-10. All the layers are 3x3 convolutional layers with stride
2. No pooling is performed. Batch Normalizations (BNs) \cite{ioffe2015batch} are used
between hidden layers. The shifting and scaling parameters in BNs are
not used. No data augmentation is performed, so that the training set
is fixed (size = 50,000). There are 188,810 parameters in the DCNN.

\textbf{Multidimensional Scaling} The core of our visualization
approach is Multidimensional Scaling (MDS) \cite{borg2005modern}. We
record a large number of intermediate models during the process of
several training schemes. Each model is a high dimensional point with
the number of dimensions being the number of parameters. The
strain-based MDS algorithm is applied to such points and a
corresponding set of 2D points are found such that the dissimilarity
matrix between the 2D points are as similar to those of the
high-dimensional points as possible. One minus cosine distance is used
as the dissimilarity metric. This is more robust to scaling of the
weights, which is usually normalized out by BNs. Euclidean distance
gives qualitatively similar results though.

\subsection{Global Visualization of SGD Training Trajectories}

\begin{figure*}\centering
  \centering
\makebox[0pt]{\includegraphics[width=1.05\textwidth]{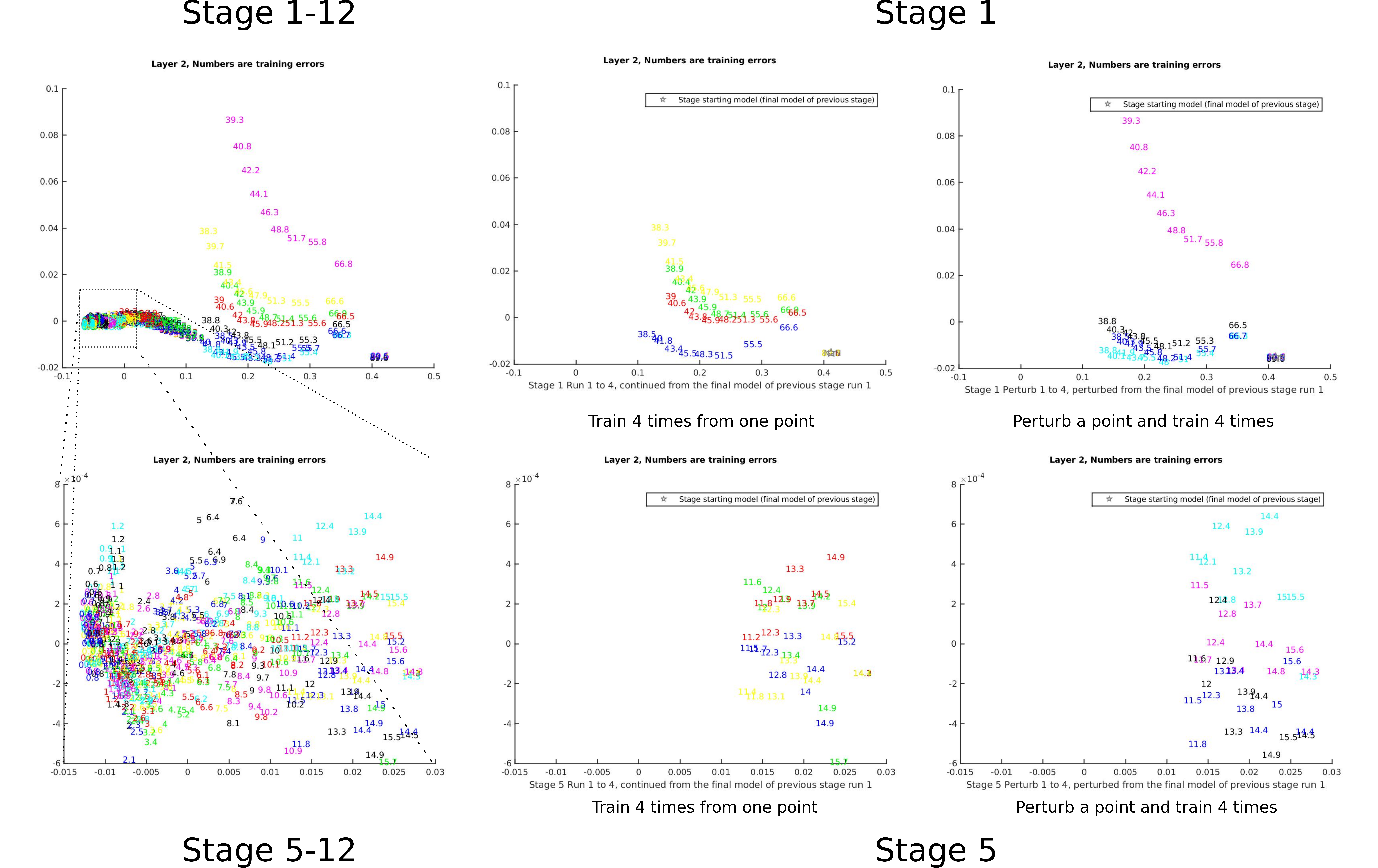}}   
\caption{We train a 6-layer (with the 1st layer being the input) convolutional
network on CIFAR-10 with stochastic gradient descent (batch size =
100). We divide the training process
  into 12 stages. In each stage, we perform \textbf{8 parallel} SGDs
  with learning rate 0.01 for 10 epochs, resulting in 8 parallel
  trajectories denoted by different colors. Trajectories 1 to 4 in
  each stage start from the final model (denoted by $P$) of trajectory
  1 of the previous stage. Trajectories 5 to 8 in each stage start
  from a perturbed version of $P$. The perturbation is performed by
  adding a gaussian noise to the weights of each layer with the
  standard deviation being 0.01 times layer's standard deviation. In
  general, we observe that running any trajectory with SGD again
  almost always leads to a slightly different convergence path. We
  plot the MDS results of all the layer 2 weights collected throughout
  all the training epochs from stage 1 to 12.
  Each number in the figure represents a
  model we collected during the above procedures. The points are in a
  2D space generated by the MDS algorithm such that their pairwise distances
  are optimized to try to reflect those distances in the original high-dimensional space. 
  The results of stages more than 5 are quite cluttered. So we applied a
  separate MDS to the stages 5 to 12. We also plot stage 1 and 5 separately for example.
  The trajectories of more stages are plotted in the Appendix. 
}
\label{fig:global_vis_layer2_stage_0}
\end{figure*}

We show in Figure \ref{fig:global_vis_layer2_stage_0} the optimization
trajectories of Stochastic Gradient Descent (SGD) throughout the
entire optimization process of training a DCNN on CIFAR-10. The SGD
trajectories follow the mini-batch approximations of the training loss
surface. Although the trajectories are noisy due to SGD, the collected
points along the trajectories provide a good visualization of the
landscape of empirical risk. We show the visualization based on the
weights of layer 2. The results from other layers (e.g., layer 5) are
qualitatively similar and are shown in the Appendix.
 
 


\subsection{Global Visualization of Training Loss Surface with Batch Gradient Descent}
 
\begin{figure*}\centering
  \centering 
\makebox[0pt]{\includegraphics[width=1.05\textwidth]{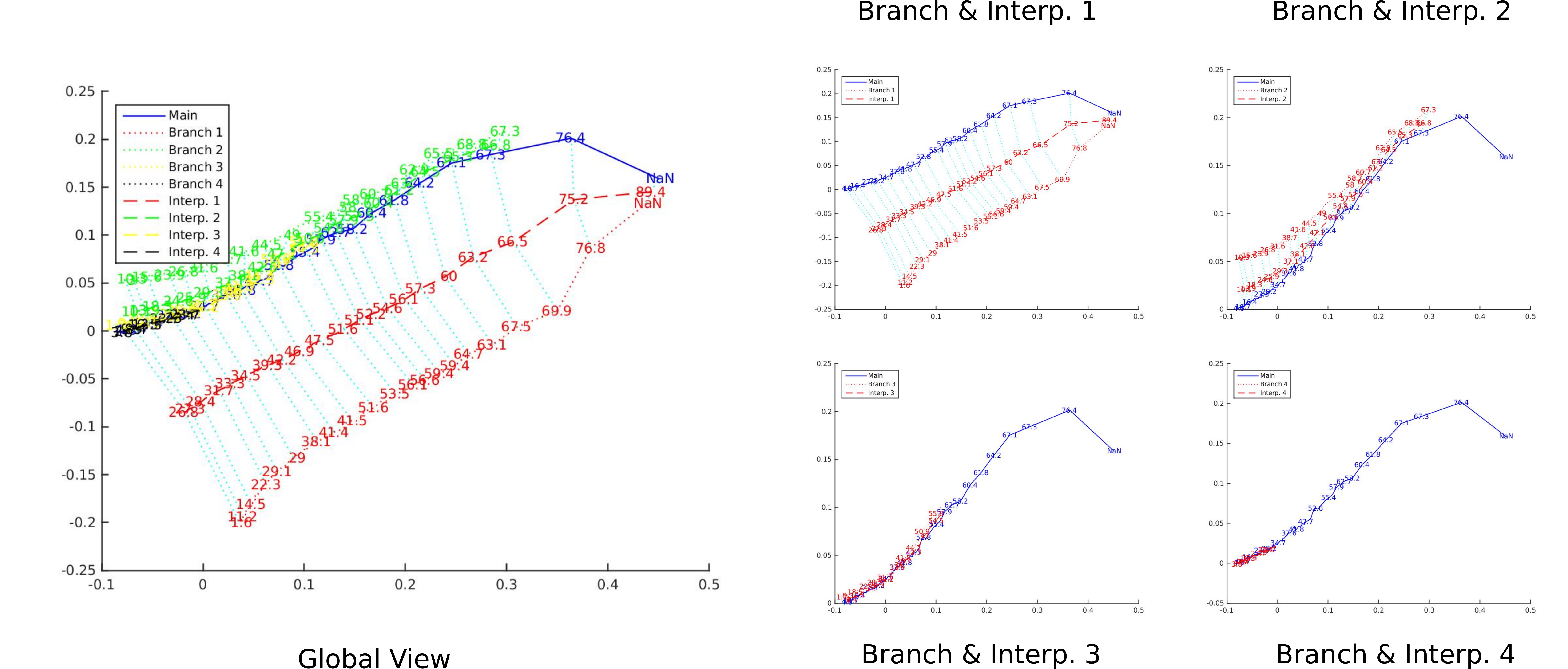}}   
\caption{ Visualizing the exact training loss surface using Batch
  Gradient Descent (BGD). A DCNN is trained on CIFAR-10 from scratch
  using Batch Gradient Descent (BGD). The numbers are training errors.
  ``NaN'' corresponds to randomly initialized models (we did not
  evaluate them and assume they perform at chance). At epoch 0, 10, 50
  and 200, we create a branch by perturbing the model by adding a
  Gaussian noise to all layers. The standard deviation of the Gaussian
  is 0.25*S, where S denotes the standard deviation of the weights in
  each layer, respectively. We also interpolate (by averaging) the
  models between the branches and the main trajectory, epoch by epoch.
  The interpolated models are evaluated on the entire training set to
  get a performance. First, surprisingly, BGD does not get stuck in
  any local minima, indicating some good properties of the landscape.
  The test error of solutions found by BGD is somewhat worse than
  those found by SGD, but not too much worse (BGD ~ 40\%, SGD ~ 32\%)
  . Another interesting observation is that as training proceeds, the
  same amount of perturbation are less able to lead to a drastically
  different trajectory. Nevertheless, a perturbation almost always
  leads to at least a slightly different model. The local neighborhood
  of the main trajectory seems to be relatively flat and contain many
  good solutions, supporting our theoretical predictions. It is also
  intriguing to see interpolated models to have very reasonable
  performance. The results here are based on weights from layer 2. The
  results of other layers are similar and are shown in the Appendix. }  
\label{fig:branch_layer_2_all_perturb_0.25}
\end{figure*}

Next, we visualize in Figure \ref{fig:branch_layer_2_all_perturb_0.25}
the exact training loss surface by training the models using Batch
Gradient Descent (BGD). In addition to training, we also performed
perterbation and interpolation experiments as described in Figure
\ref{fig:branch_layer_2_all_perturb_0.25}. The main trajectory,
branches and the interpolated models together provides a good
visualization of the landscape of the empirical risk. 



\subsection{More Detailed Analyses of Several Local Landscapes (especially the flat global minima)} 
\label{vis:sec3} 

\begin{figure*}[!h]
\centering
\includegraphics[width=\textwidth]{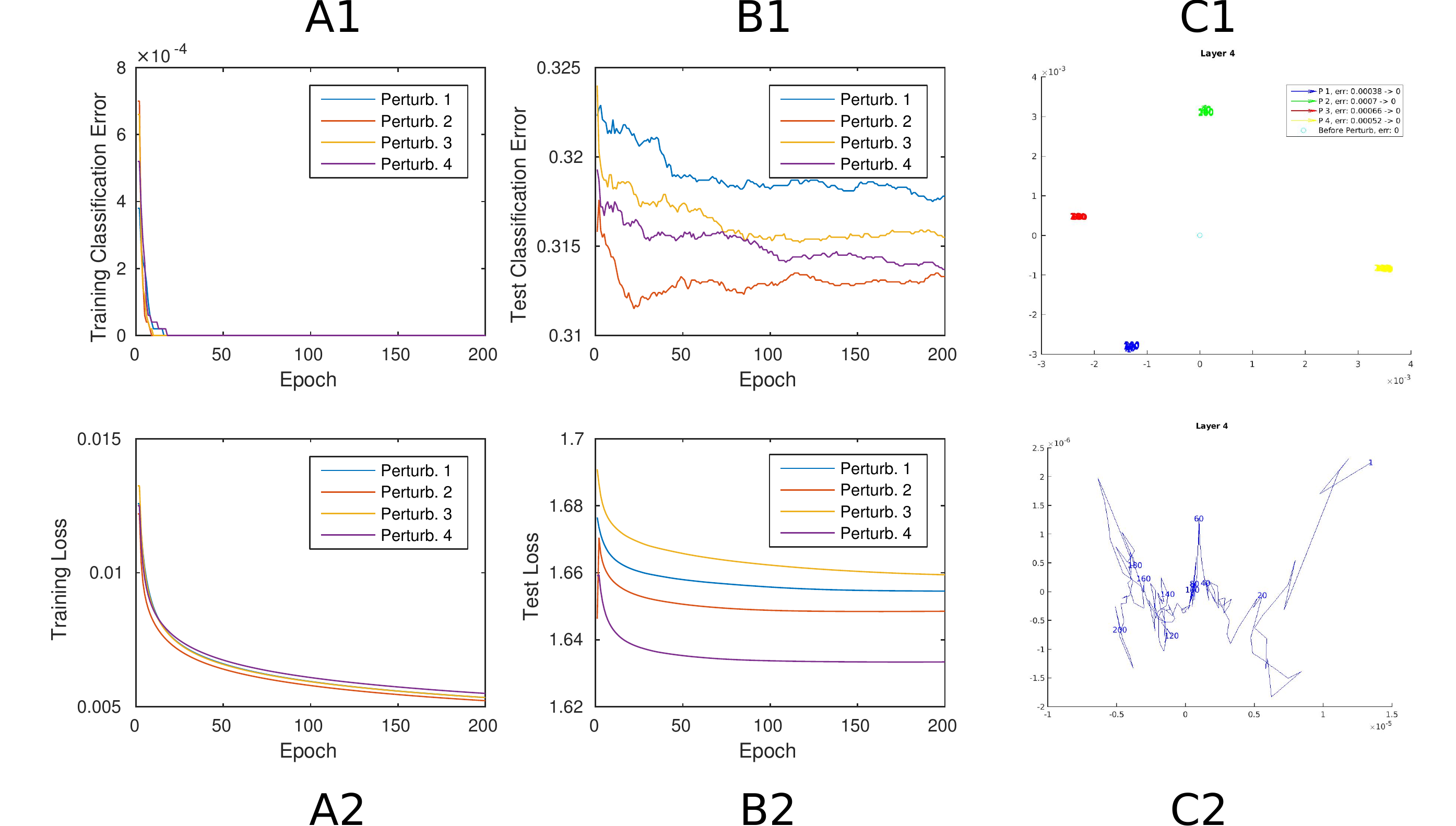}   
\caption{Verifying the flatness of global minima: We layerwise perturb the weights of model $M_{final}$ (which is at a global minimum) by adding a gaussian noise with standard deviation = 0.1 * S, where S is the standard deviation of the weights. After perturbation, we continue training the model with 200 epochs of gradient descent (i.e., batch size = training set size). The same procedure was performed 4 times, resulting in 4 curves shown in the figures. The training and test classification errors and losses are shown in A1, A2, B1 and B2. The MDS visualization of 4 trajectories (denoted by 4 colors) is shown in C1 --- the 4 trajectories converge to different solutions. The MDS visualization of one trajectory is in C2. In addition, we show the confusion matrices of converged models in the Appendix to verify that they are indeed different models. More similar experiments can be found in the Appendix. }    
\label{fig:perturb_err_loss_from_epoch_sgd60_plus_gd400}
\end{figure*}

We perform some more detailed analyses at several locations of the
landscape. Especially, we would like to check if the global minima is
flat. We train a 6-layer (with the 1st layer being the input) DCNN on
CIFAR-10 with 60 epochs of SGD (batch size = 100) and 400 epochs of
Batch Gradient Descent (BGD). BGD is performed to get to as close to
the global minima as possible. Next we select three models from this
learning trajectory (1) $M_5$: the model at SGD epoch 5. (2) $M_{30}$:
the model at SGD epoch 30. (3) $M_{final}$: the final model after 60
epochs of SGD and 400 epochs of BGD. The results of (3) are shown in 
Figure \ref{fig:perturb_err_loss_from_epoch_sgd60_plus_gd400} while
those of (1) and (2) are shown in the the Appendix.

\section{The Landscape of the Empirical Risk: Towards an Intuitive Baseline Model} 
\label{sec:intuitive}

In this section, we propose a simple baseline model for the landscape of empirical risk that is consistent with all of our theoretical and experimental findings.
In the case of overparametrized DCNNs, here is a recapitulation of our main observations: \\
$\bullet$ Theoretically, we show that there are a large number of global minimizers with zero (or small) empirical error. The same minimizers are degenerate. \\
$\bullet$ Regardless of Stochastic Gradient Descent (SGD) or Batch Gradient Descent (BGD), a small perturbation of the model almost always leads to a slightly different convergence path.  The earlier the perturbation is in the training process the more different the final model would be. \\
$\bullet$ Interpolating two ``nearby'' convergence paths lead to another convergence path with similar errors every epoch.  Interpolating two ``distant'' models lead to raised errors. \\
$\bullet$ We do not observe local minima, even when training with BGD. \\   

There is a simple model that is consistent with above observations. As 
a first-order characterization, we believe that the landscape of
empirical risk is simply \textbf{a collection of (hyper) basins that each has
  a flat global minima}. Illustrations are provided in Figure
\ref{fig:landscape_model}. 
  
As shown in Figure \ref{fig:landscape_model}, the building block of the landscape is a 
basin. \textbf{How does a basin look like in high dimension? Is there any evidence for this model?} One definition of a hyper-basin    
would be that as loss decreases, the hypervolume of the parameter
space decreases (see Figure \ref{fig:landscape_model} (H) for example).  As we
can see, with the same amount of scaling in each dimension, the volume 
shrinks much faster as the number of dimension increases --- with a
linear decrease in each dimension, the hypervolume decreases as a
exponential function of the number of dimensions. With the number of
dimensions being the number of parameters, the volume shrinks
incredibly fast. This leads to a phenomenon that we all observe
experimentally: whenever one perturb a model by adding some
significant noise, the loss almost always never go down. The larger
the perturbation is, the more the error increases. The reasons are
simple if the local landscape is a hyper-basin: the volume of a lower
loss area is so small that by randomly perturbing the point, there is
almost no chance getting there. The larger the perturbation is, the
more likely it will get to a much higher loss area.


There are, nevertheless, other plausible variants of this model that
can explain our experimental findings. In Figure
\ref{fig:landscape_model} (G), we show one alternative model we call
``basin-fractal''. This model is more elegant while being also
consistent with most of the above observations. The key difference
between simple basins and ``basin-fractal'' is that in
``basin-fractal'', one should be able to find ``walls'' (raised
errors) between two models within the same basin. Since it is a
fractal, these ``walls'' should be present at all levels of errors.
For the moment, we only discovered ``walls'' between two models the
trajectories lead to which are very different (obtained either by
splitting very early in training, as shown in Figure
\ref{fig:branch_layer_2_all_perturb_0.25} branch 1 or by a very significant
perturbation, as shown in the Appendix). We have not found other  
significant ``walls'' in all other perturbation and interpolation
experiments. So a first order model of the landscape would be just a
collection of simple basins. Nevertheless, we do find
``basin-fractal'' elegant, and perhaps the ``walls'' in the low loss
areas are just too flat to be noticed.
 
Another surprising finding about the basins is that, they seem to be
so ``smooth'' such that there is no local minima. Even when training
with batch gradient descent, we do not encounter any local minima.
When trained long enough with small enough learning rates, one always
gets to 0 classification error and negligible cross entropy loss.

\section{Previous theoretical work}

Deep Learning references start with Hinton's backpropagation and with
LeCun's convolutional networks (see for a nice review
\cite{LeCunBengioHinton2015}). Of course, multilayer convolutional
networks have been around at least as far back as the optical
processing era of the 70s. The Neocognitron\cite{fukushima:1980} was a
convolutional neural network that was trained to recognize characters.
The property of {\it compositionality} was a main motivation for
hierarchical models of visual cortex such as HMAX which can be
regarded as a pyramid of AND and OR layers\cite{Riesenhuber1999}, that
is a sequence of conjunctions and disjunctions.   \cite{poggio2016and} provided formal conditions under which deep
networks can avoid the curse of dimensionality. More specifically,
several papers have appeared on the landscape of the training error
for deep networks.  Techniques borrowed from the physics of spin
glasses (which in turn were based on old work by Marc Kac on the zeros
of algebraic equations) were used \cite{landscape2015} to suggest the
existence of a band of local minima of high quality as measured by the
test error. The argument however depends on a number of assumptions
which are rather implausible (see \cite{SoudryCarmon2016} and
\cite{kawaguchi2016deep} for comments and further work on the
problem). Soudry and Carmon \cite{SoudryCarmon2016} show that with
mild over-parameterization and dropout-like noise, training error for
a neural network with one hidden layer and piece-wise linear
activation is zero at every local minimum. All these results suggest
that the energy landscape of deep neural networks should be easy to
optimize. They more or less hold in practice —it is easy to
optimize a prototypical deep network to near-zero loss on the training
set.
 

\section{Discussion and Conclusions}
\textbf{Are the results shown in this work data dependent?} We visualized the SGD trajectories in the case of fitting random labels. There is no qualitative difference between the results from those of normal labels. So it is safe to say the results are at least not label dependent. We will further check if fitting random input data to random labels will give similar results. \\
\textbf{What about Generalization?} It is experimentally observed that (see Figure \ref{appendix:fig:generalization} in the Appendix), at least in all our experiments, overparametrization (e.g., 60x more parameters than data) does not hurt generalization at all. \\  
\textbf{Conclusions:} Overall, we characterize the landscape of empirical risk of overparametrized DCNNs with a mix of theoretical analyses and experimental explorations. We provide a simple baseline model of the landscape that can account for all of our theoretical and experimental results. Nevertheless, as the final model is so simple, it is hard to believe that it would completely characterize the true loss surface of DCNN. Further research is warranted.     

\bibliographystyle{ieeetr}  
\bibliography{Boolean}

\newpage

\appendix

 

\begin{figure}\centering 
\includegraphics[width=\textwidth]{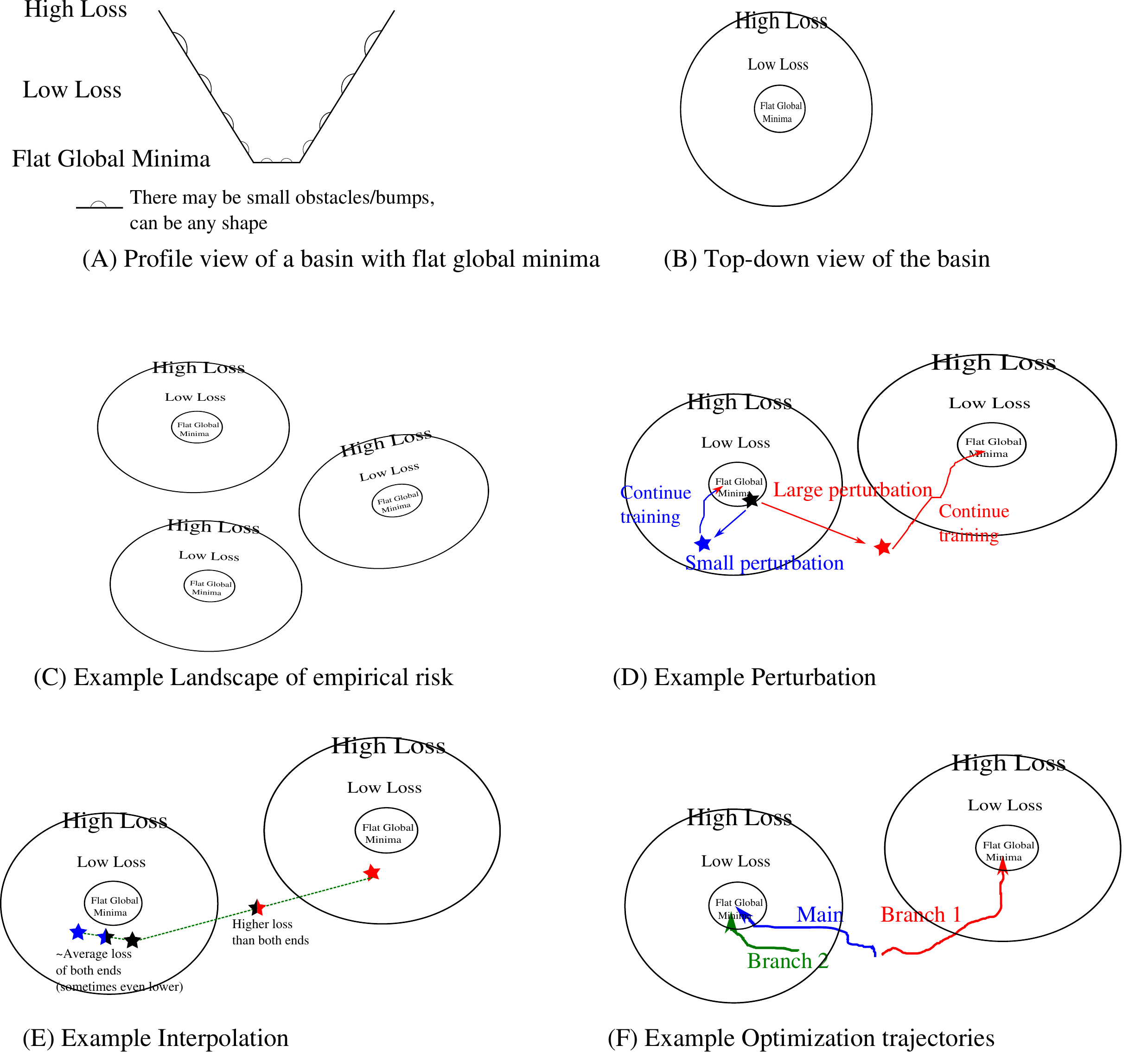} 
\caption{The Landscape of empirical risk of overparametrized DCNN may
  be simply a collection of (perhaps slightly rugged) basins. (A) the
  profile view of a basin (B) the top-down view of a basin (C) example
  landscape of empirical risk (D) example perturbation: a small
  perturbation does not move the model out of its current basin, so
  re-training converges back to the bottom of the same basin. If the
  perturbation is large, re-training converges to another basin. (E)
  Example Interpolation: averaging two models within a basin tend to
  give a error that is the average of the two models (or less).
  Averaging two models between basins tend to give an error that is
  higher than both models. (F) Example optimization trajectories that
  correspond to Figure \ref{appendix:fig:branch_layer_2_all_perturb_0.25} and
  Figure \ref{appendix:fig:branch_layer_3_all_perturb_1}. }
\label{appendix:fig:landscape_model}   
\end{figure}

\section{Landscape of the Empirical Risk: Theoretical Analyses}
\label{appendix:landscape}
\label{appendix:sec:theoretical}


\subsection{Optimization of compositional functions: Bezout theorem}

The following analysis of the landscape of the empirical risk is based
on two assumptions that hold true in most applications of deep
convolutional networks:

\begin{enumerate}
\item overparametrization of the network, typically using several
  times more parameters (the weights of the network) than data points.
  In practice, even with data augmentation, one can always make the
  model larger to achieve overparametrization without sacrificing
  either the training or generalization performance.
\item each of the equations corresponding to zeros of the empirical
  risk (we assume a regression framework attempting to minimize a loss
  such as square loss) can be approximated by a polynomial equation in
  the weights, by a polynomial approximaton within $\epsilon$ (in the
  sup norm) of the RELU nonlinearity.
\end{enumerate}

The main observation is that the degree of each approximating equation
$\ell^d(\epsilon)$ is determined by the accuracy $\epsilon$ we desire
for approximating the ReLU activation by a univariate polynomial $P$
of degree $\ell(\epsilon)$ and by the number of layers $d$.\footnote
{Of course we have to constrain the range of values in the argument of
  the RELUs in order to set the degree of the polynomial $P$ that
  achieves accuracy $\epsilon$.} This approximation argument is of
course a qualitative one sice good approximation of the function does
not imply by itself good approximation of the number of its
zeros. Notice that we could abandon completely the approximation
approach and just consider the number of zeros when the RELUs are
exactly a low order univariate polinomial. The argument then would not
apply to RELU networks but would still carry weight because the two
types of networks -- with polynomial activation and with RELUs --
behave theoretically (see ) and empirically in a similar way.  In the
well-determined case (as many unknown weights as equations, that is
data points), Bezout theorem provides an upper bound on the number of
solutions. {\it The number of distinct zeros} (counting points at
infinity, using projective space, assigning an appropriate
multiplicity to each intersection point, and excluding degenerate
cases) would be {\it equal to $Z$ - the product of the degrees of each
  of the equations}.  Since the system of equations is usually
underdetermined -- as many equations as data points but more unknowns
(the weights) -- we expect an infinite number of global minima, under
the form of $Z$ {\it regions} of zero empirical error. If the
equations are inconsistent there are still many global minima of the
squared error that is solutions of systems of equations with a similar
form.
   

\subsection{Global minima with zero error}

We consider a simple example in which the zeros of the empirical error
(that is exact solutions of the set of equations obtained by setting
$f(x^i) - y_i= 0$, where $i=1,\cdots,N$ are the data points and $f(x)$
is the network parametrized by weights $w$). In particular, we consider
the zeros on  the training set of a network with ReLUs activation
approximating a function of four variables with the following structure:

\begin{equation}
f(x_1, x_2, x_3, x_4) = g (h(x_1, x_2), h'(x_3, x_4)).
\end{equation}
\noindent We assume that the deep approximation network uses ReLUs as follows
\begin{equation}
g(h,h)= A (W_1 h + W_2 h +W_3)_++B (W'_1 h + W'_2 h +W'_3)_+
\end{equation}

\noindent and
\begin{equation} 
h(x_1, x_2) = a (w_1 x_1 +w_2
x_2 +w_3)_+ + b (v_1 x_1 + v_2
x_2 +v_3)_+,
\end{equation}

\begin{equation} 
h'(x_3, x_4) = a' (w'_1 x_1 +w'_2
x_2 +w'_3)_+ + b' (v_1' x_1 + v_2'
x_2 +v_3')_+ .
\end{equation}

There are usually quite a few more units in the first and second layer
than the $2$ of the example above.

This example generalizes to the case in which the kernel support is
larger than $2$ (for instance is $3x3=9$ as in ResNets). In the
standard case each node (for instance in the first layer) contains quite a
few units ($O(100)$) and as many outputs. However, the effective
outputs are much fewer so that each is the linear combination of
several ReLUs units. 

\begin{figure*}
\centering
\includegraphics[width=0.75\textwidth]{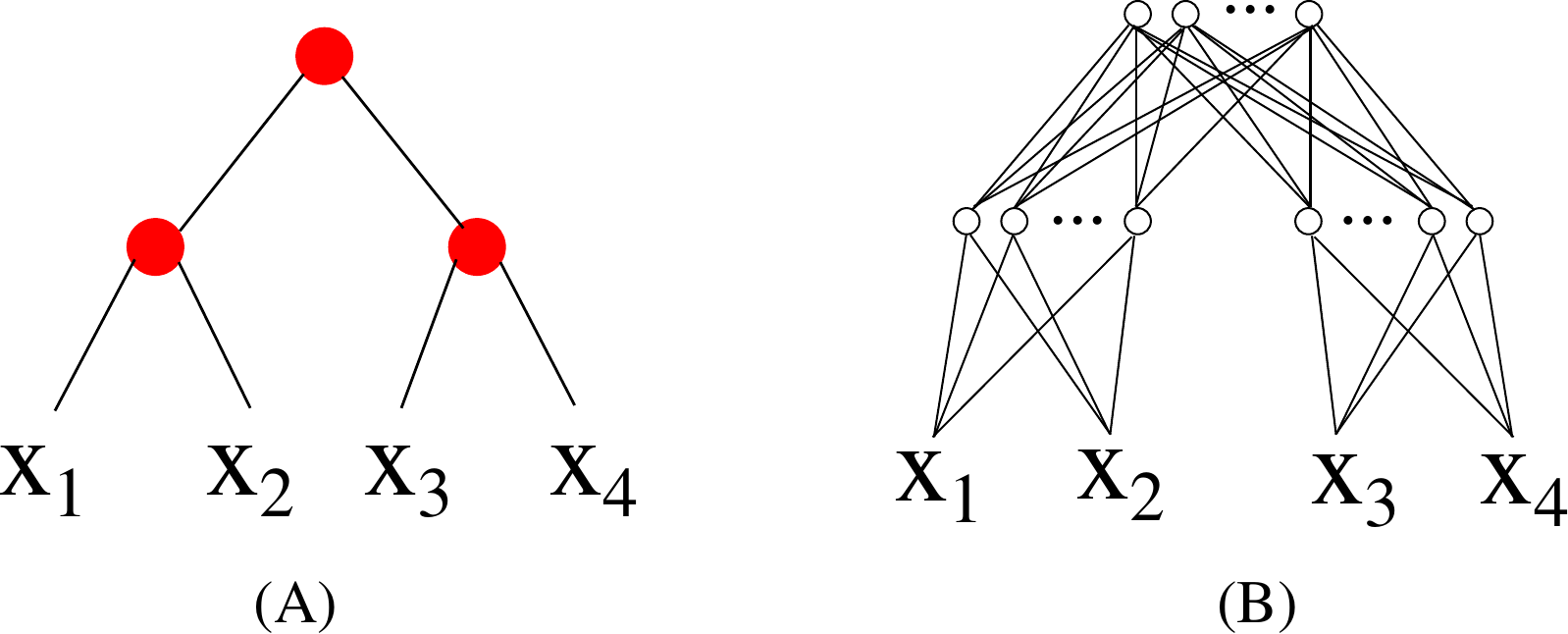}  
\caption{ The diagram (B) shows a 3 nodes part of a network; the
  corresponding compositional function of the form
  $f(x_1, \cdots, x_4) = h_{2}(h_{11} (x_1, x_2), h_{12}(x_3, x_4))$
  represented by the binary graph (A).  In the approximating network
  each of the $3$ nodes in the graph  corresponds to
  a set of $Q$ ReLU units computing the ridge function
  $\sum_{i=1}^Q a_i(\scal{\mathbf{v}_i}{\mathbf{x}}+t_i)_+$, with
  $\mathbf{v}_i, \mathbf{x} \in \R^2$, $a_i, t_i\in\R$. Corollary
  \ref{appendix:rank} implies that in a network that has a consistent accuracy
  across the layers the matrix of the weights to the top node should 
  have a rank in the order of the number of inputs of the connected 
  nodes at the lower layer: in this case the rank is $4$.} 
\label{appendix:3nodes_a+b}
\end{figure*}

Consider Figure \ref{appendix:3nodes_a+b}. The approximating polynomial
equations for the zero of the empirical errors for this network, which
could be part of a larger network, are, for $i=1,\cdots,N$ where $N$
are the data points:

\begin{equation}
P(W_1 P(w_{1}x_{i,1} +w_2 x_{i,2} +w_3)+W_2 P(w_{1}x_{i,3} +w_2
x_{i,4} +w_3) + 
\label{appendix:poly_of_poly}
\end{equation}
\begin{equation}
+W_3) -y_i=0
\end{equation}

The above equations describe the simple case of one ReLU per node for
the case of the network of Figure \ref{appendix:3nodes_a+b}.  Equations
\ref{appendix:poly_of_poly} are a system of underconstrained polynomial
equations of degree $l^d$. In general, there are as many constraints
as data points $i=1,\cdots,N$ for a much larger number $K$ of unknown
weights $W, w, \cdots$.  There are no solutions if the system is
inconsistent -- which happens if and only if $0 = 1$ is a linear
combination (with polynomial coefficients) of the equations (this is
Hilbert's Nullstellensatz).  Otherwise, it has infinitely many complex
solutions: the set of all solutions is an algebraic set of dimension
at least $K-N$. If the underdetermined system is chosen at random the
dimension is equal to $K-N$ with probability one.

Even in the non-degenerate case (as many data as parameters), Bezout
theorem suggests that there are many solutions. 
With $d$ layers the degree of the polynomial equations is $\ell^d$. With
$N$ datapoints the Bezout upper bound in the zeros of the weights is
$\ell^{Nd}$. Even if the number of real zero -- corresponding to zero
empirical error -- is much smaller (Smale and Shub estimate
\cite{ShubSmale94} $l^{\frac{Nd}{2}}$), the number is
still enormous: for a CiFAR situation this may be as high as $2^{10^5}$.

\subsection{Minima}

As mentioned, in several cases we expect absolute zeros to exist with
zero empirical error. If the equations are inconsistent it seems
likely that global minima with similar properties exist.

\begin{corollary}
  In general, non-zero minima  exist with higher dimensionality than
  the zero-error global minima: their dimensionality is the
  number of weights $K$ vs. the number of data points $N$. This
  is true in the linear case and  also in the presence of ReLUs.
\end{corollary}

Let us  consider the same example as before looking at the critical
points of the gradient. With a square loss function the critical
points of the gradient are:

\begin{equation}
\nabla_w\sum_{i=1}^N(f(x_i) - y_i)^2=0
\end{equation}

\noindent which gives $K$ equations

\begin{equation}
\sum_{i=1}^N(f(x_i) - y_i) \nabla_w f(x_i)=0.     
\end{equation}

Approximating within $\epsilon$ in the sup norm each ReLU in $f(x_i)$
with a fixed polynomial $P(z)$ yields again a system of $K$ polynomial
equations in the weights of higher order than in the case of
zero-minimizers. They are of course satisfied by the degenerate zeros
of the empirical error but also by additional non-degenerate (in the
general case) solutions.

We summarize our main observations on the $\epsilon$ approximating
system of equations in the following

\begin{proposition}
 There are a very large number of zero-error minima which are highly degenerate unlike the local
  non-zero minima.
\end{proposition}

\section{The Landscape of the Empirical Risk: Visualizing and Analysing the Loss Surface During the Entire Training Process (on CIFAR-10)}
\label{appendix:sec:vis}

In the empirical work described below we analyze a classification
problem with cross entropy loss. Our theoretical analyses with the
regression framework provide a \textit{lower bound} of the number of
solutions of the classification problem.


\subsection{Experimental Settings}
Unless mentioned otherwise, we trained a 6-layer (with the 1st layer
being the input) Deep Convolutional Neural Network (DCNN) on
CIFAR-10. All the layers are 3x3 convolutional layers with stride
2. No pooling is performed. Batch Normalizations (BNs) are used
between hidden layers. The shifting and scaling parameters in BNs are
not used. No data augmentation is performed, so that the training set
is fixed (size = 50,000). There are 188,810 parameters in the DCNN.

\textbf{Multidimensional Scaling} The core of our visualization
approach is Multidimensional Scaling (MDS) \cite{borg2005modern}. We
record a large number of intermediate models during the process of
several training schemes. Each model is a high dimensional point with
the number of dimensions being the number of parameters. The
strain-based MDS algorithm is applied to such points and a
corresponding set of 2D points are found such that the dissimilarity
matrix between the 2D points are as similar to those of the
high-dimensional points as possible. One minus cosine distance is used
as the dissimilarity metric. This is more robust to scaling of the
weights, which is usually normalized out by BNs. Euclidean distance
gives qualitatively similar results though.

\subsection{Global Visualization of SGD Training Trajectories}

We show the optimization trajectories of Stochastic Gradient Descent
(SGD), since this is what people use in practice.  The SGD
trajectories follow the mini-batch approximations of the training loss
surface. Although the trajectories are noisy due to SGD,
the collected points along the trajectories provide a visualization of the
landscape of empirical risk.

We train a 6-layer (with the 1st layer being the input) convolutional
network on CIFAR-10 with stochastic gradient descent (batch size =
100) We divide the training process into 12 stages. In each stage, we
perform \textbf{8 parallel} SGDs with learning rate 0.01 for 10
epochs, resulting in 8 parallel trajectories denoted by different
colors in each subfigure of Figure \ref{appendix:fig:layer_2_a_stage_0} and
\ref{appendix:fig:layer_2_a_stage_4}. Trajectories 1 to 4 in each stage start
from the final model (denoted by $P$) of trajectory 1 of the previous
stage. Trajectories 5 to 8 in each stage start from a perturbed
version of $P$. The perturbation is performed by adding a gaussian
noise to the weights of each layer with the standard deviation being
0.01 times layer's standard deviation. In general, we observe that 
running any trajectory with SGD again almost always leads to a
slightly different model.

\begin{figure*}\centering 
\includegraphics[width=0.9\textwidth]{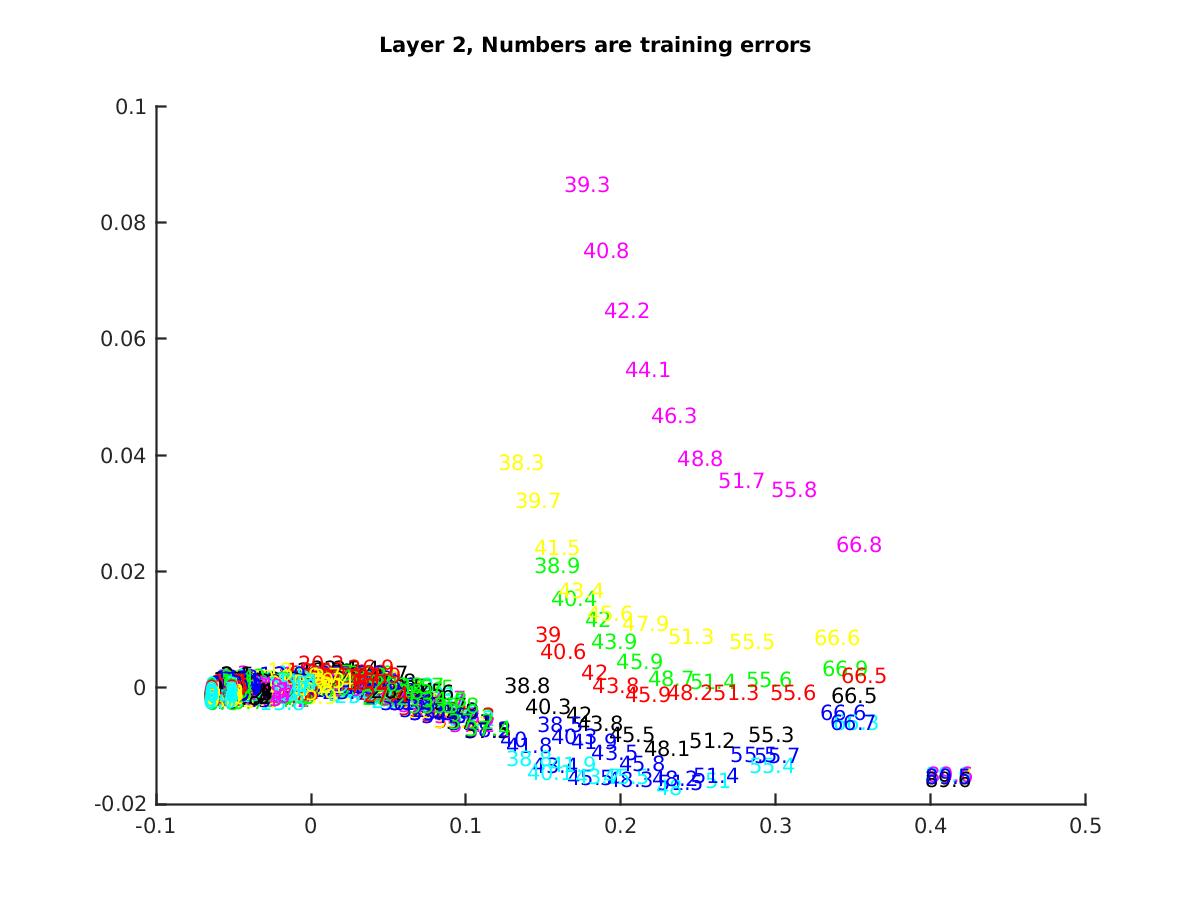}
\caption{Training a DCNN on CIFAR-10. We divide the training process
  into 12 stages. In each stage, we perform \textbf{8 parallel} SGDs
  with learning rate 0.01 for 10 epochs, resulting in 8 parallel
  trajectories denoted by different colors. Trajectories 1 to 4 in
  each stage start from the final model (denoted by $P$) of trajectory
  1 of the previous stage. Trajectories 5 to 8 in each stage start
  from a perturbed version of $P$. The perturbation is performed by
  adding a gaussian noise to the weights of each layer with the
  standard deviation being 0.01 times layer's standard deviation. In
  general, we observe that running any trajectory with SGD again
  almost always leads to a slightly different convergence path. We
  plot the MDS results of all the layer 2 weights collected throughout
  all the training epochs from stage 1 to 12. The detailed parallel
  trajectories of stage 1 to 3 are plotted separately in Figure
  \ref{appendix:fig:layer_2_a_stage_0}. Each number in the figure represents a
  model we collected during the above procedures. The points are in a
  2D space generated by the MDS algorithm such that their pairwise distances
  are optimized to try to reflect those distances in the original high-dimensional space. 
  The results of stages more than 5 are quite cluttered. So we applied a
  separate MDS to the stages 5 to 12 and show the results in Figure
  \ref{appendix:fig:global_vis_layer2_stage_4}. The detailed parallel
  trajectories of stage 5 to 7 are plotted separately in Figure
  \ref{appendix:fig:layer_2_a_stage_4}. }
\label{appendix:fig:global_vis_layer2_stage_0} 
\end{figure*}

\begin{figure}
%
\begin{tabular}{cc}
  \subfloat[In stage 1, the starting model is the randomly initialized model.]{\includegraphics[width=0.5\textwidth]{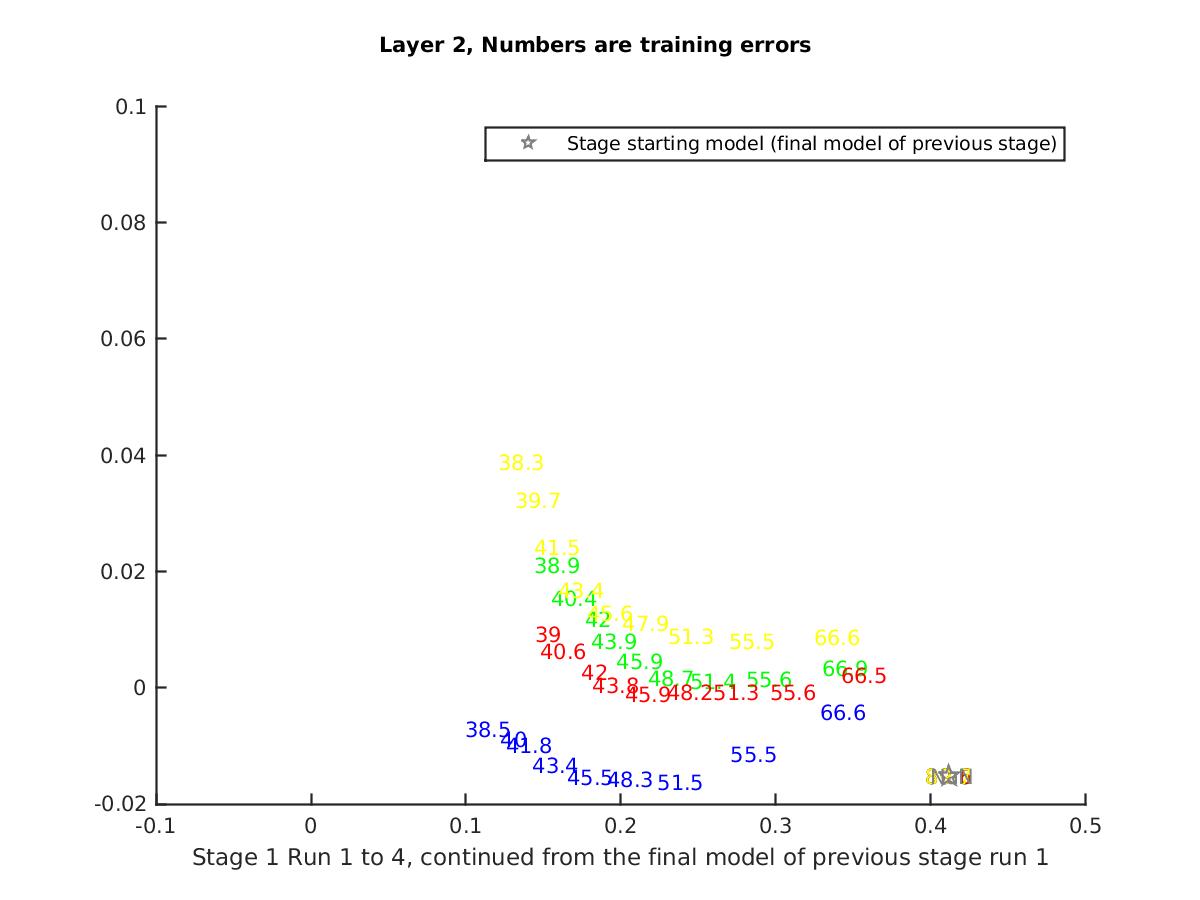}}   & \subfloat[]{\includegraphics[width=0.5\textwidth]{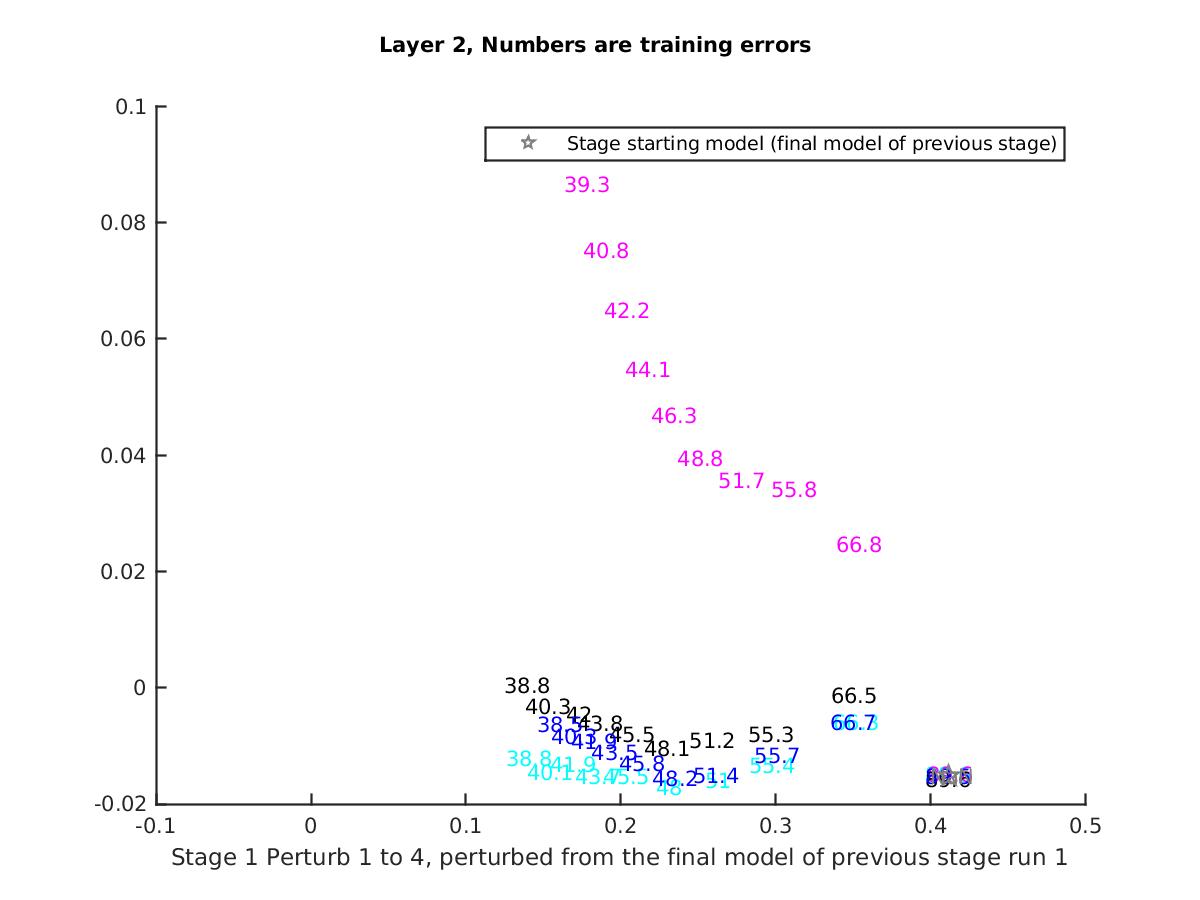}} \\
  \subfloat[]{\includegraphics[width=0.5\textwidth]{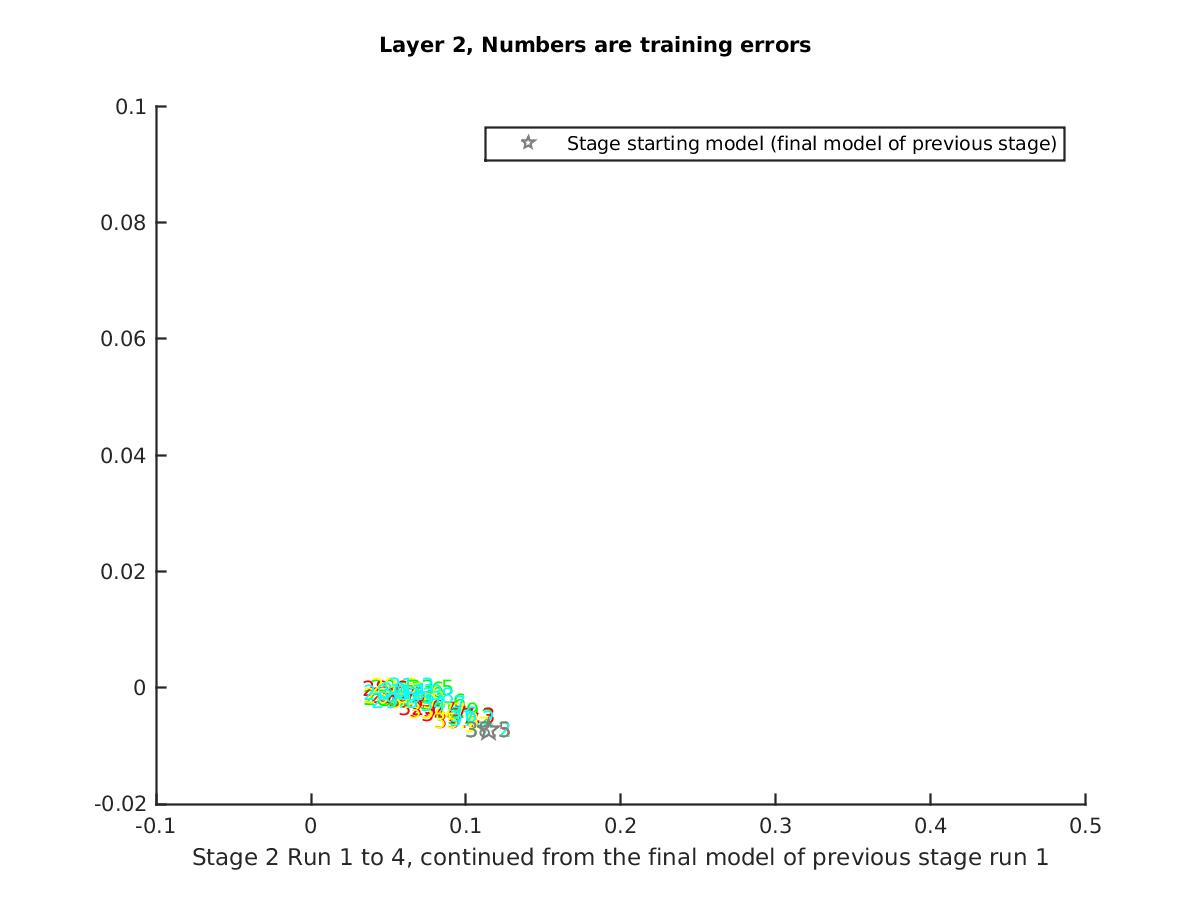}}   & \subfloat[]{\includegraphics[width=0.5\textwidth]{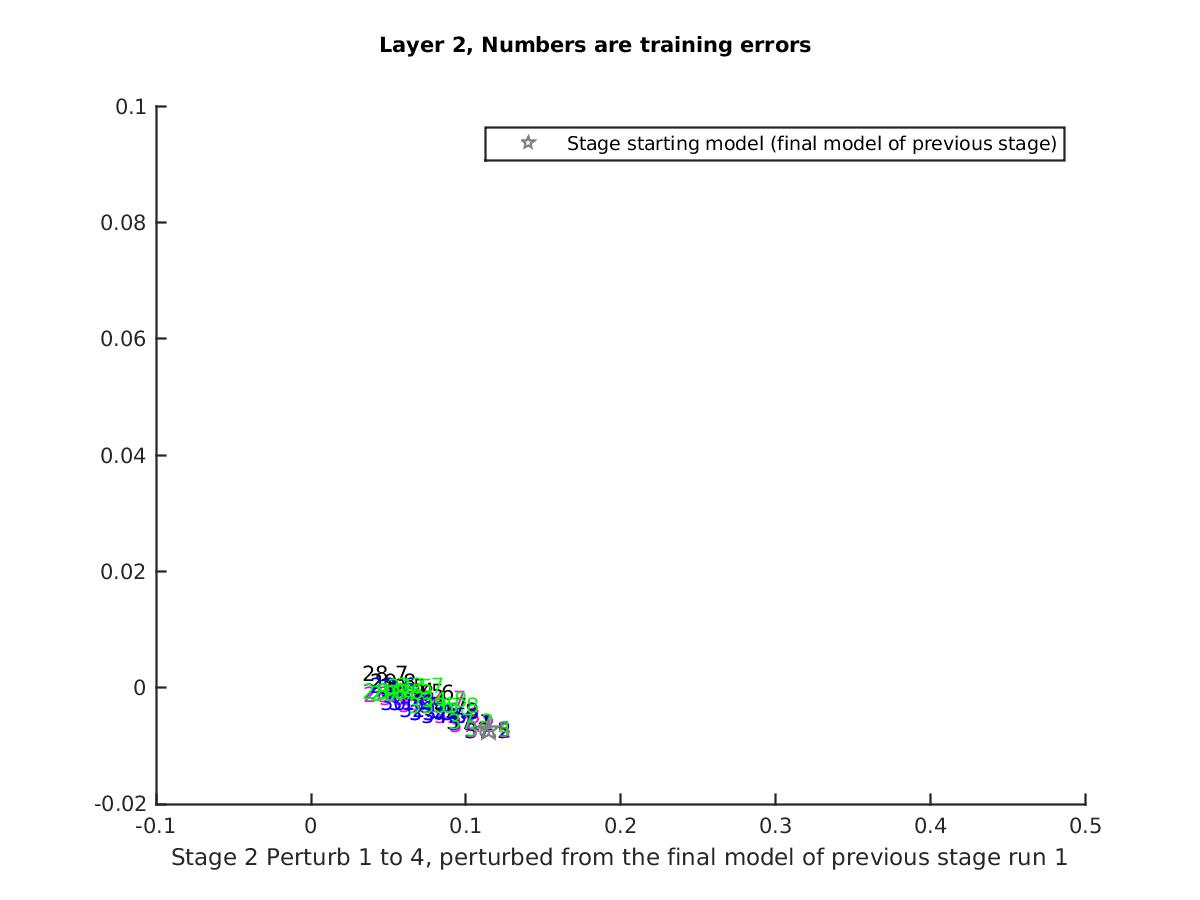}} \\
  \subfloat[]{\includegraphics[width=0.5\textwidth]{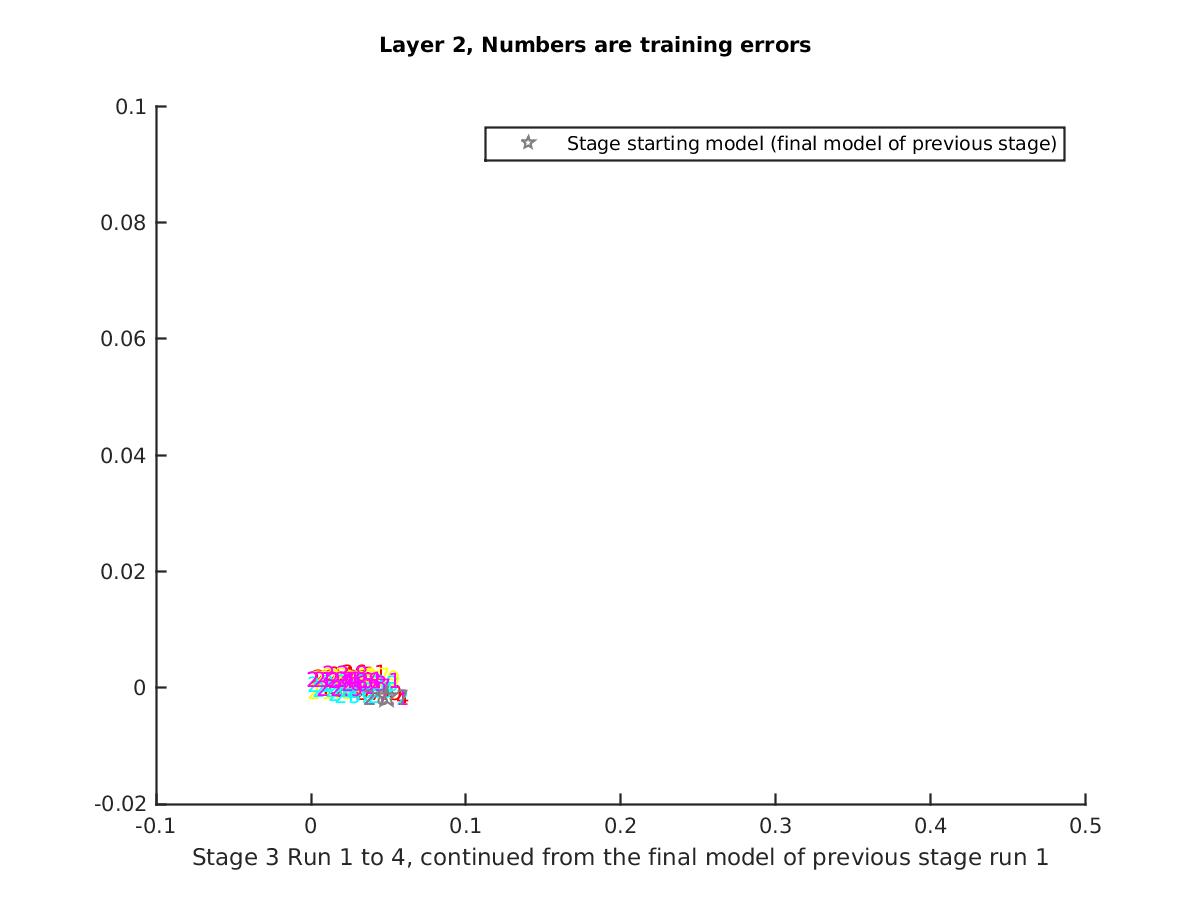}}   & \subfloat[]{\includegraphics[width=0.5\textwidth]{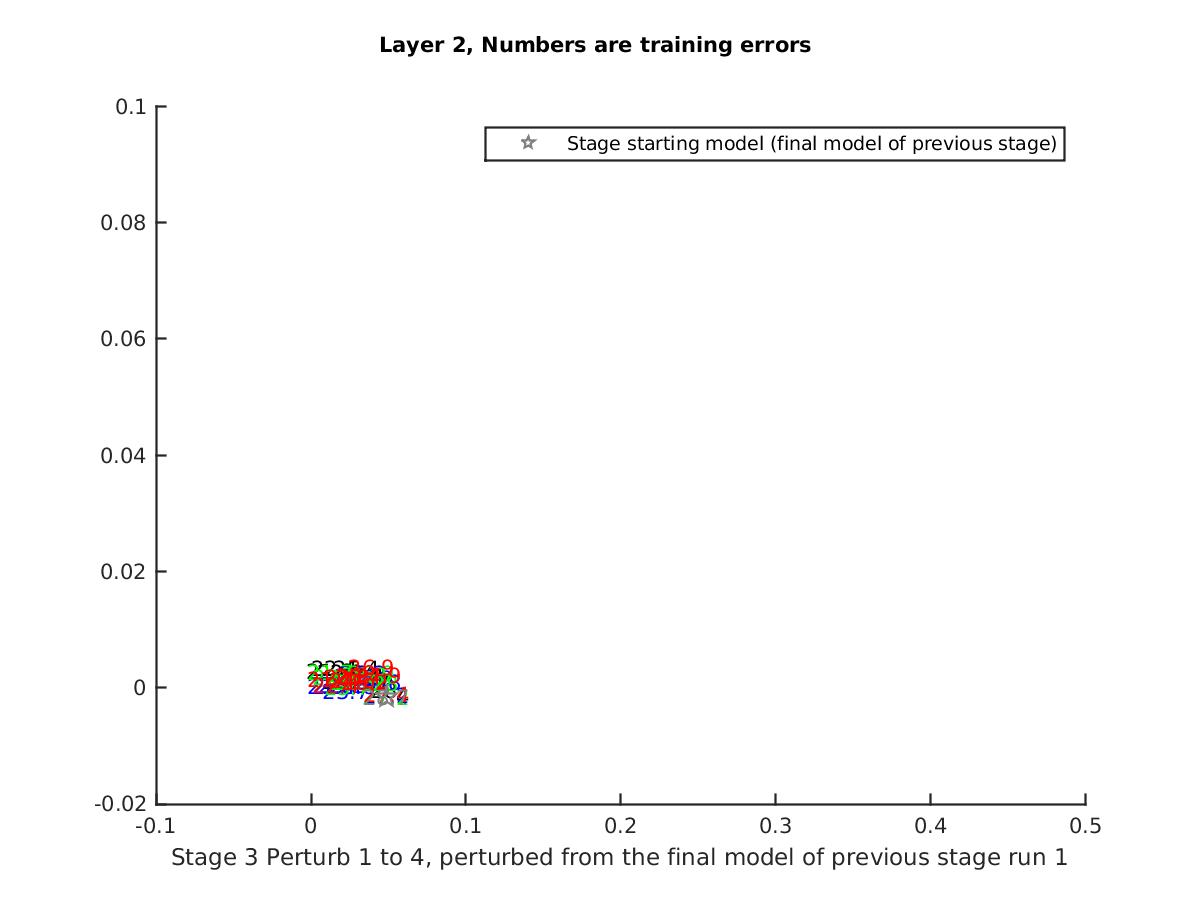}} \\
\end{tabular} 
\caption{Parallel trajectories from stage 1 to 3 of Figure \ref{appendix:fig:global_vis_layer2_stage_0} are plotted separately. More details are described in Figure \ref{appendix:fig:global_vis_layer2_stage_0} }\label{appendix:fig:layer_2_a_stage_0} 
\end{figure}

\begin{figure}
%
\begin{tabular}{cc}
  \subfloat[]{\includegraphics[width=0.5\textwidth]{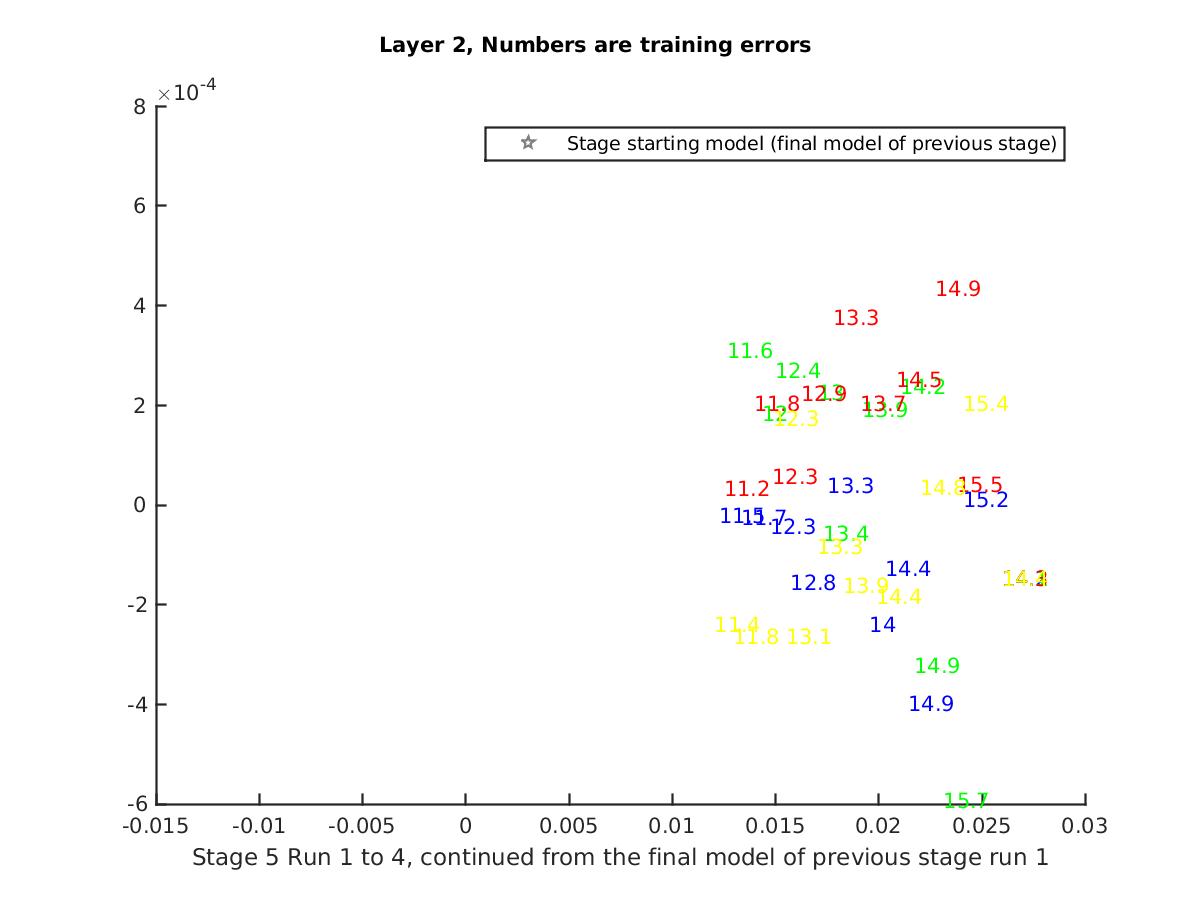}}   & \subfloat[]{\includegraphics[width=0.5\textwidth]{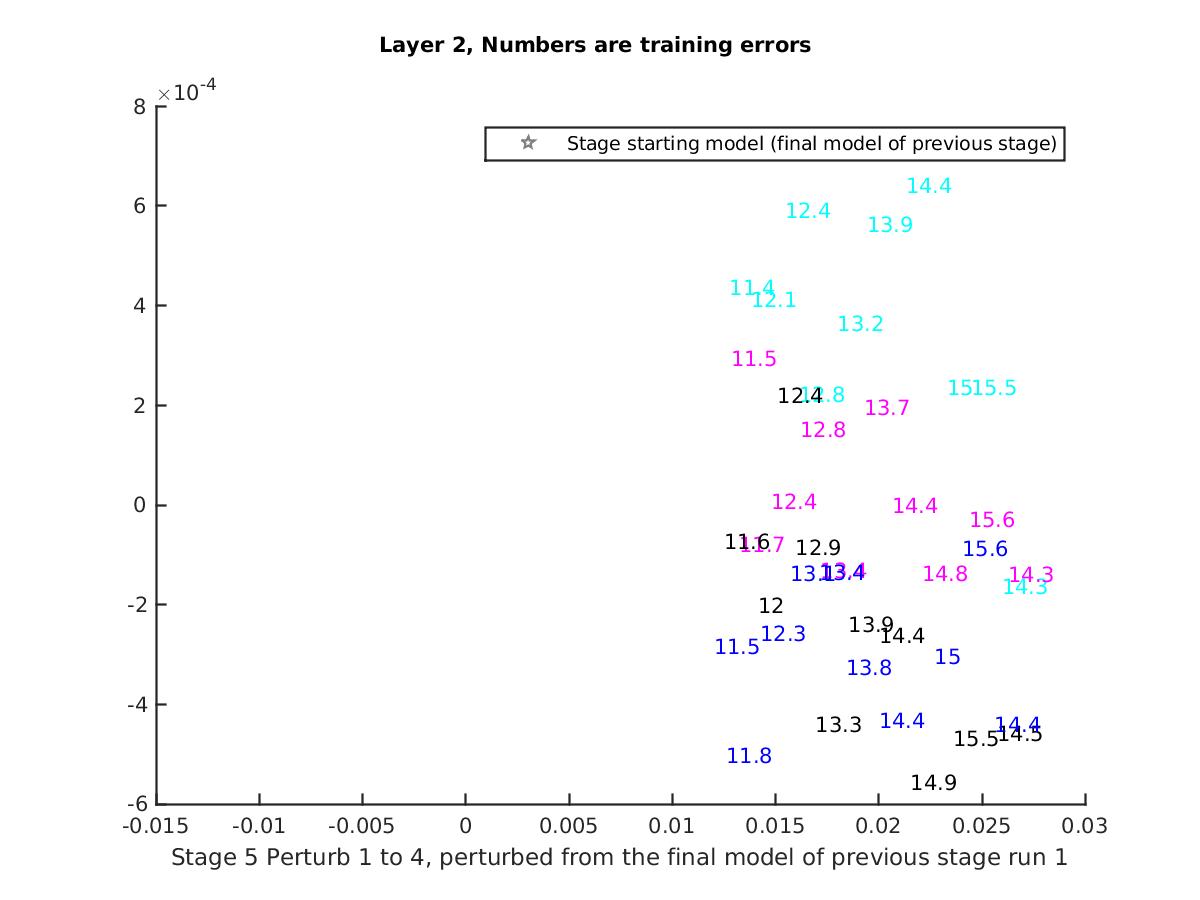}} \\
  \subfloat[]{\includegraphics[width=0.5\textwidth]{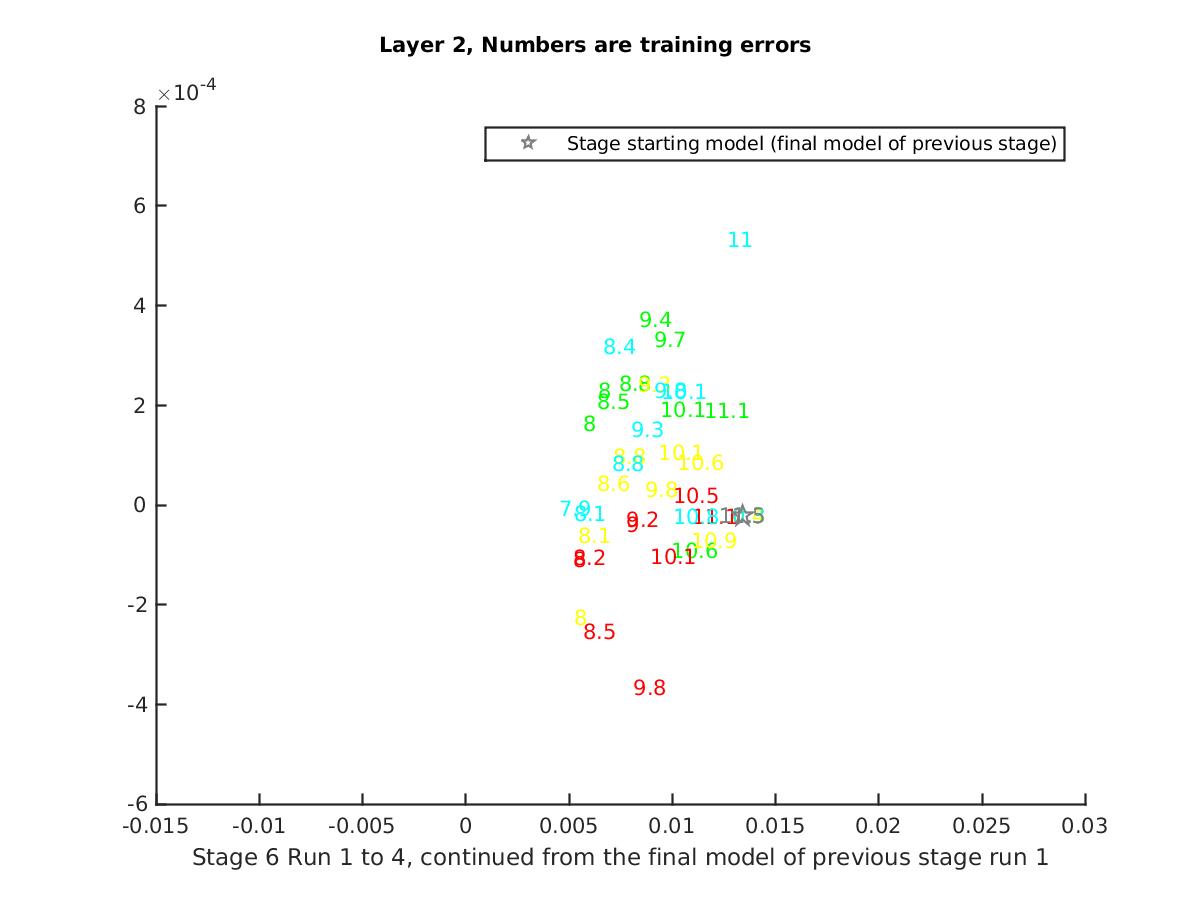}}   & \subfloat[]{\includegraphics[width=0.5\textwidth]{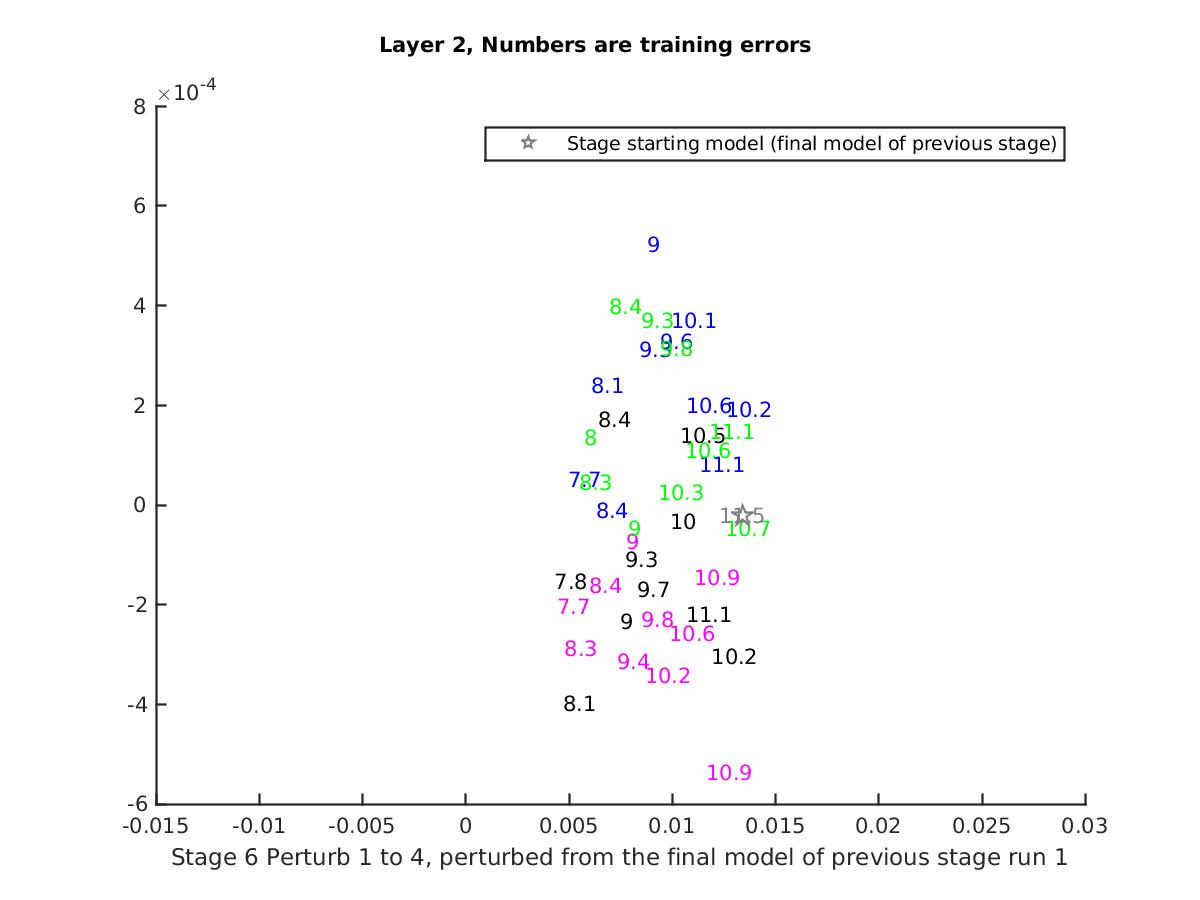}} \\
  \subfloat[]{\includegraphics[width=0.5\textwidth]{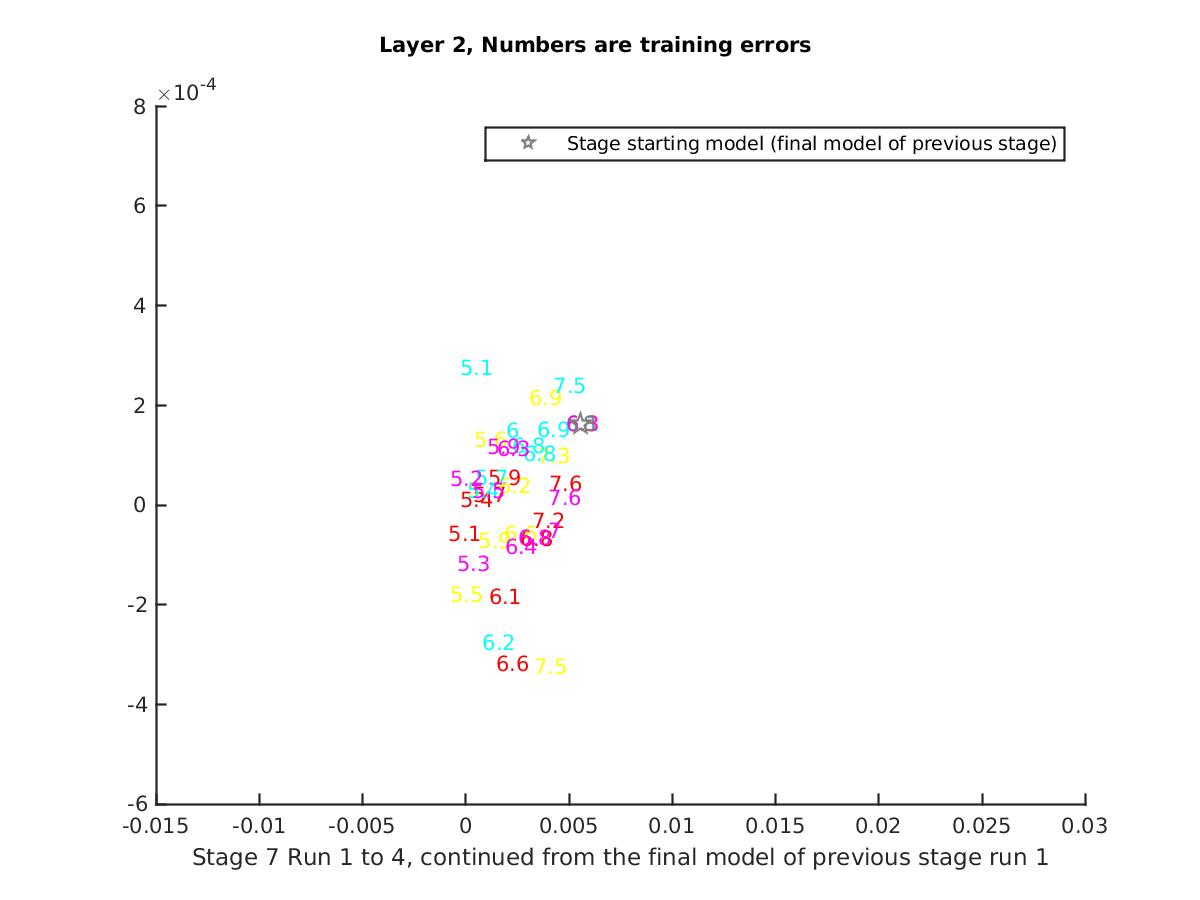}}   & \subfloat[]{\includegraphics[width=0.5\textwidth]{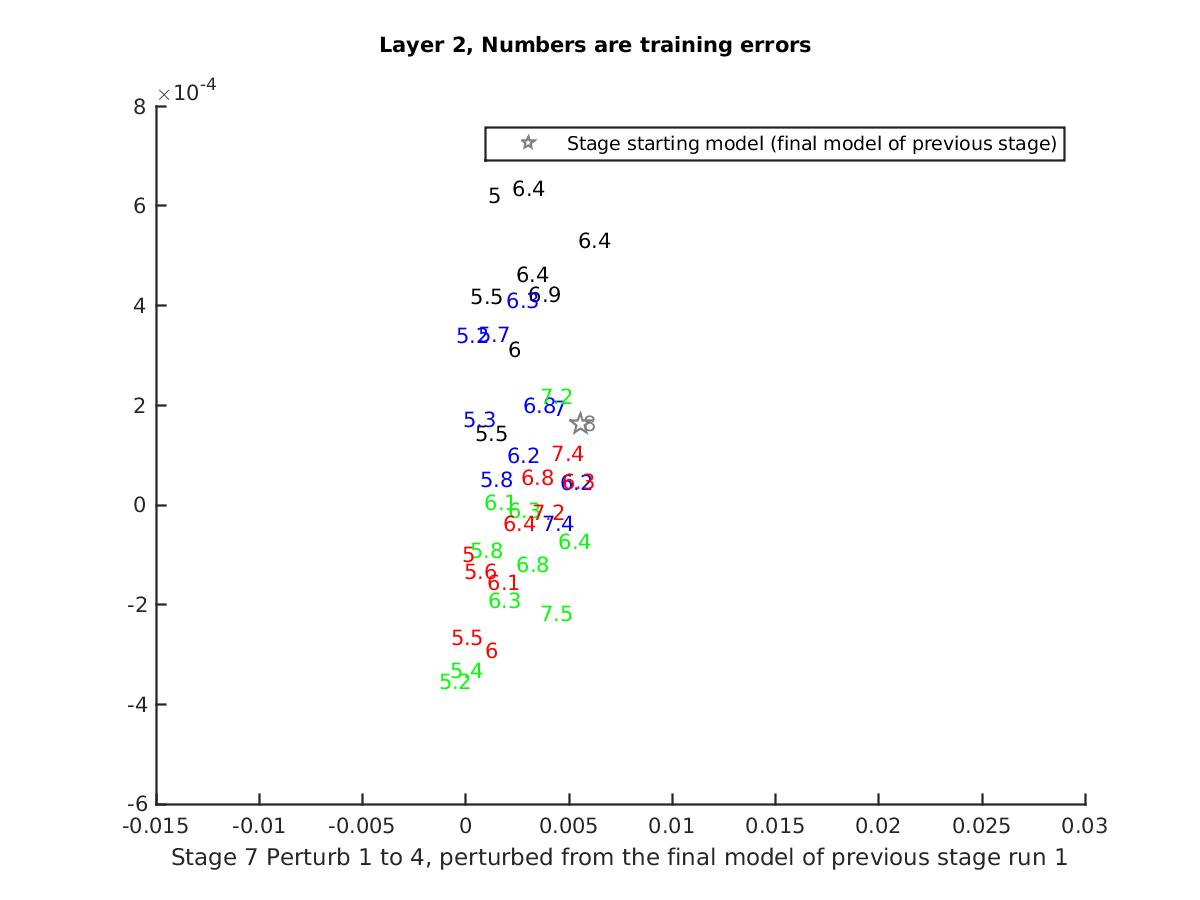}} \\ 
\end{tabular}   
\caption{Parallel trajectories from stage 5 to 7 of Figure \ref{appendix:fig:global_vis_layer2_stage_0} are plotted separately. More details are described in Figure \ref{appendix:fig:global_vis_layer2_stage_0} }\label{appendix:fig:layer_2_a_stage_4} 
\end{figure}

Taking layer 2 weights for example, we plot the global MDS results of
stage 1 to 12 in Figure \ref{appendix:fig:global_vis_layer2_stage_0}. The
detailed parallel trajectories of stage 1 to 3 are plotted separately
in Figure \ref{appendix:fig:layer_2_a_stage_0}.

\begin{figure*}\centering 
\includegraphics[width=0.9\textwidth]{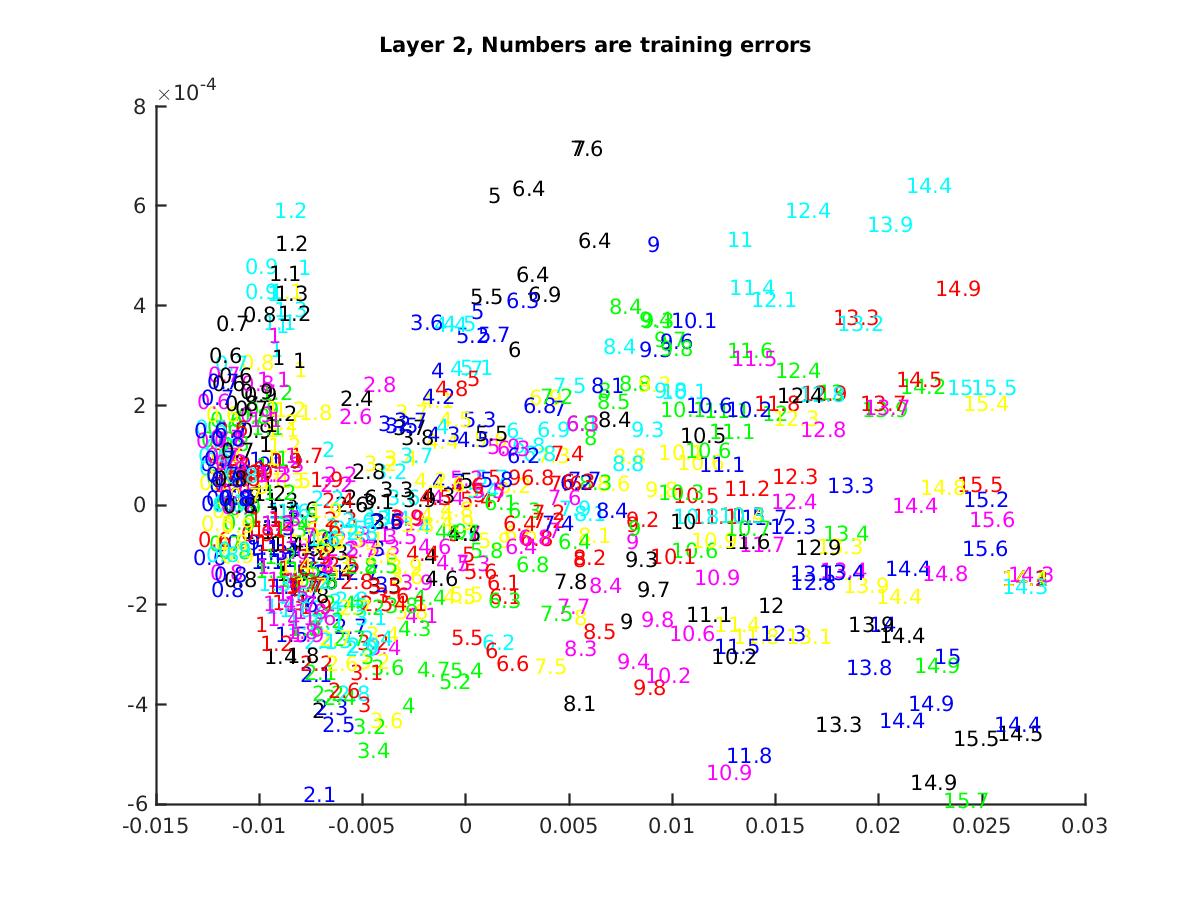}
\caption{A separate MDS performed on stage 5 to 12 of Figure \ref{appendix:fig:global_vis_layer2_stage_0} to provide more resolution. Details are described in the caption of Figure \ref{appendix:fig:global_vis_layer2_stage_0}}  
\label{appendix:fig:global_vis_layer2_stage_4} 
\end{figure*}

The results of stages more than 5 are quite cluttered. So we applied a
separate MDS to the stages 5 to 12 and show the results in Figure
\ref{appendix:fig:global_vis_layer2_stage_4}. The detailed parallel
trajectories of stage 5 to 7 are plotted separately in Figure
\ref{appendix:fig:layer_2_a_stage_4}.

\begin{figure*}\centering
  \includegraphics[width=0.9\textwidth]{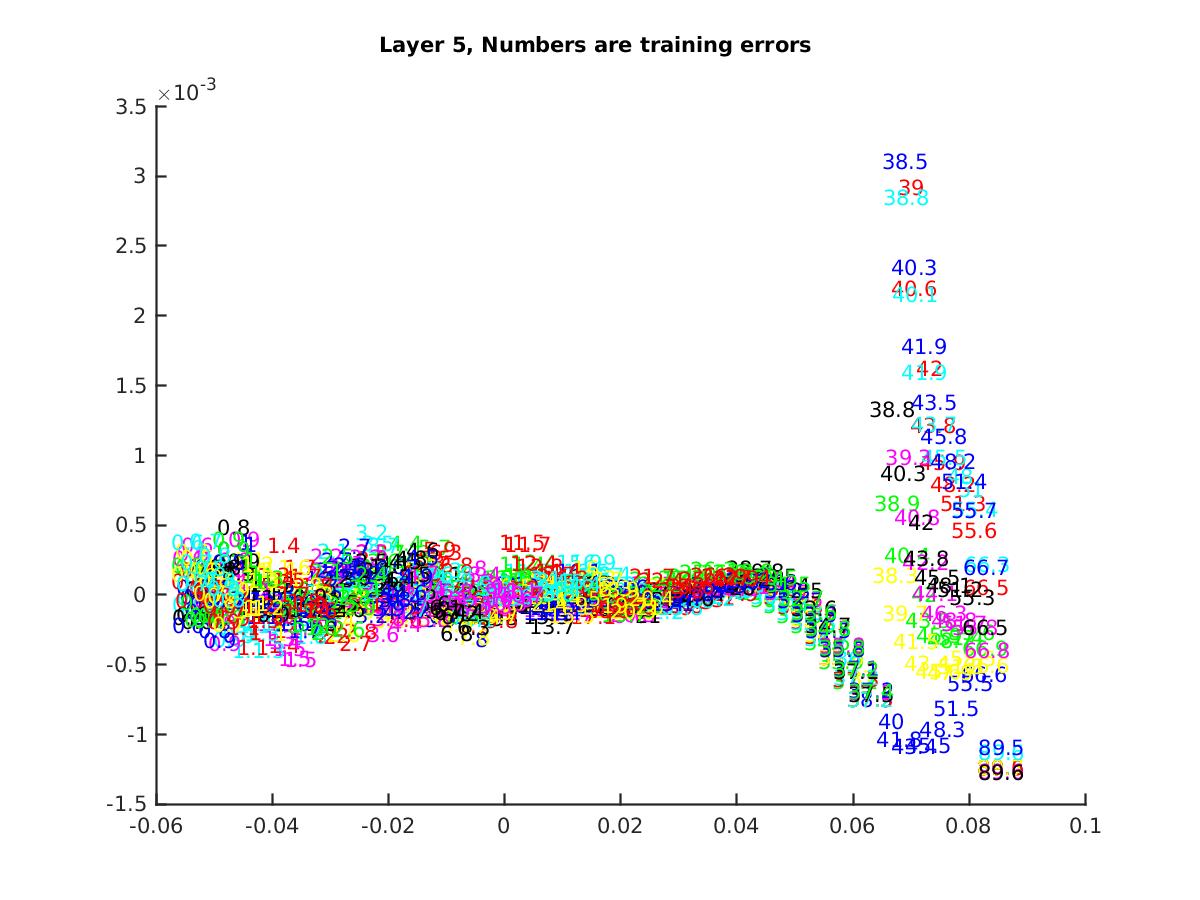}
\caption{Same as Figure \ref{appendix:fig:global_vis_layer2_stage_0}, but all weights are collected from Layer 5.}
\label{appendix:fig:global_vis_layer5_stage_0}
\end{figure*}

\begin{figure*}\centering
  \includegraphics[width=0.9\textwidth]{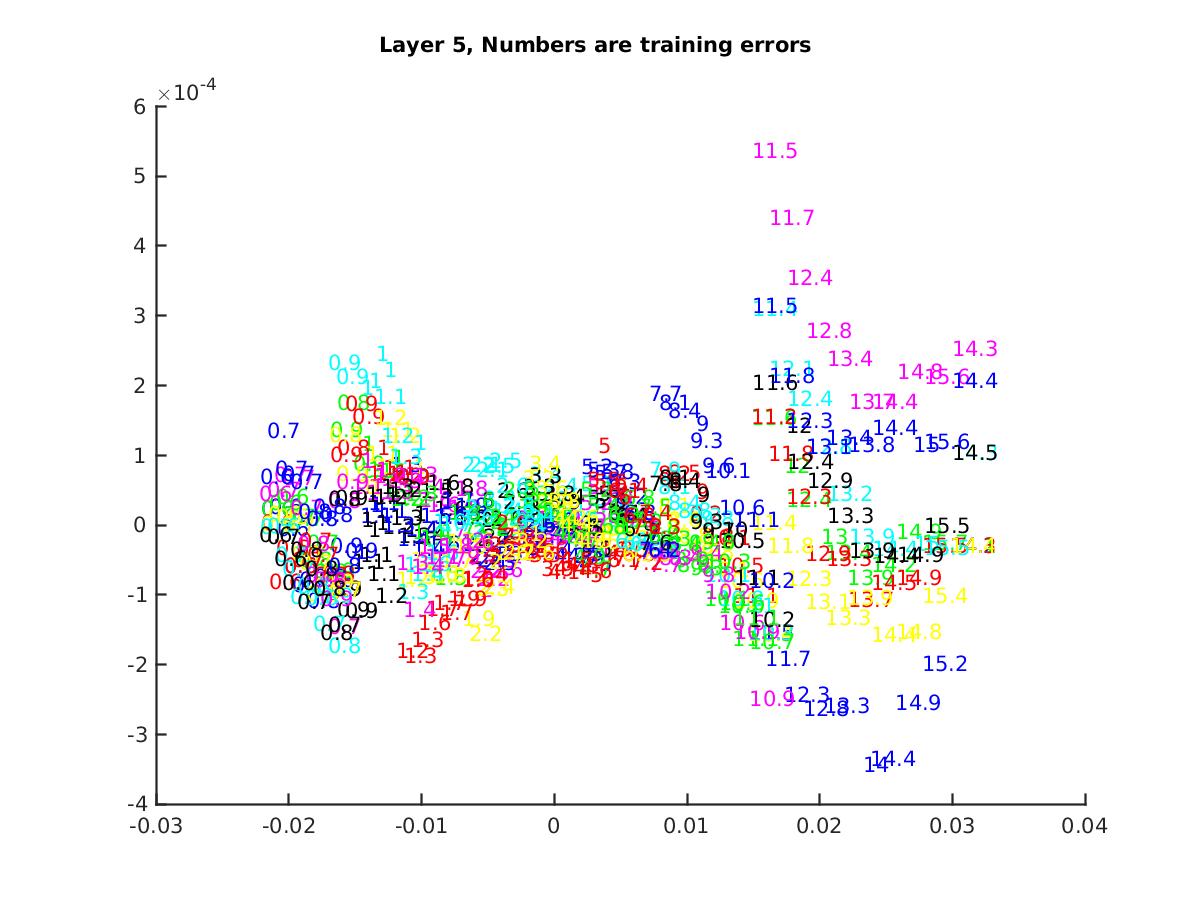} 
\caption{Same as Figure \ref{appendix:fig:global_vis_layer2_stage_4}, but all weights are collected from Layer 5.} 
\label{appendix:fig:global_vis_layer5_stage_4}
\end{figure*}

The weights of different layers tend to produce qualitatively similar
results. We show the results of layer 5 in Figure
\ref{appendix:fig:global_vis_layer5_stage_0} and Figure
\ref{appendix:fig:global_vis_layer5_stage_4}.

\subsection{Global Visualization of Training Loss Surface with Batch Gradient Descent}

Next, we visualize the exact training loss surface by training the
models using Batch Gradient Descent (BGD). We adopt the following
procedures: We train a model from scratch using BGD. At epoch 0, 10,
50 and 200, we create a branch by perturbing the model by adding a
Gaussian noise to all layers. The standard deviation of the Gaussian
noise is a meta parameter, and we tried 0.25*S, 0.5*S and 1*S, where S
denotes the standard deviation of the weights in each layer,
respectively.

We also interpolate (by averaging) the models between the branches and the main trajectory, epoch by epoch.
The interpolated models are evaluated on the entire training set to get a performance (in terms of error percentage).

The main trajectory, branches and the interpolated models together
provides a good visualization of the landscape of the empirical risk.

\begin{figure*}\centering 
\includegraphics[width=0.9\textwidth]{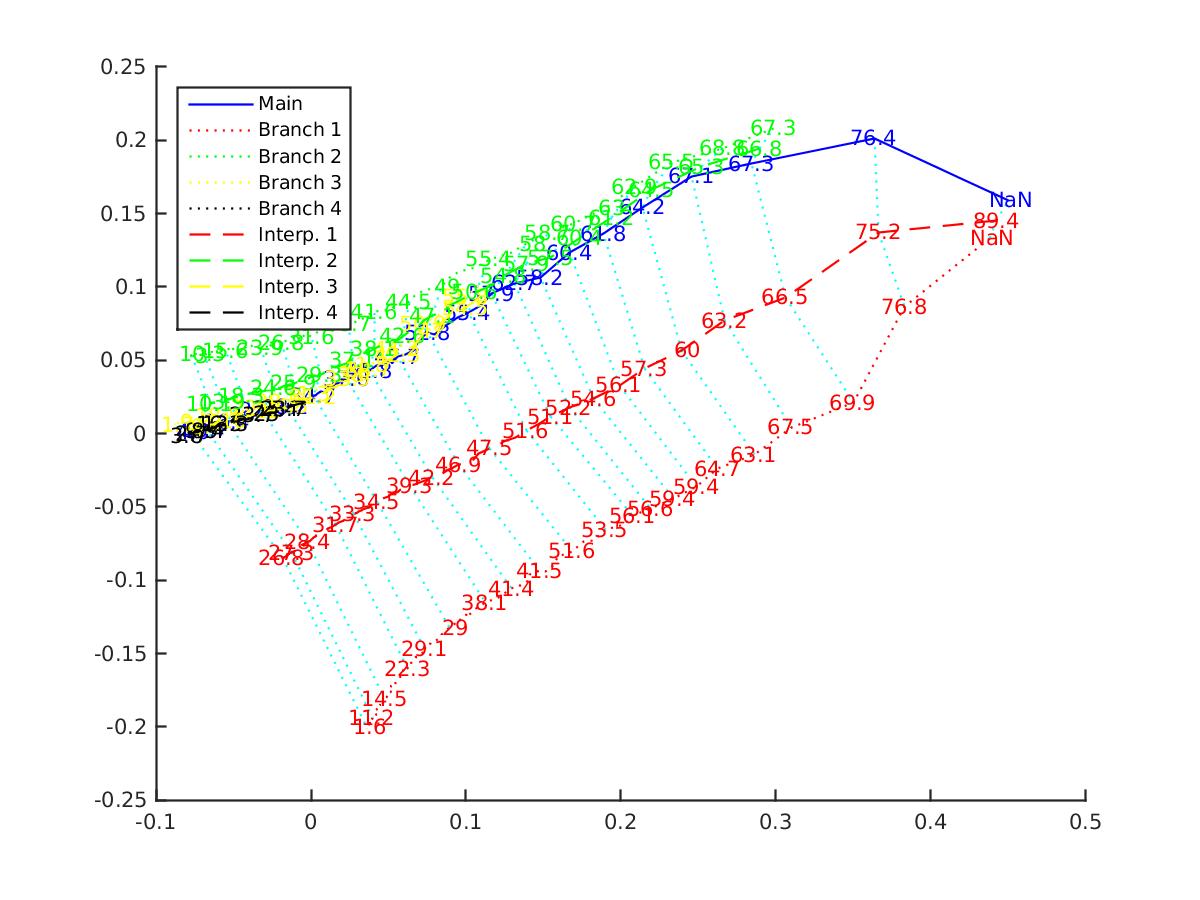}
\caption{The 2D visualization using MDS of weights in layer 2. A DCNN
  is trained on CIFAR-10 from scratch using Batch Gradient Descent
  (BGD). The numbers are training errors. ``NaN'' corresponds to
  randomly initialized models (we did not evaluate them and assume
  they perform at chance). At epoch 0, 10, 50 and 200, we create a
  branch by perturbing the model by adding a Gaussian noise to all
  layers. The standard deviation of the Gaussian is 0.25*S, where S
  denotes the standard deviation of the weights in each layer,
  respectively. We also interpolate (by averaging) the models between
  the branches and the main trajectory, epoch by epoch. The
  interpolated models are evaluated on the entire training set to get
  a performance. First, surprisingly, BGD does not get stuck in any
  local minima, indicating some good properties of the landscape. The
  test error of solutions found by BGD is somewhat worse than those
  found by SGD, but not too much worse (BGD ~ 40\%, SGD ~ 32\%) .
  Another interesting observation is that as training proceeds, the
  same amount of perturbation are less able to lead to a drastically
  different trajectory. Nevertheless, a perturbation almost always
  leads to at least a slightly different model. The local neighborhood
  of the main trajectory seems to be relatively flat and contain many
  good solutions, supporting our theoretical predictions. It is also
  intriguing to see interpolated models to have very reasonable
  performance. }
\label{appendix:fig:branch_layer_2_all_perturb_0.25}  
\end{figure*}

\begin{figure}
%
\begin{tabular}{cc}
  \subfloat[]{\includegraphics[width=0.5\textwidth]{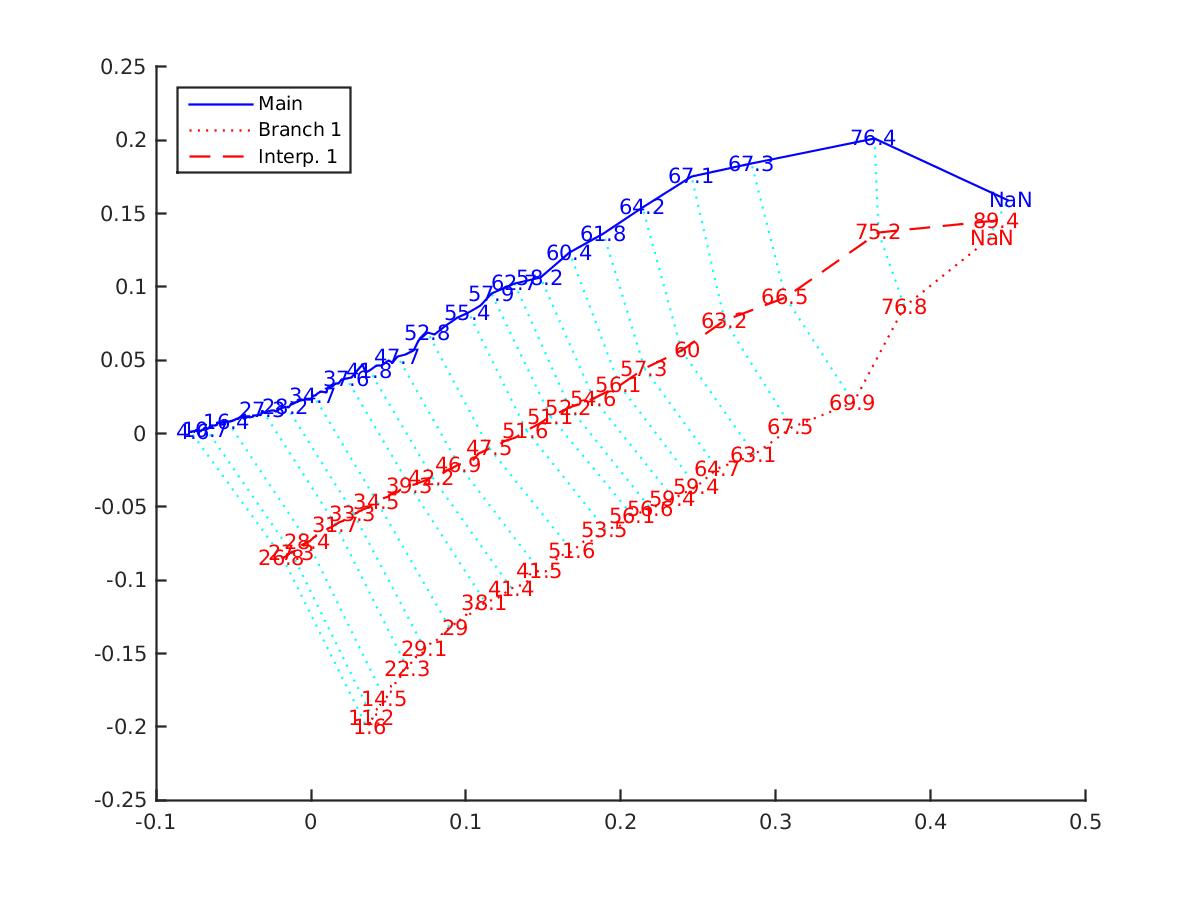}}   & \subfloat[]{\includegraphics[width=0.5\textwidth]{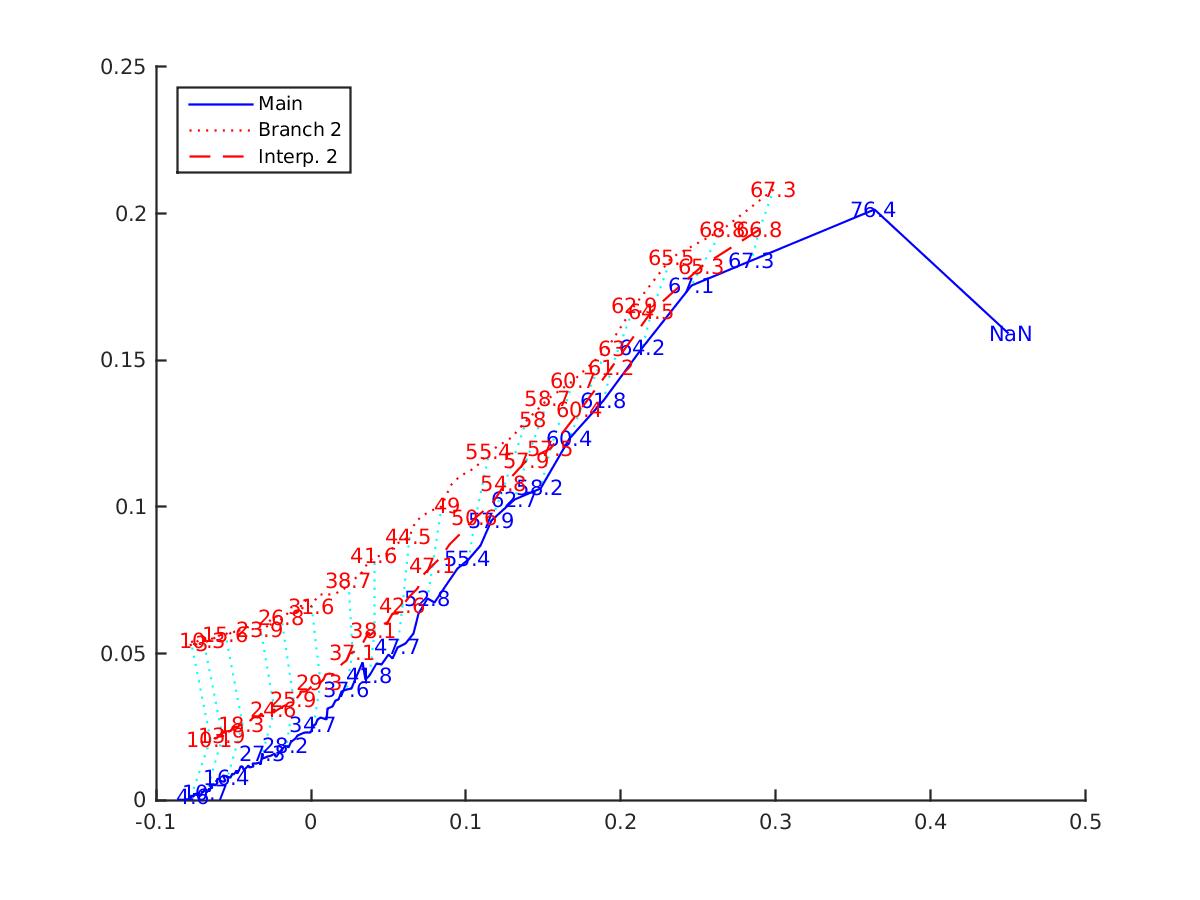}} \\  
  \subfloat[]{\includegraphics[width=0.5\textwidth]{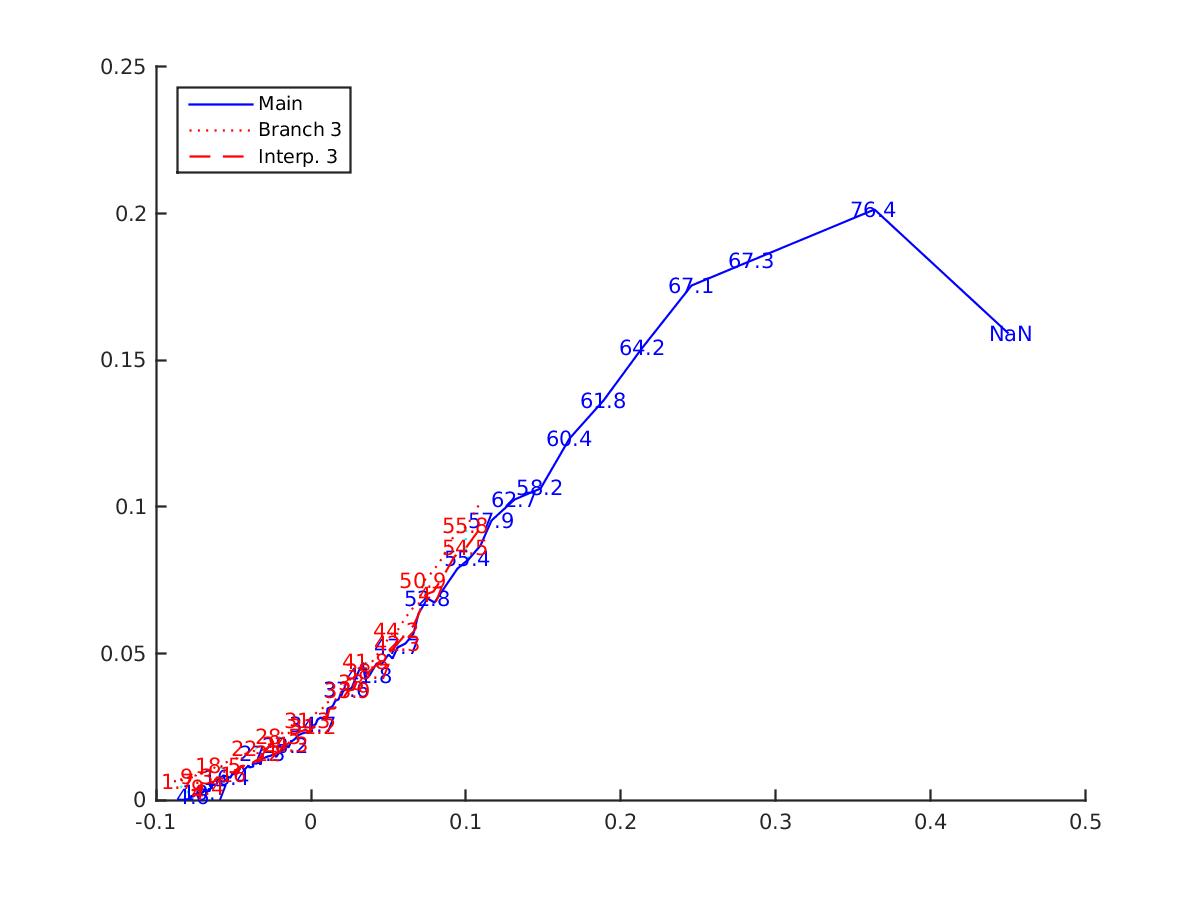}}   & \subfloat[]{\includegraphics[width=0.5\textwidth]{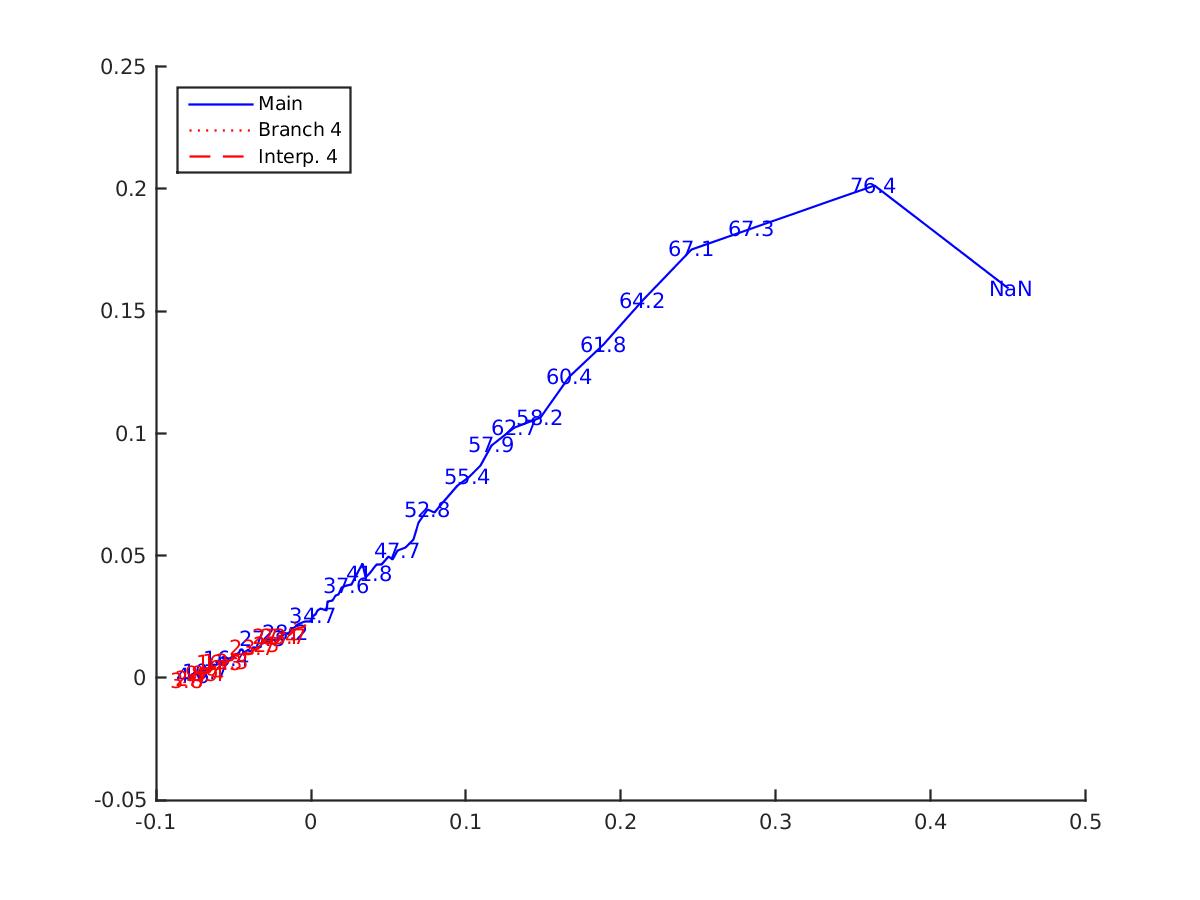}} \\
\end{tabular}   
\caption{Same as Figure \ref{appendix:fig:branch_layer_2_all_perturb_0.25}, but 4 branches are plotted separately to avoid clutter. }\label{appendix:fig:branch_layer_2_sep_perturb_0.25}
\end{figure}

\begin{figure*}\centering 
\includegraphics[width=0.9\textwidth]{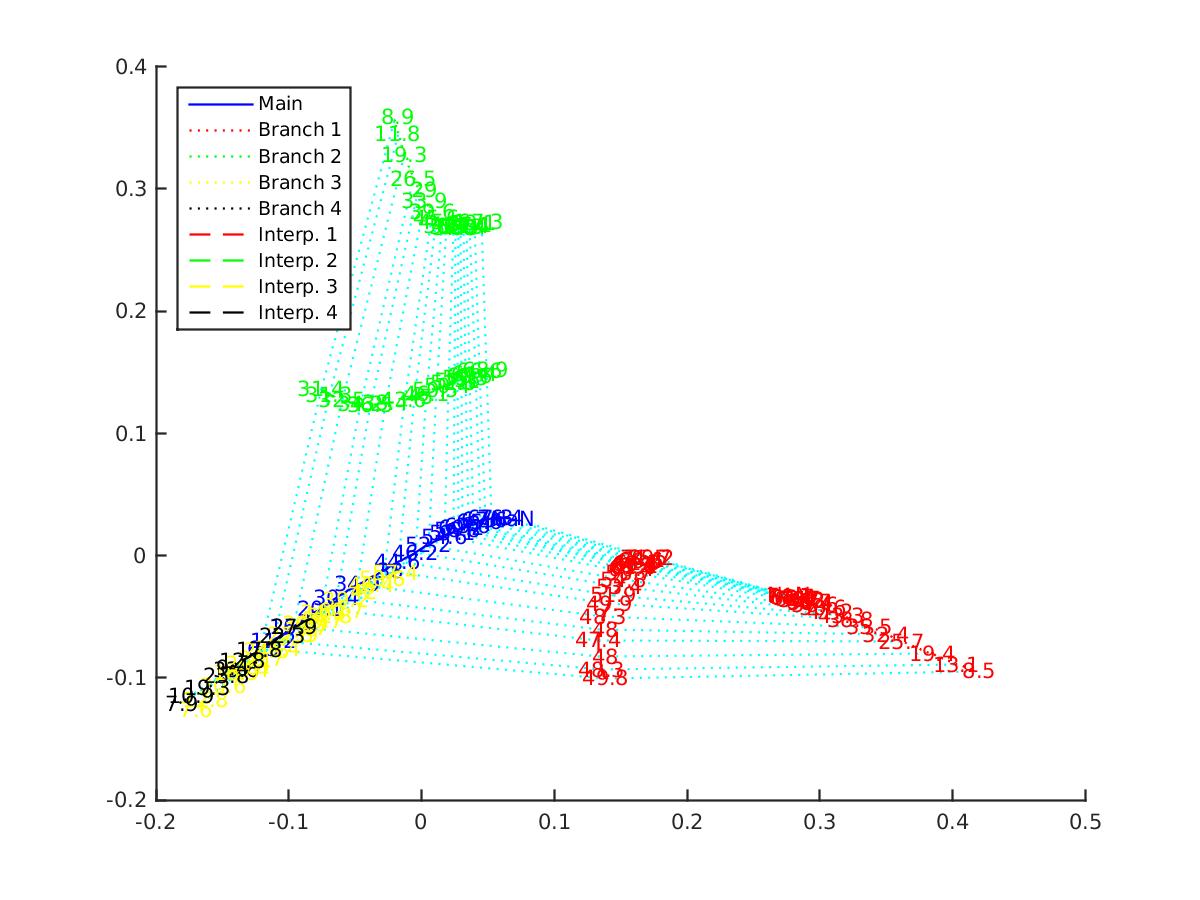}
\caption{Same as Figure \ref{appendix:fig:branch_layer_2_all_perturb_0.25}
  except: (1) the weights of layer 3 are visualized, instead of layer
  2. (2) the perturbation is significantly larger: we create a branch
  by perturbing the model by adding a Gaussian noise to all layers.
  The standard deviation of the Gaussian is S, where S denotes the
  standard deviation of the weights in each layer, respectively. Larger perturbations leads to more separated branches. } 
\label{appendix:fig:branch_layer_3_all_perturb_1} 
\end{figure*}

\begin{figure}
%
\begin{tabular}{cc}
  \subfloat[]{\includegraphics[width=0.5\textwidth]{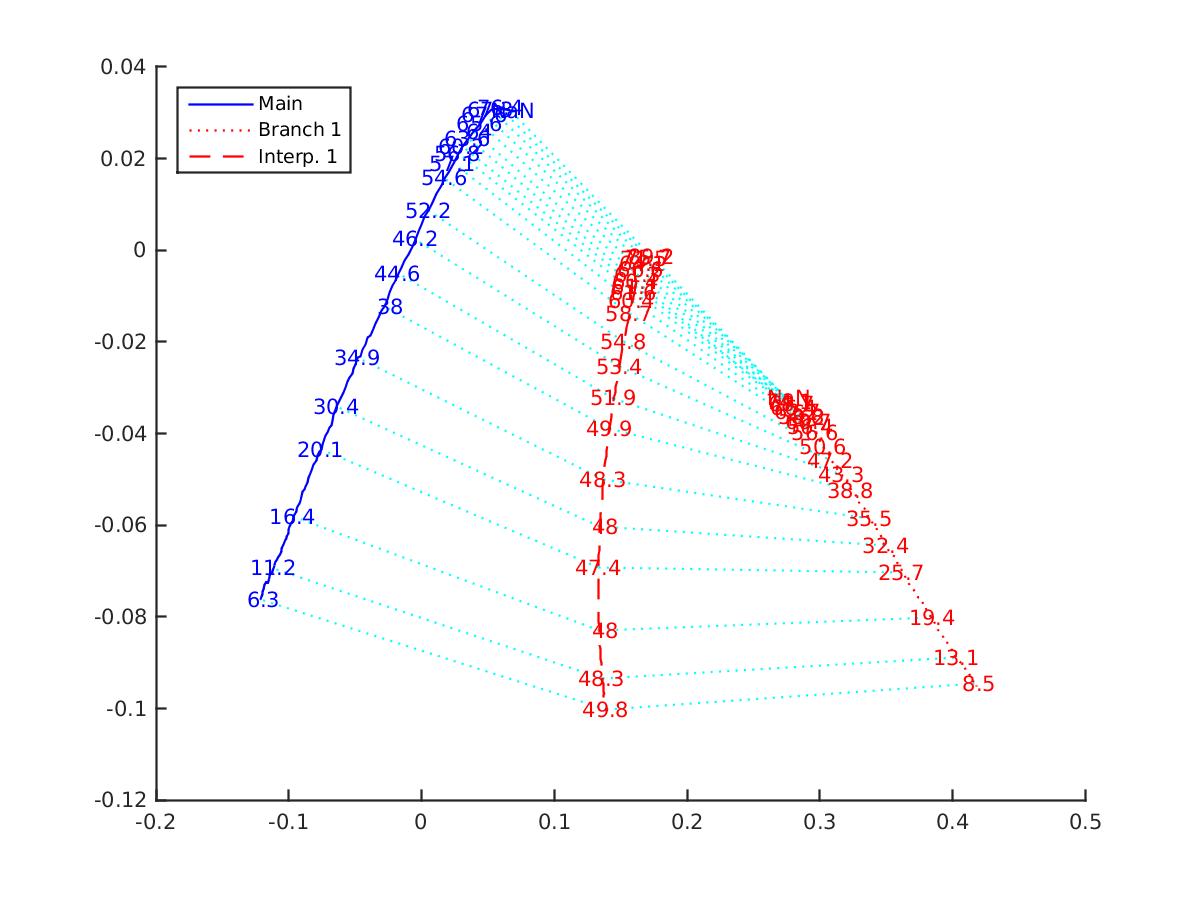}}   & \subfloat[]{\includegraphics[width=0.5\textwidth]{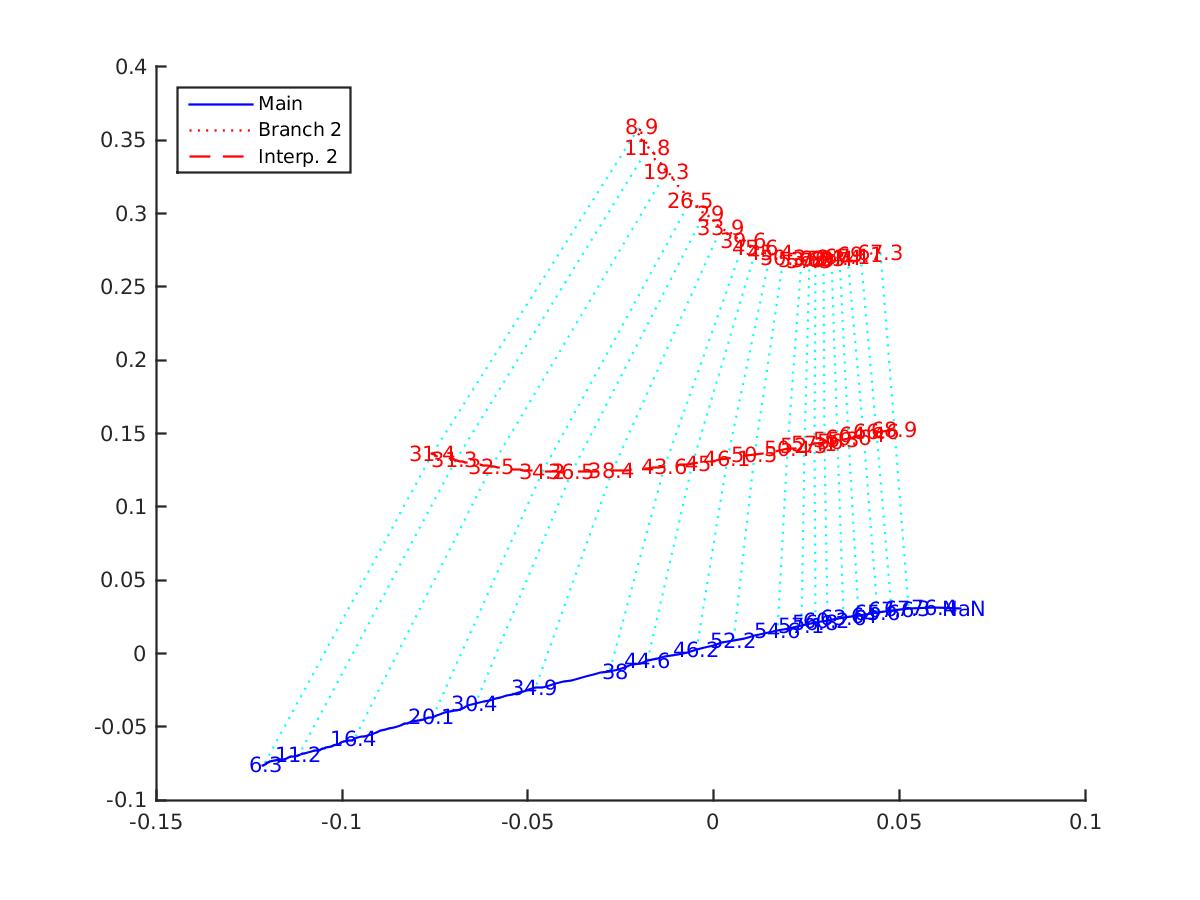}} \\  
  \subfloat[]{\includegraphics[width=0.5\textwidth]{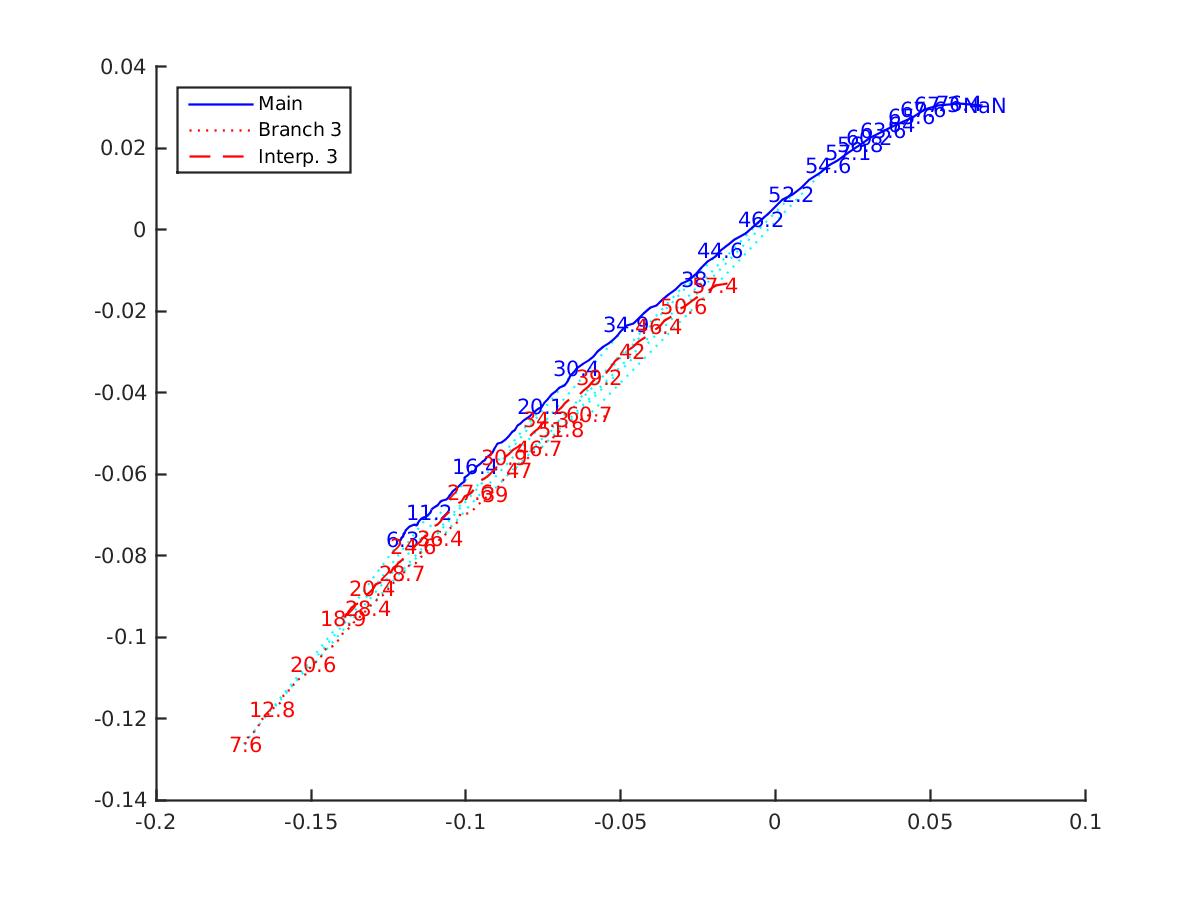}}   & \subfloat[]{\includegraphics[width=0.5\textwidth]{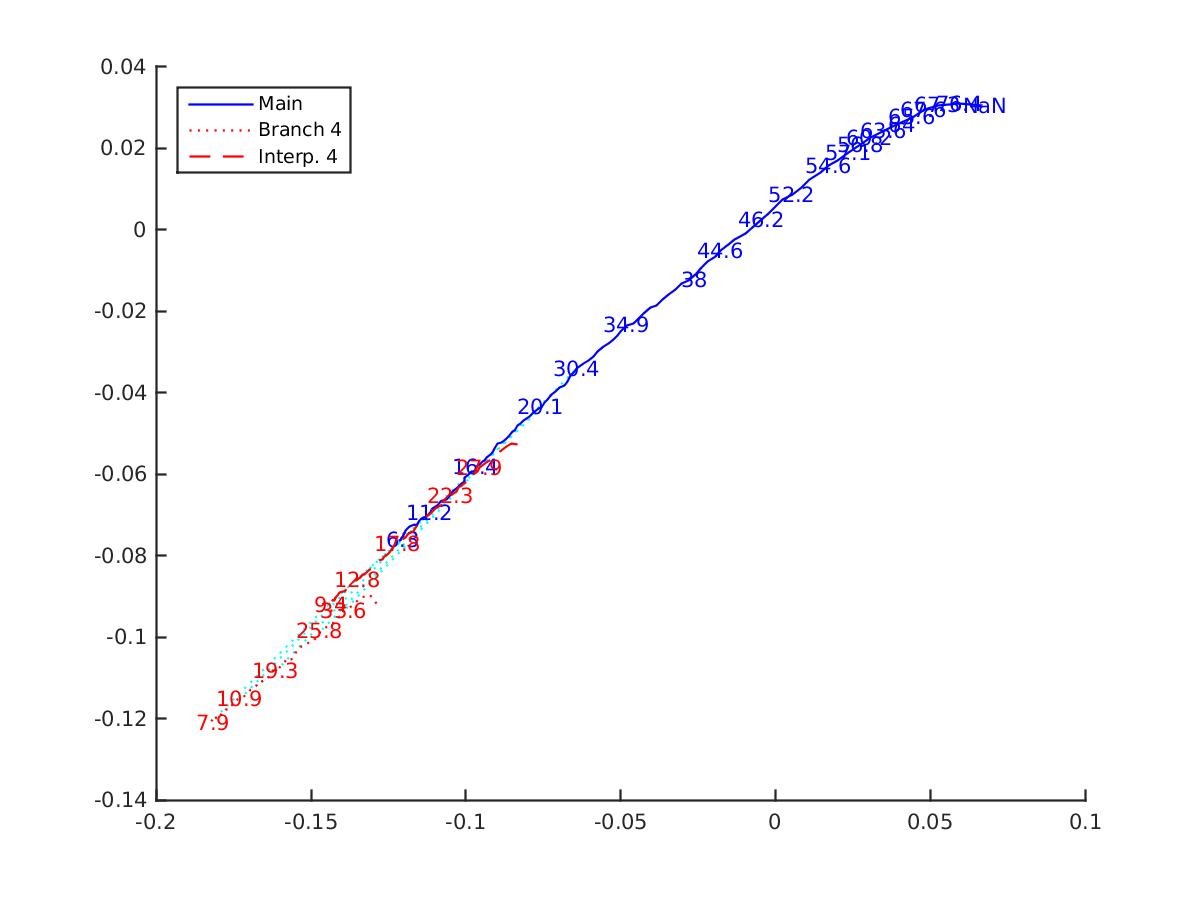}} \\
\end{tabular}    
\caption{Same as Figure \ref{appendix:fig:branch_layer_3_all_perturb_1}, but 4 branches are plotted separately to avoid clutter. Compared to Figure \ref{appendix:fig:branch_layer_2_sep_perturb_0.25}, here we have larger perturbations when creating a branch, which leads to more separated trajectories (e.g., subfigure (b)).   }\label{appendix:fig:branch_layer_3_sep_perturb_1}  
\end{figure}

\subsection{More Detailed Analyses of Several Local Landscapes (especially the flat global minima)} 
\label{appendix:vis:sec3} 

After the global visualization of the loss surface, we perform some
more detailed analyses at several locations of the landscape.
Especially, we would like to check if the global minima is flat. We
train a 6-layer (with the 1st layer being the input) DCNN on CIFAR-10
with 60 epochs of SGD (batch size = 100) and 400 epochs of Batch
Gradient Descent (BGD). BGD is performed to get to as close to the
global minima as possible. 


Next we select three models from this learning trajectory
\begin{itemize}
  \item $M_5$: the model at SGD epoch 5.
  \item $M_{30}$: the model at SGD epoch 30.
  \item $M_{final}$: the final model after 60 epochs of SGD and 400 epochs of BGD.
\end{itemize}

We perturb the weights of these models and retrain them with BGD, respectively. This
procedure was done multiple times for each model to get an idea of the nearby empirical risk landscape. 

The results are consistent with the previous theoretical arguments:

\begin{itemize}
  \item global minima are easily found with zero classification error and negligible cross entropy loss.
  \item The global minima seem ``flat'' under perturbations with zero error corresponding to different parameters values.
  \item The local landscapes at different levels of error seem to be very similar. Perturbing a model always lead to a different convergence path, leading to a similar but distinct model. We tried smaller perturbations and also observed this effect.
\end{itemize}

\subsubsection{Perturbing the model at SGD Epoch 5}

\begin{figure*}[!h]
\centering
\includegraphics[width=0.8\textwidth]{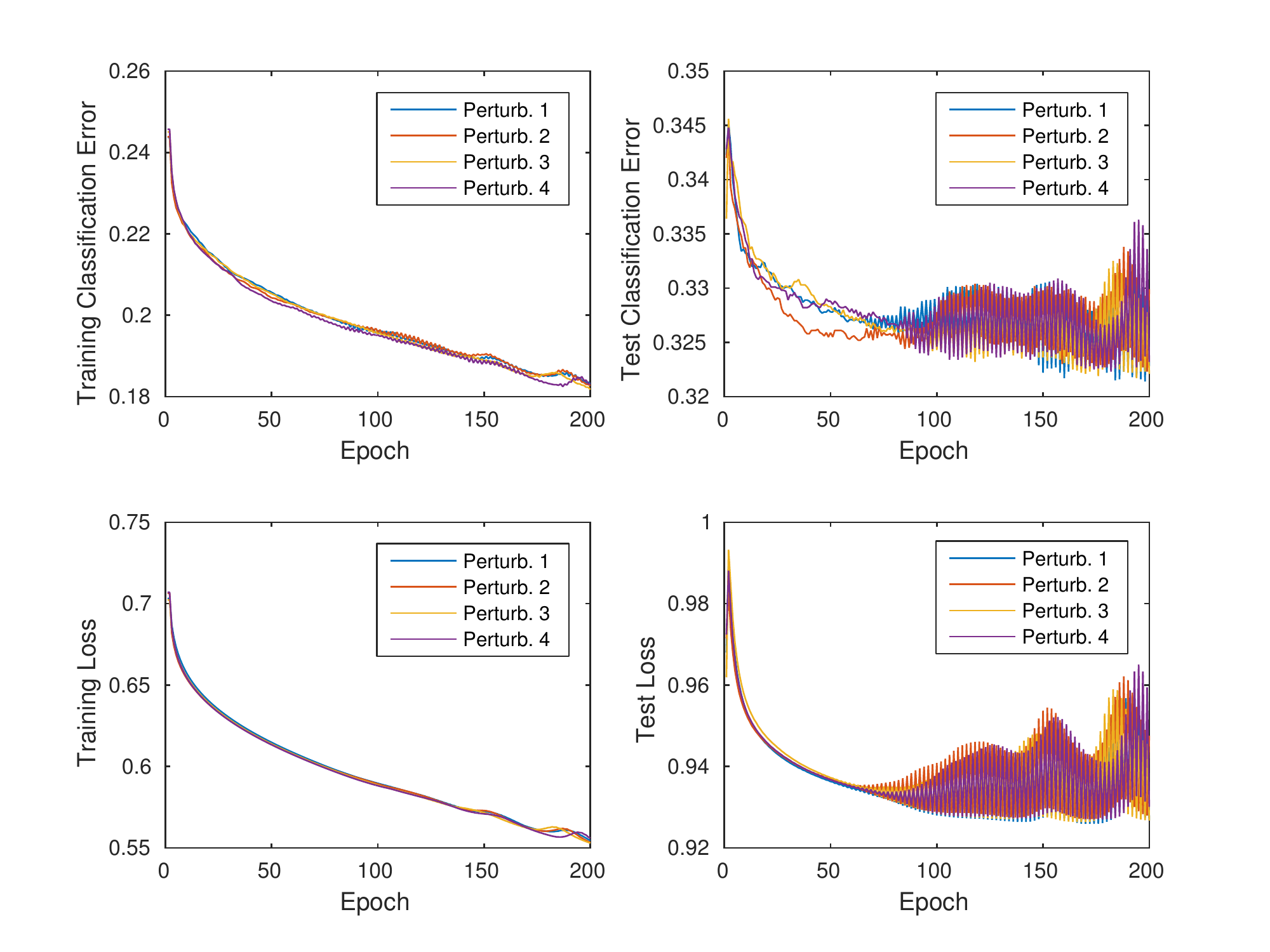} 
\caption{We layerwise perturb the weights of model $M_5$ by adding a gaussian noise with standard deviation = 0.1 * S, where S is the standard deviation of the weights. After perturbation, we continue training the model with 200 epochs of gradient descent (i.e., batch size = training set size). The same procedure was performed 4 times, resulting in 4 curves shown in the figures. The training and test classification errors and losses are shown. }
\label{appendix:fig:perturb_err_loss_from_epoch_5}
\end{figure*}

\begin{figure*}[!h]
\centering
\includegraphics[width=0.7\textwidth]{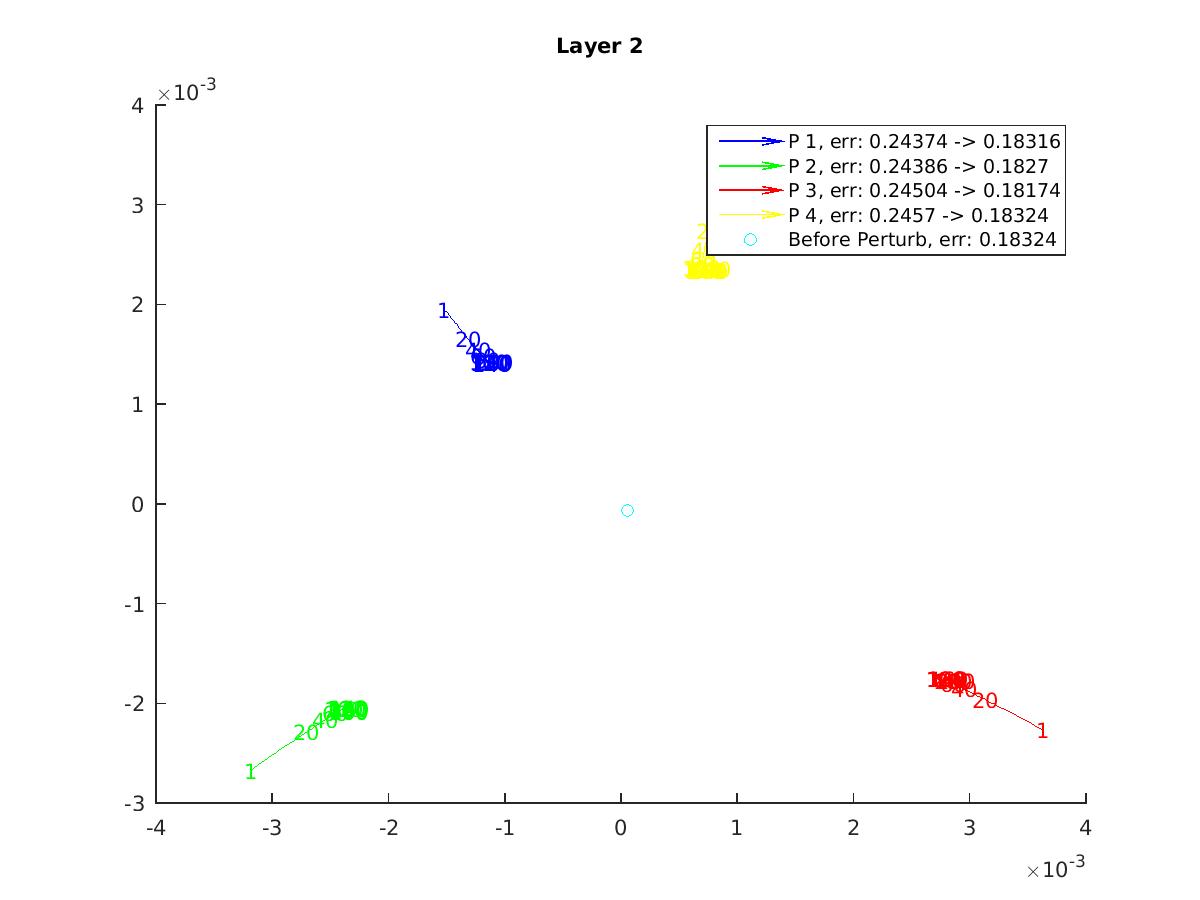}
\caption{Multidimensional scaling of the layer 2 weights throughout the retraining process. There are 4 runs indicated by 4 colors/trajectories, corresponding exactly to Figure \ref{appendix:fig:perturb_err_loss}. Each run corresponds to one perturbation of the model $M_{5}$ and subsequent 200 epochs' training. The number in the figure indicates the training epoch number after the initial perturbation.}
\label{appendix:fig:perturb_layer_2A_from_epoch_5}
\end{figure*}
\begin{figure*}[!h]
\centering
\includegraphics[width=0.7\textwidth]{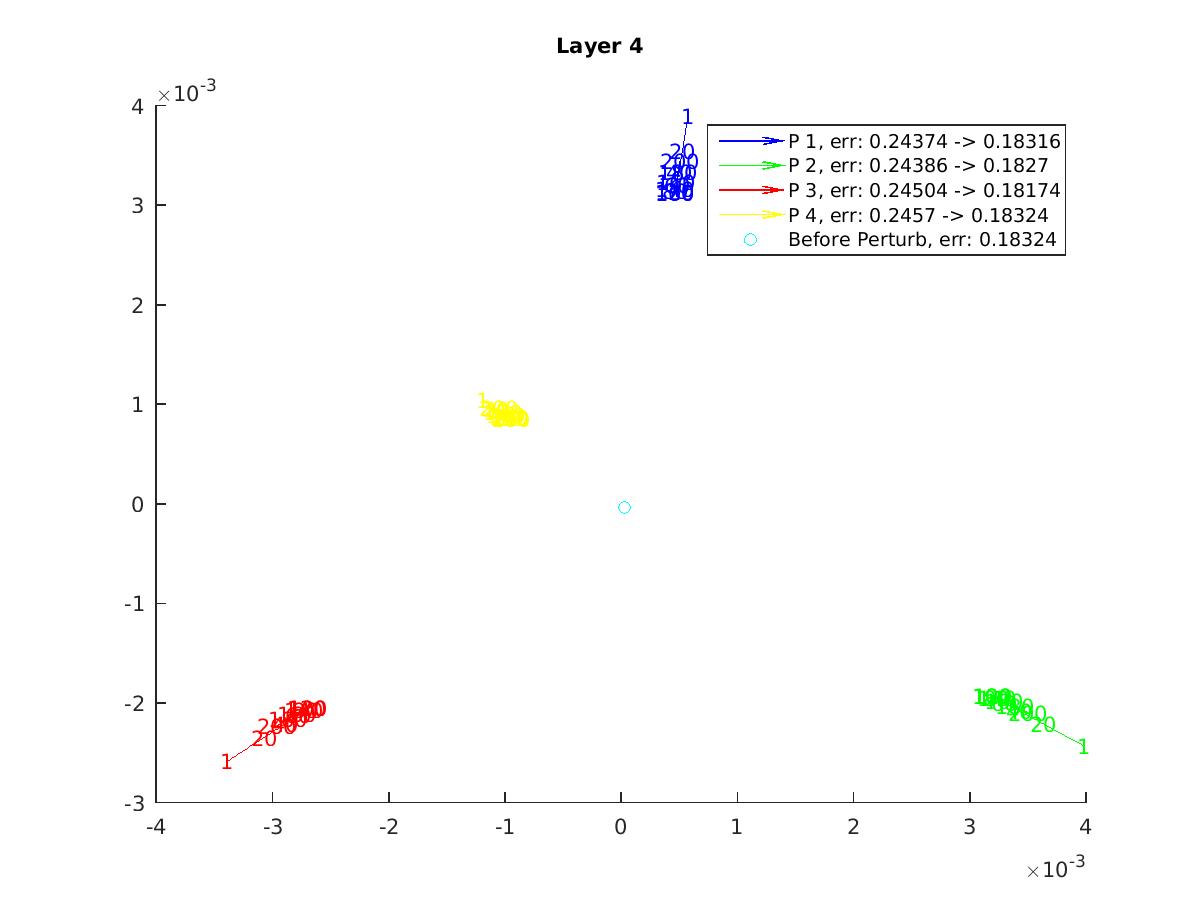}
\caption{Multidimensional scaling of the layer 4 weights throughout the training process. There are 4 runs indicated by 4 colors/trajectories, corresponding exactly to Figure \ref{appendix:fig:perturb_err_loss}. Each run corresponds to one perturbation of the model $M_{5}$ and subsequent 200 epochs' training. The number in the figure indicates the training epoch number after the initial perturbation.}
\label{appendix:fig:perturb_layer_4A_from_epoch_5}
\end{figure*}
\begin{figure*}[!h]
\centering
\includegraphics[width=0.7\textwidth]{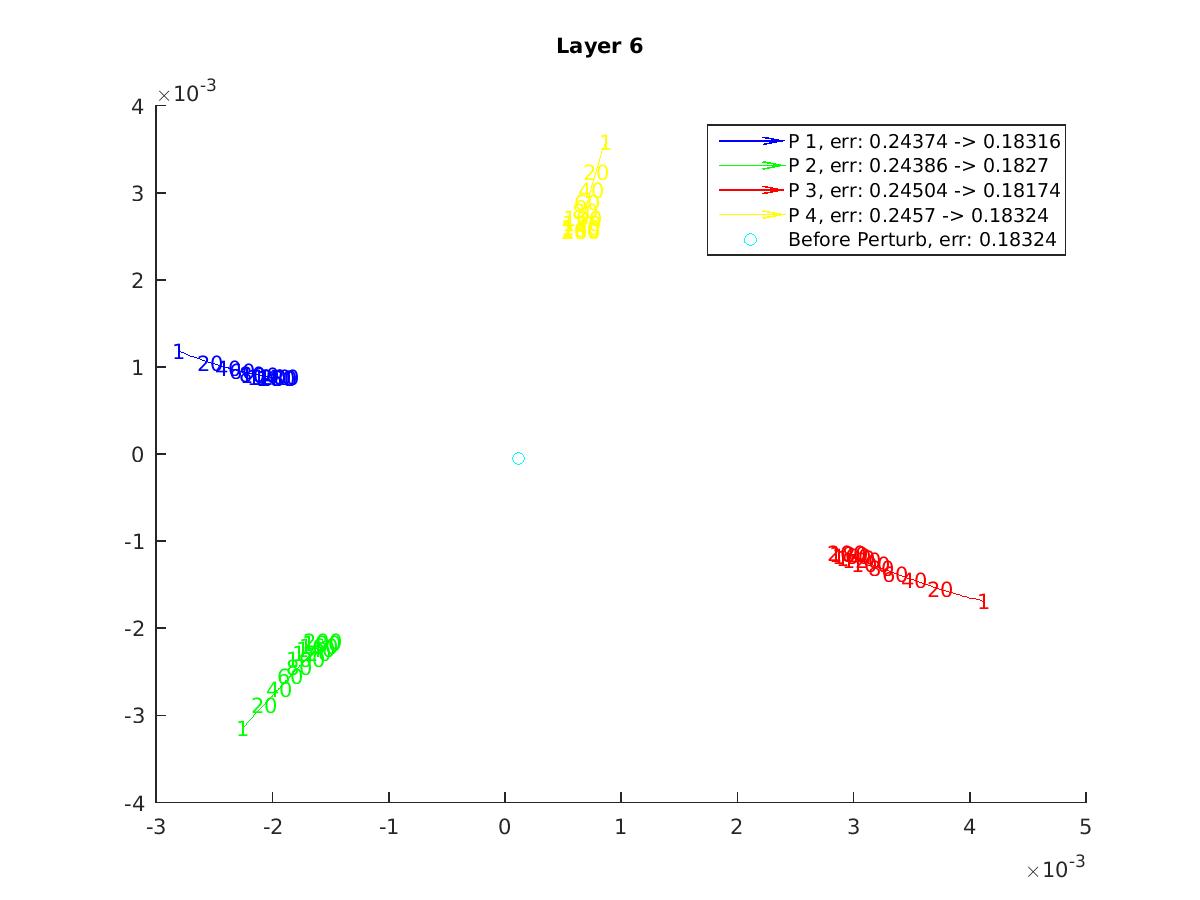}
\caption{Multidimensional scaling of the layer 6 weights throughout the training process. There are 4 runs indicated by 4 colors/trajectories, corresponding exactly to Figure \ref{appendix:fig:perturb_err_loss}. Each run corresponds to one perturbation of the model $M_{5}$ and subsequent 200 epochs' training. The number in the figure indicates the training epoch number after the initial perturbation.}  
\label{appendix:fig:perturb_layer_6A_from_epoch_5}
\end{figure*}

\begin{figure}  
\begin{tabular}{ccc}
  \subfloat[]{\includegraphics[width=0.31\textwidth]{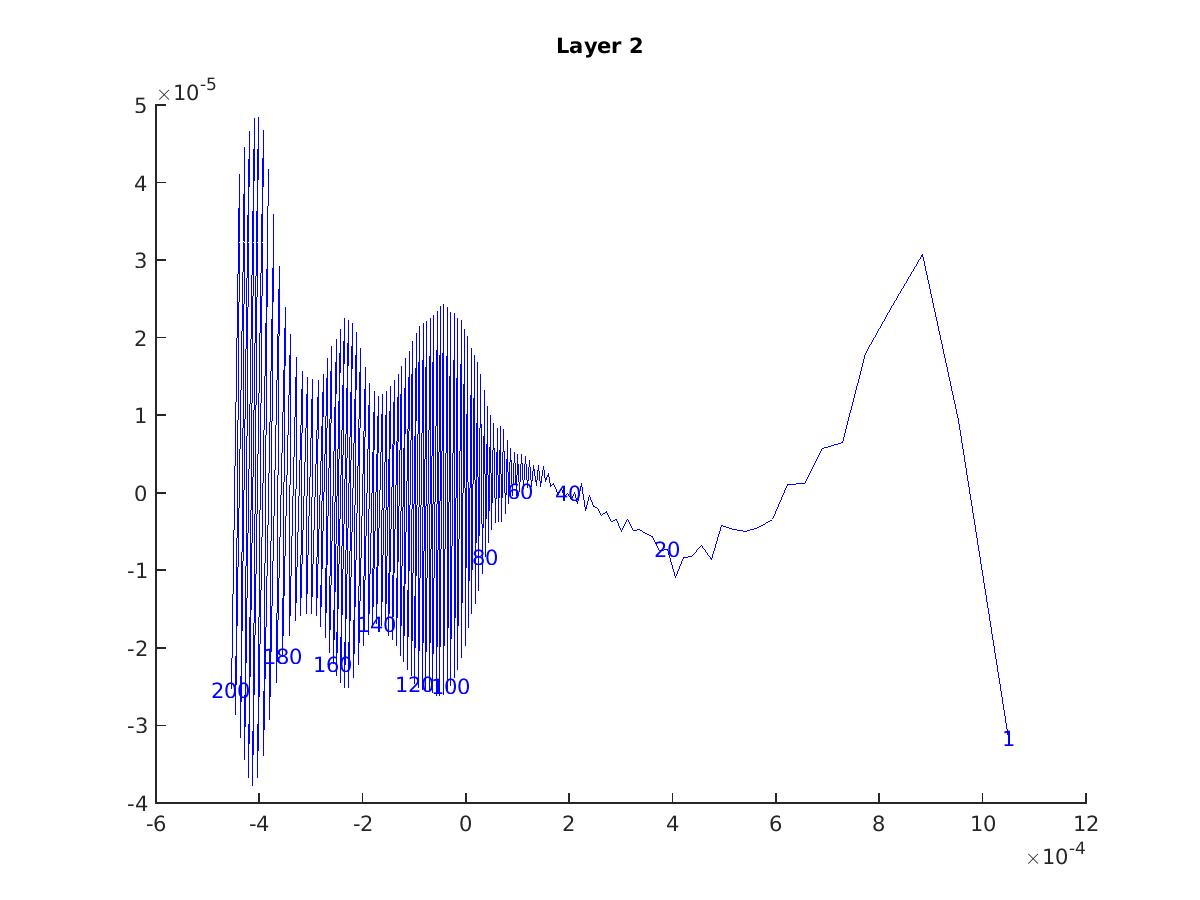}}   & \subfloat[]{\includegraphics[width=0.31\textwidth]{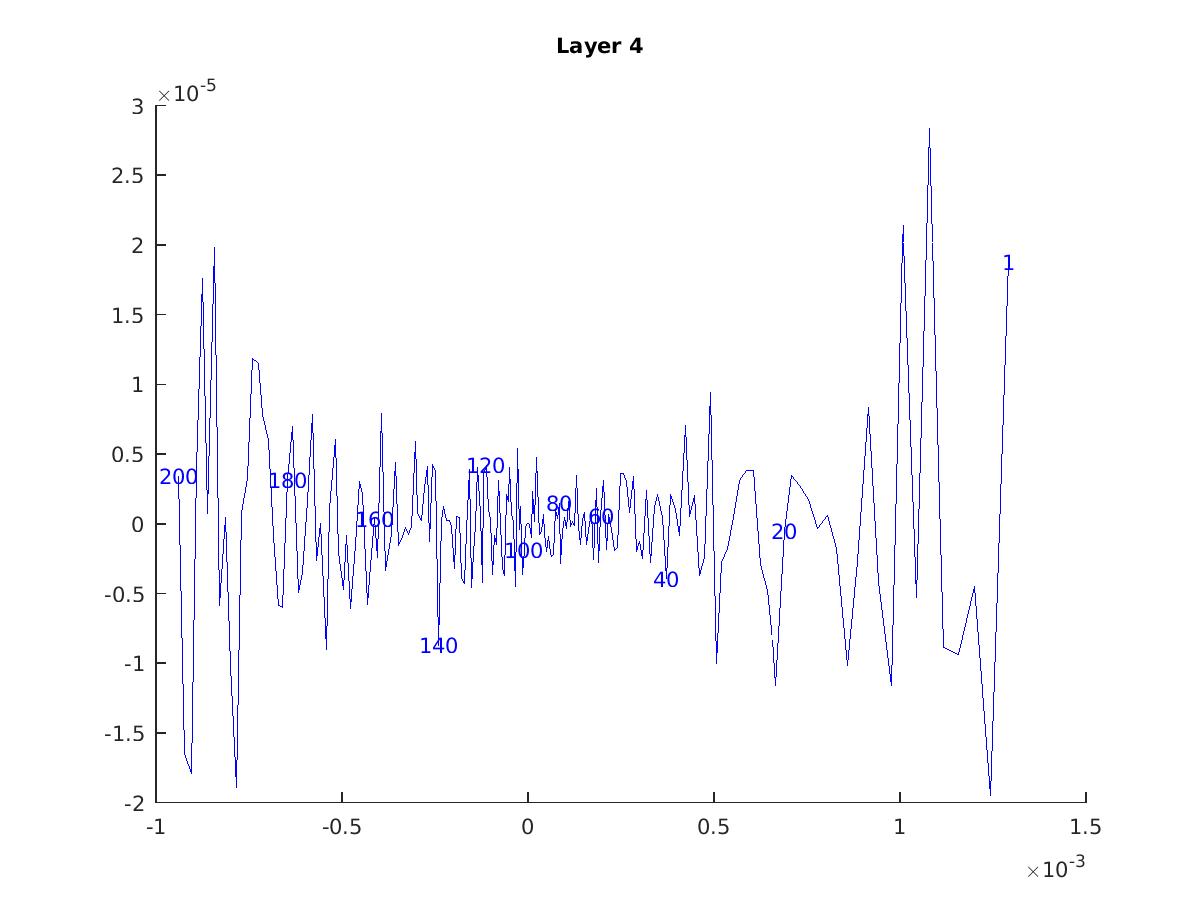}}   & \subfloat[]{\includegraphics[width=0.31\textwidth]{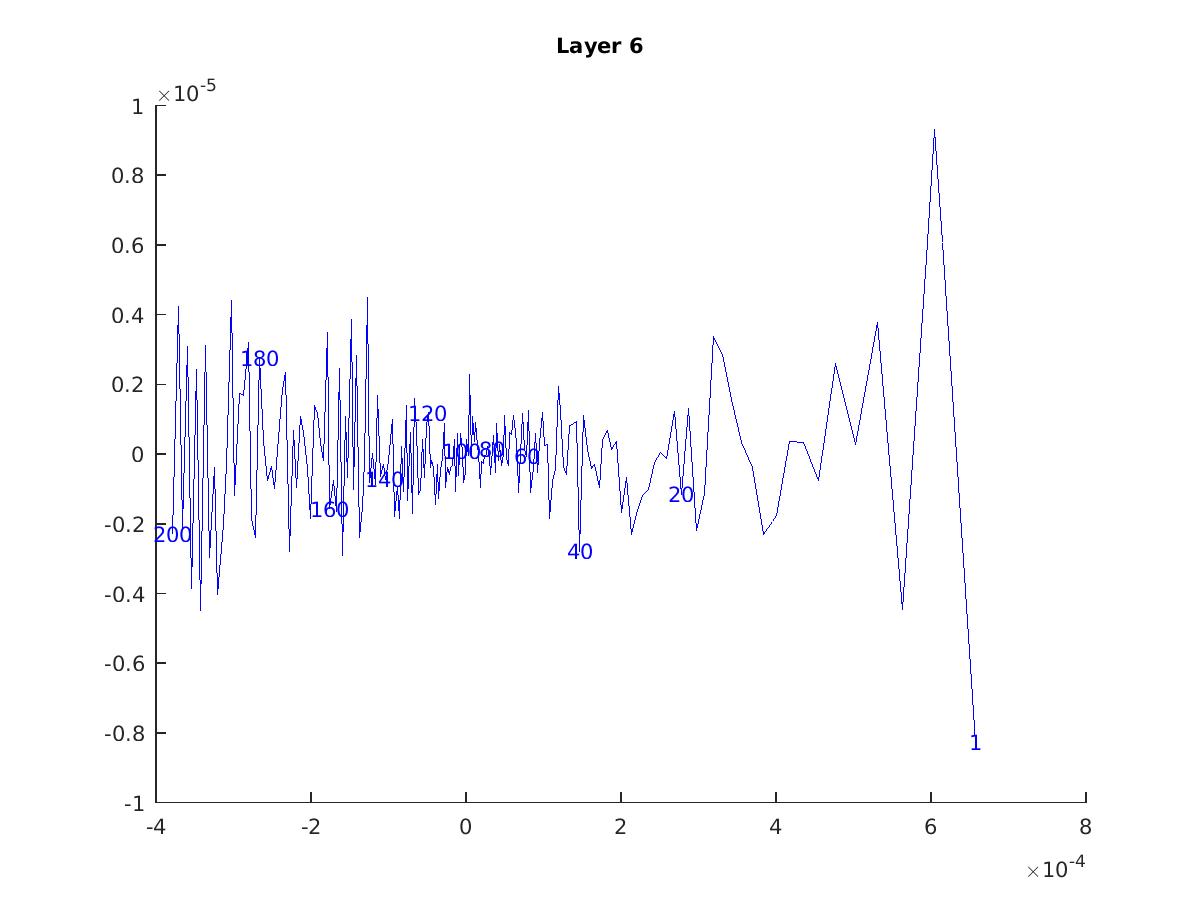}} \\ 
\end{tabular}
\caption{The Multidimensional Scaling (MDS) of the layer 2/4/6 weights of model  $M_{5}$ throughout the retraining process. Only the weights of one run (run 1 in Figure \ref{appendix:fig:perturb_err_loss}) are fed into MDS to provide more resolution. The number in the figure indicates the training epoch number after the initial perturbation.}  
\end{figure}

\begin{figure*}[!h]
\centering
\includegraphics[width=0.7\textwidth]{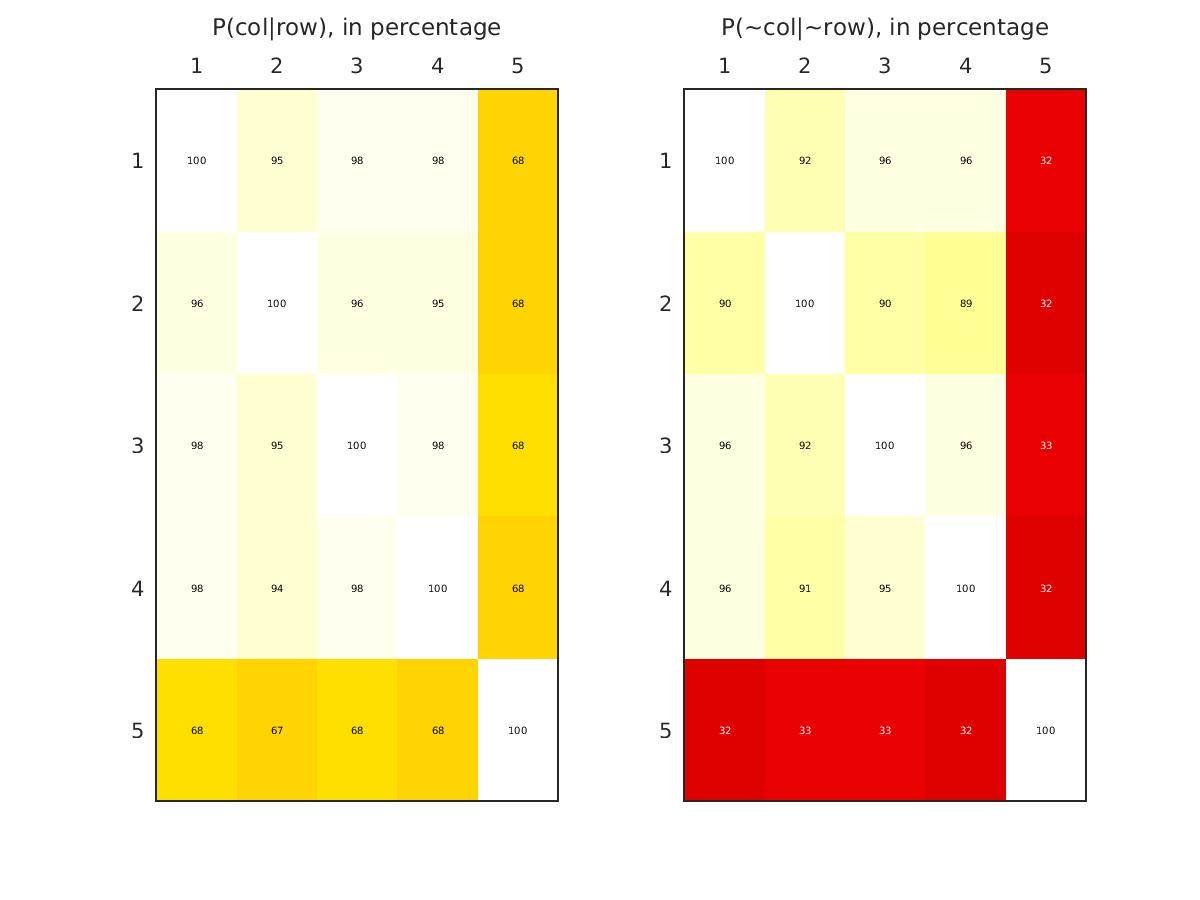}
\caption{The similarity between the final models of the 4 perturbations (row/column 1-4) of $M_{5}$ and a random reference model (the 5-th row/column). The left figure shows the probability of the column model being correct given the row model is correct.  The right figure shows the probability of the column model being incorrect given the row model is incorrect. These two figures show that the perturbed models really become different (although a little similar) models after continued training.} 
\label{appendix:fig:similarity_from_epoch_5} 
\end{figure*}

\subsubsection{Perturbing the model at SGD Epoch 30}

\begin{figure*}[!h]
\centering
\includegraphics[width=0.8\textwidth]{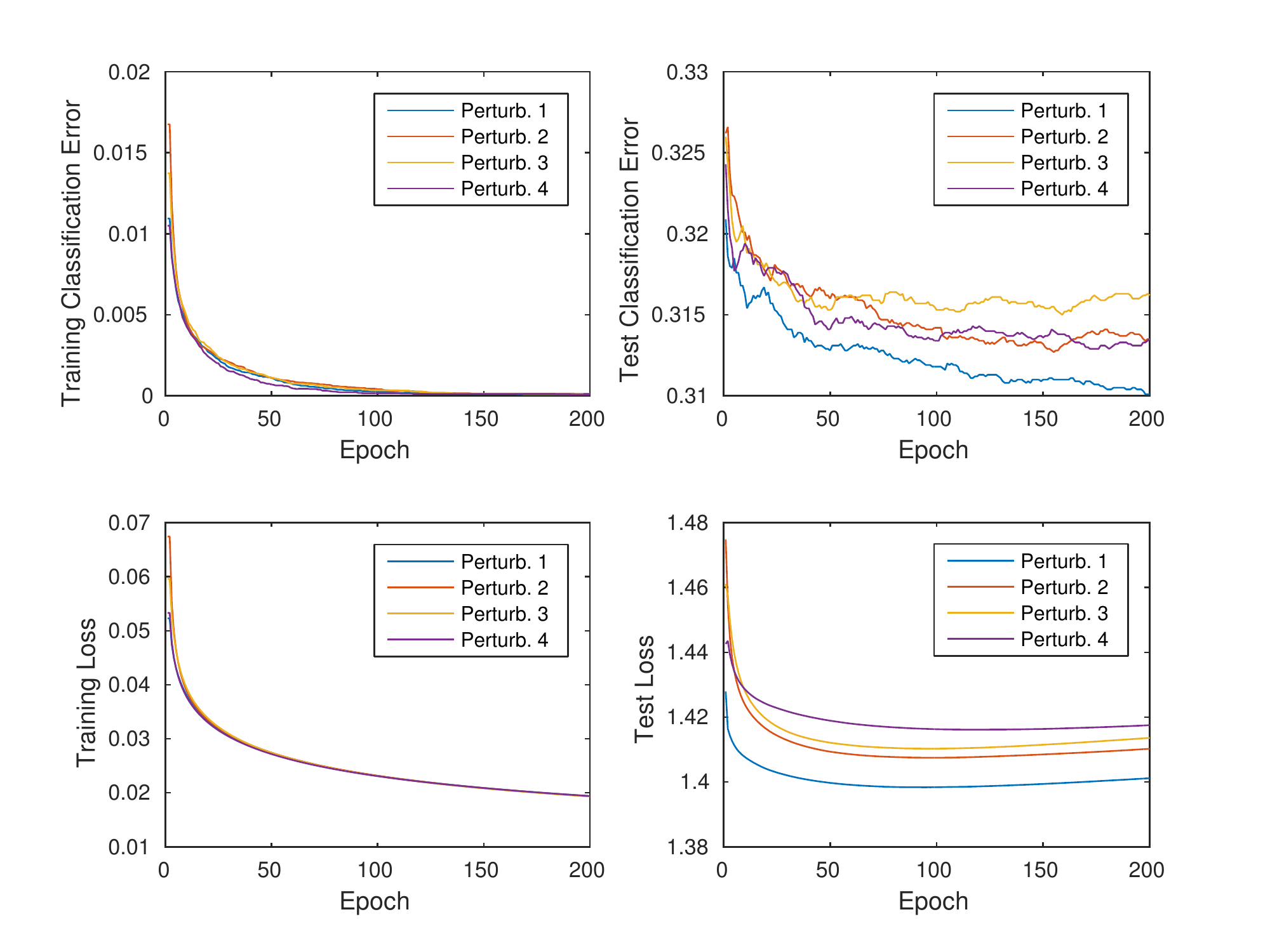} 
\caption{We layerwise perturb the weights of model $M_{30}$ by adding a gaussian noise with standard deviation = 0.1 * S, where S is the standard deviation of the weights. After perturbation, we continue training the model with 200 epochs of gradient descent (i.e., batch size = training set size). The same procedure was performed 4 times, resulting in 4 curves shown in the figures. The training and test classification errors and losses are shown. } 
\label{appendix:fig:perturb_err_loss_from_epoch_30}
\end{figure*}

\begin{figure*}[!h]
\centering
\includegraphics[width=0.7\textwidth]{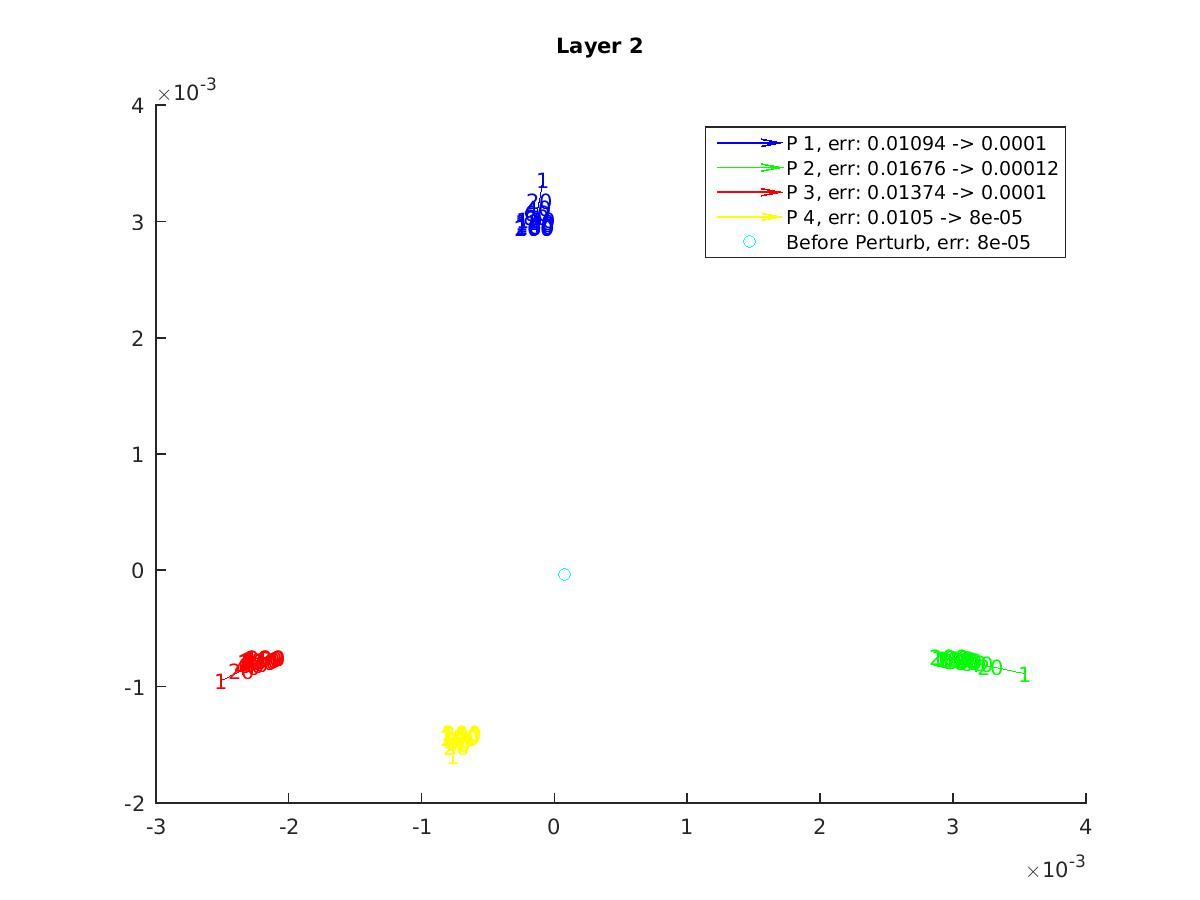}
\caption{Multidimensional scaling of the layer 2 weights throughout the retraining process. There are 4 runs indicated by 4 colors/trajectories, corresponding exactly to Figure \ref{appendix:fig:perturb_err_loss}. Each run corresponds to one perturbation of the model $M_{30}$ and subsequent 200 epochs' training. The number in the figure indicates the training epoch number after the initial perturbation.} 
\label{appendix:fig:perturb_layer_2A_from_epoch_30}
\end{figure*}

\begin{figure*}[!h]
\centering
\includegraphics[width=0.7\textwidth]{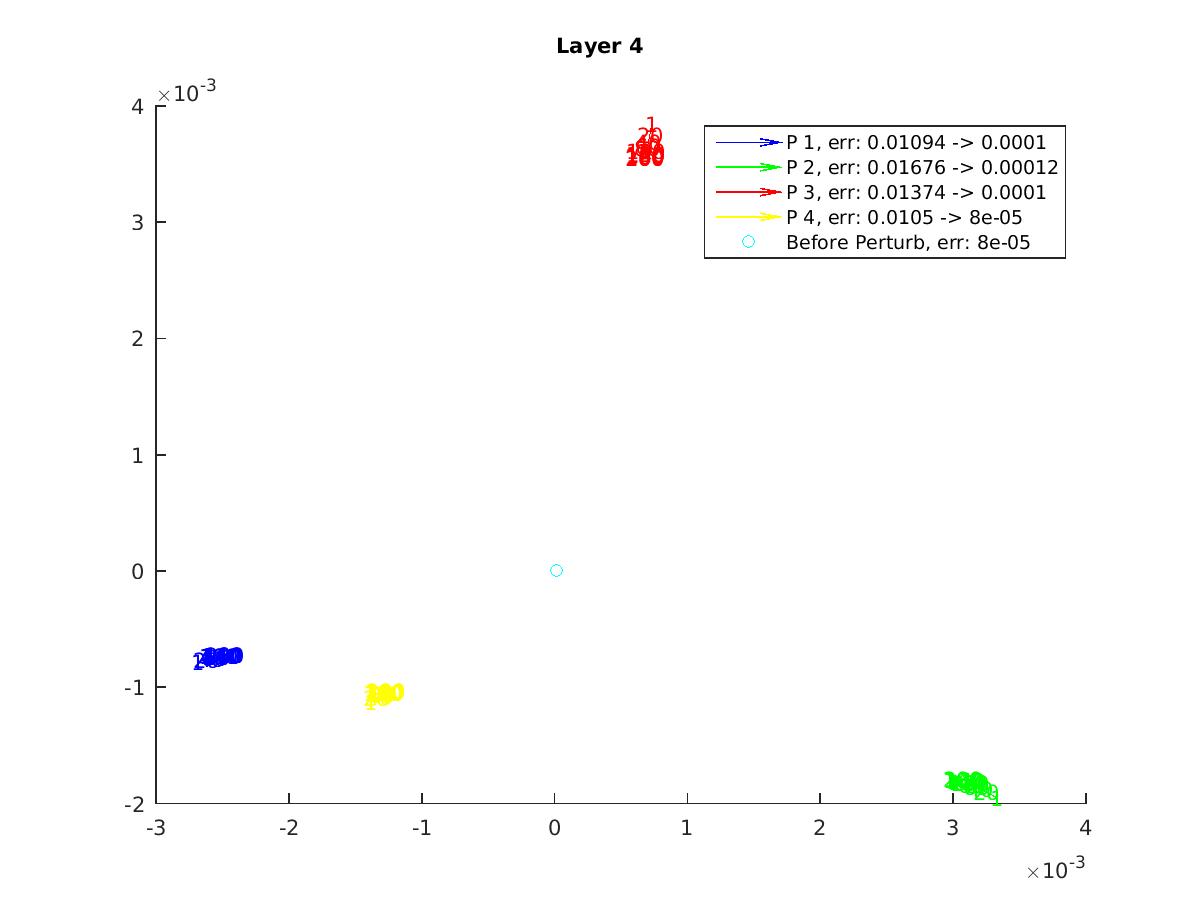}
\caption{Multidimensional scaling of the layer 4 weights throughout the training process. There are 4 runs indicated by 4 colors/trajectories, corresponding exactly to Figure \ref{appendix:fig:perturb_err_loss}. Each run corresponds to one perturbation of the model $M_{30}$ and subsequent 200 epochs' training. The number in the figure indicates the training epoch number after the initial perturbation.}
\label{appendix:fig:perturb_layer_4A_from_epoch_30}
\end{figure*}
 
\begin{figure*}[!h]
\centering
\includegraphics[width=0.7\textwidth]{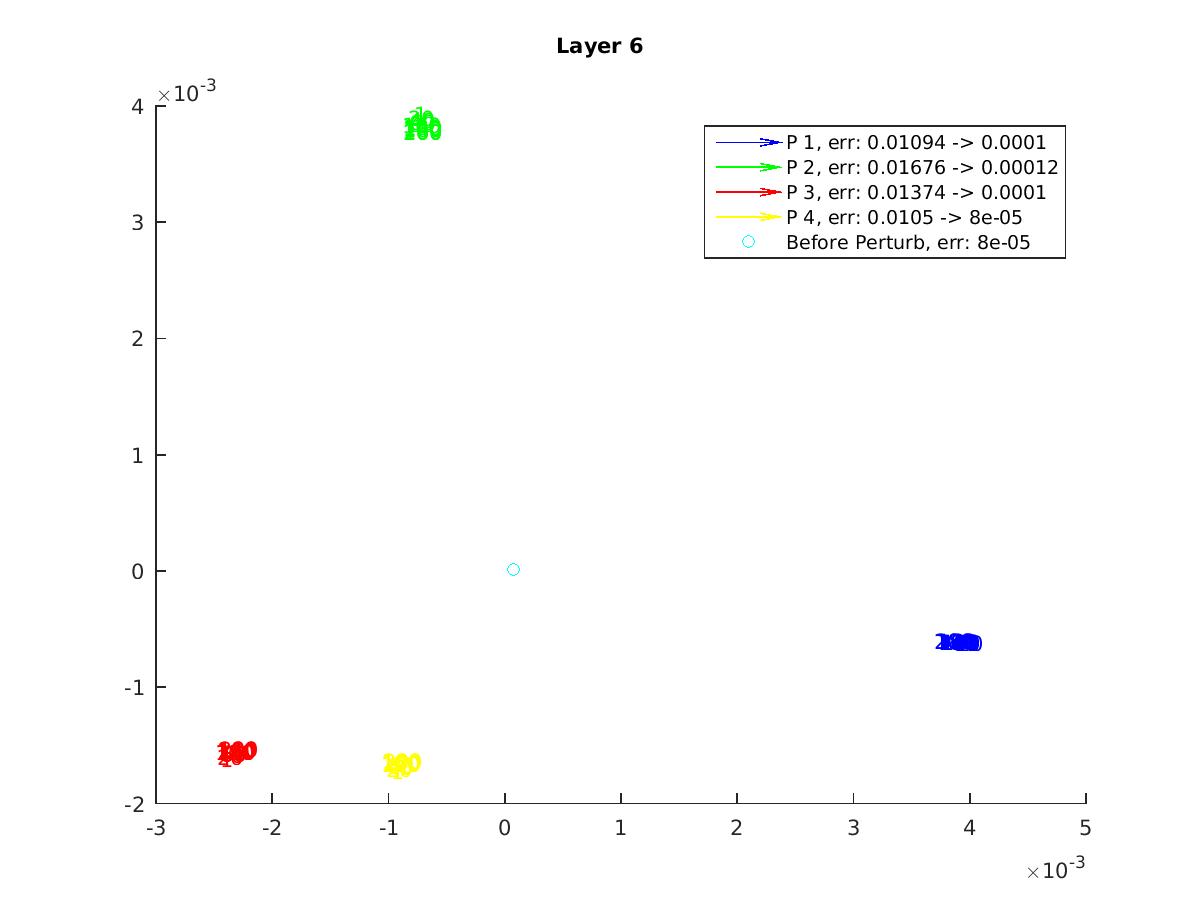}
\caption{Multidimensional scaling of the layer 6 weights throughout the training process. There are 4 runs indicated by 4 colors/trajectories, corresponding exactly to Figure \ref{appendix:fig:perturb_err_loss}. Each run corresponds to one perturbation of the model $M_{30}$ and subsequent 200 epochs' training. The number in the figure indicates the training epoch number after the initial perturbation.}  
\label{appendix:fig:perturb_layer_6A_from_epoch_30}
\end{figure*}

\begin{figure}  
\begin{tabular}{ccc}
  \subfloat[]{\includegraphics[width=0.31\textwidth]{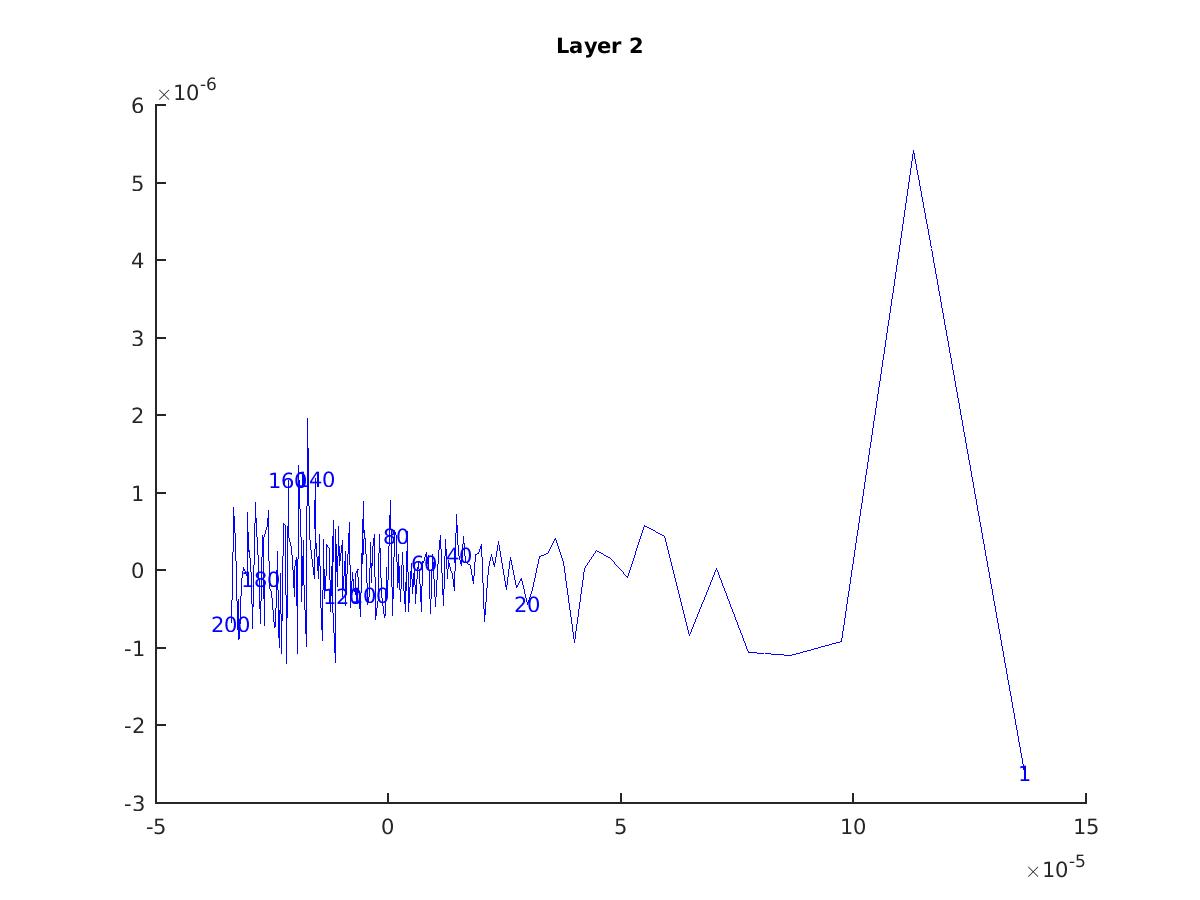}}   & \subfloat[]{\includegraphics[width=0.31\textwidth]{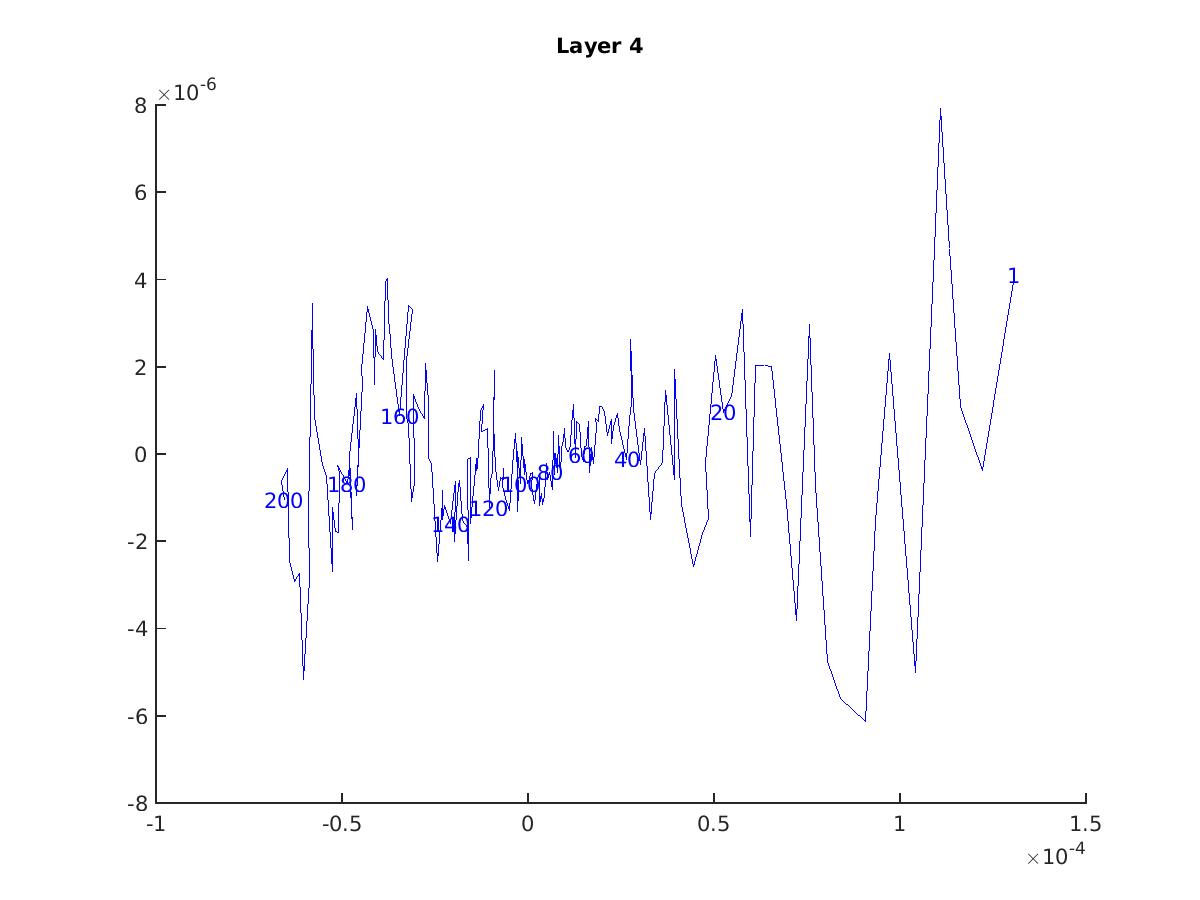}}   & \subfloat[]{\includegraphics[width=0.31\textwidth]{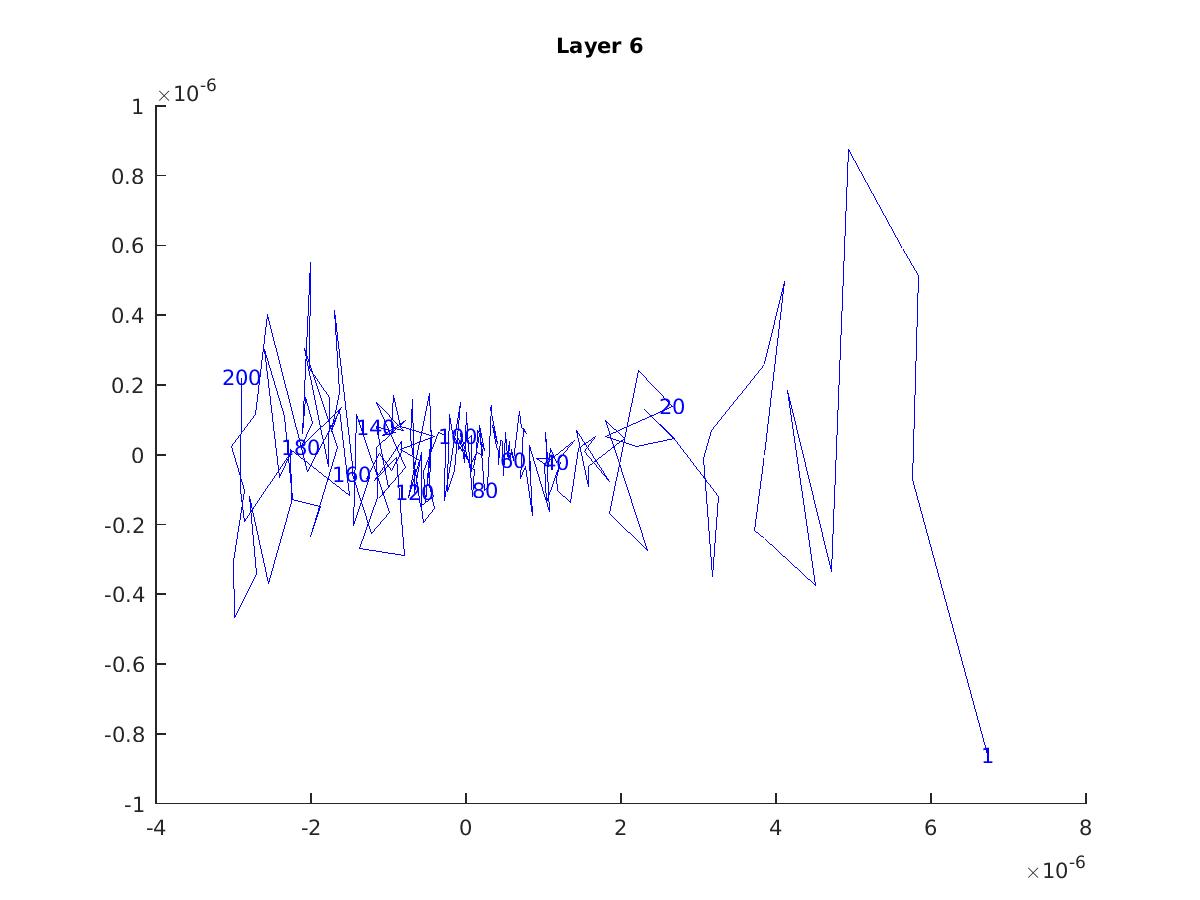}} \\ 
\end{tabular}
\caption{The Multidimensional Scaling (MDS) of the layer 2/4/6 weights of model  $M_{30}$ throughout the retraining process. Only the weights of one run (run 1 in Figure \ref{appendix:fig:perturb_err_loss}) are fed into MDS to provide more resolution. The number in the figure indicates the training epoch number after the initial perturbation.}   
\end{figure}

\begin{figure*}[!h]
\centering
\includegraphics[width=0.7\textwidth]{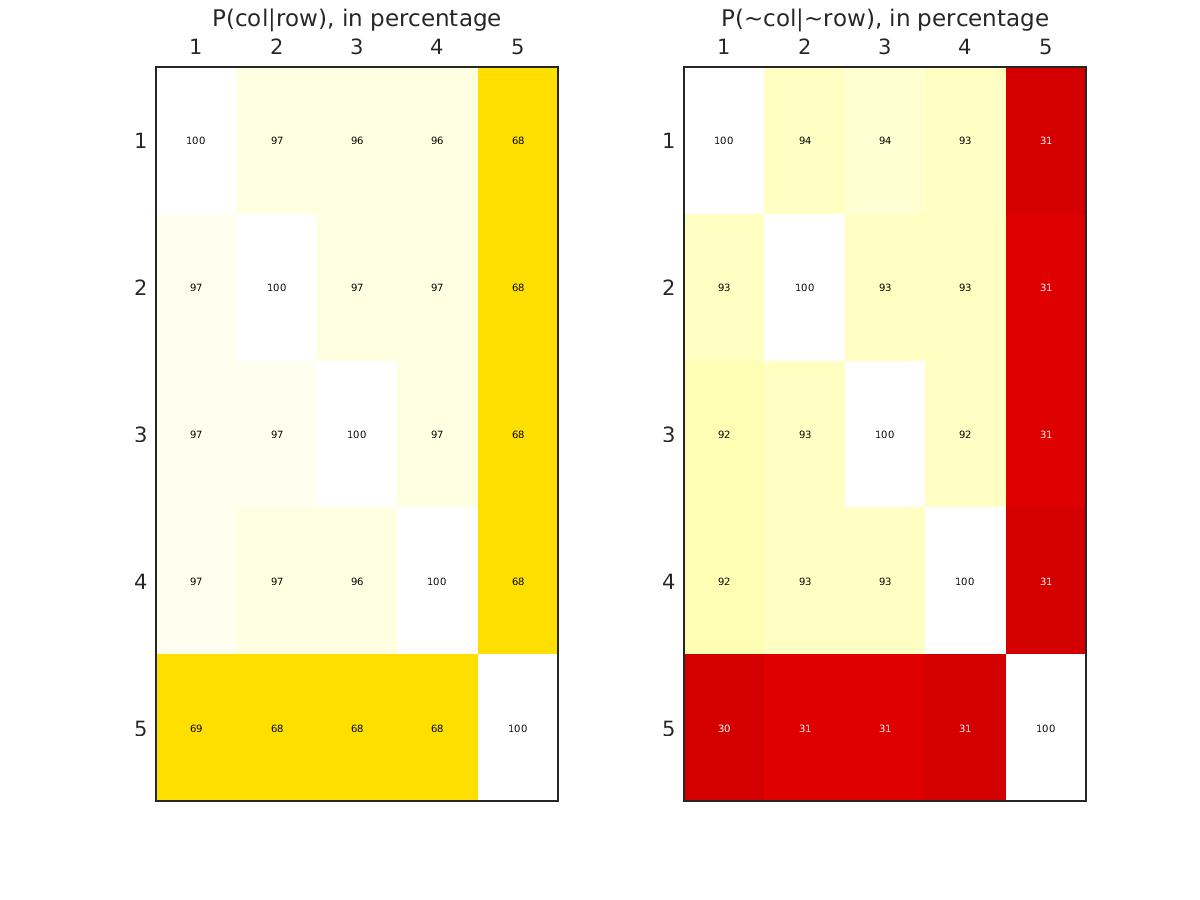}
\caption{The similarity between the final models of the 4 perturbations (row/column 1-4) of $M_{30}$ and a random reference model (the 5-th row/column). The left figure shows the probability of the column model being correct given the row model is correct.  The right figure shows the probability of the column model being incorrect given the row model is incorrect. These two figures show that the perturbed models really become different (although a little similar) models after continued training.} 
\label{appendix:fig:similarity_from_epoch_30} 
\end{figure*}

\begin{figure*}[!h]
\centering
\includegraphics[width=0.7\textwidth]{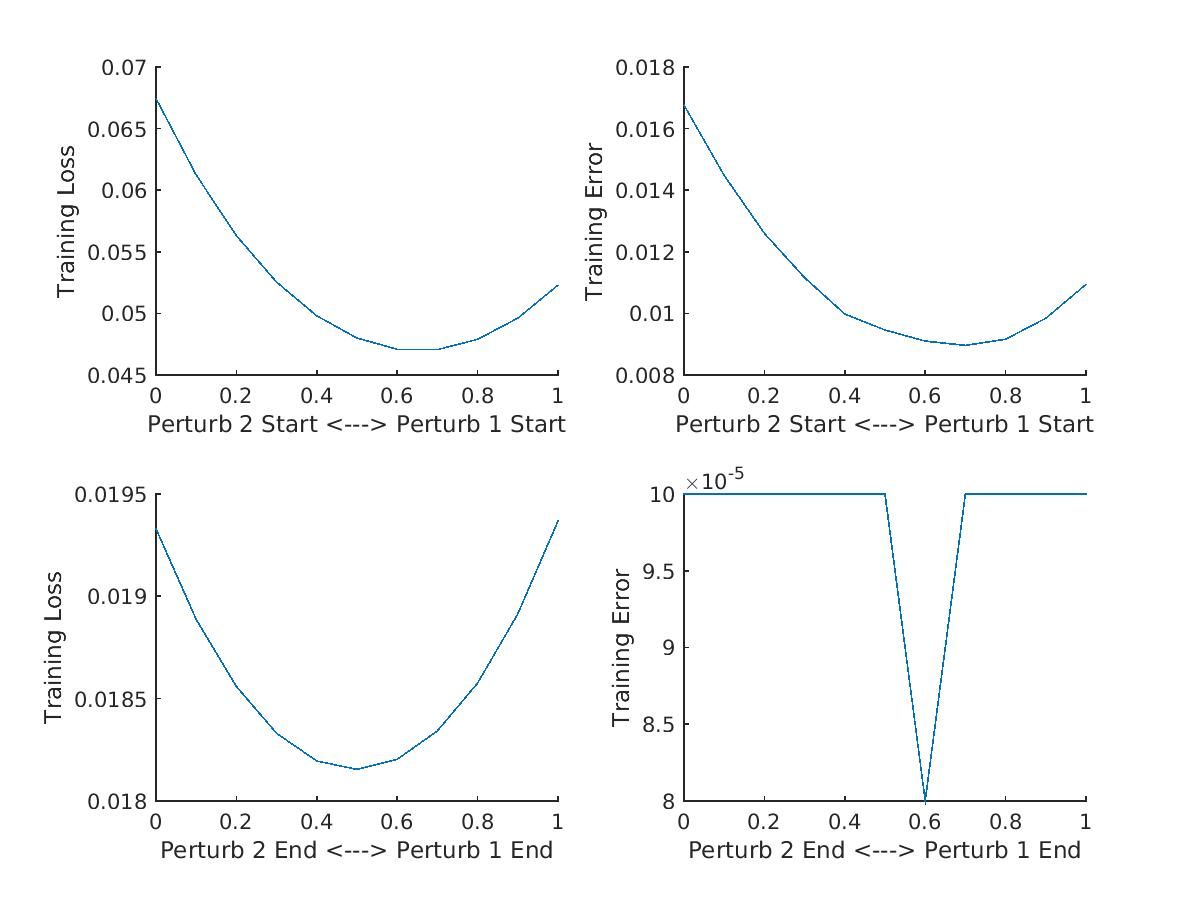}
\caption{Interpolating perturbation 1 and 2 of $M_{30}$. The top and bottom rows interpolate the perturbed models before and after retraining, respectively. The x axes are interpolating ratios --- at x=0 the model becomes perturbed model 1 and at x=1 the model becomes perturbed model 2. It is a little surprising to find that the interpolated models between perturbation 1 and 2 can even have slightly lower errors. Note that this is no longer the case if the perturbations are very large.}
\label{appendix:fig:interp_from_epoch_30} 
\end{figure*}

\subsubsection{Perturbing the final model (SGD 60 + GD 400 epochs)}

\begin{figure*}[!h]
\centering
\includegraphics[width=0.8\textwidth]{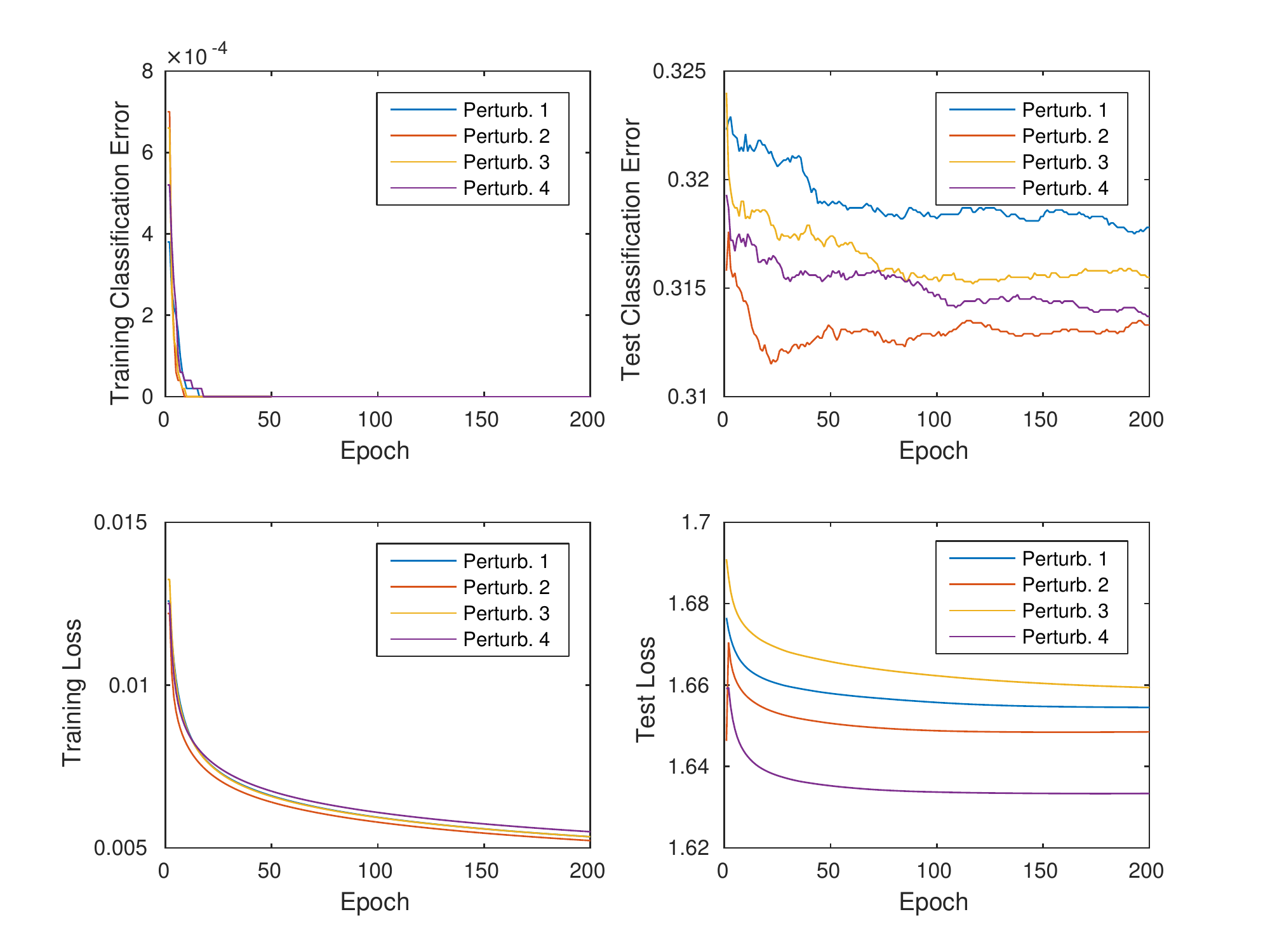}
\caption{We layerwise perturb the weights of model $M_{final}$ by adding a gaussian noise with standard deviation = 0.1 * S, where S is the standard deviation of the weights. After perturbation, we continue training the model with 200 epochs of gradient descent (i.e., batch size = training set size). The same procedure was performed 4 times, resulting in 4 curves shown in the figures. The training and test classification errors and losses are shown. }
\label{appendix:fig:perturb_err_loss_from_epoch_sgd60_plus_gd400}
\end{figure*}

\begin{figure*}[!h]
\centering
\includegraphics[width=0.7\textwidth]{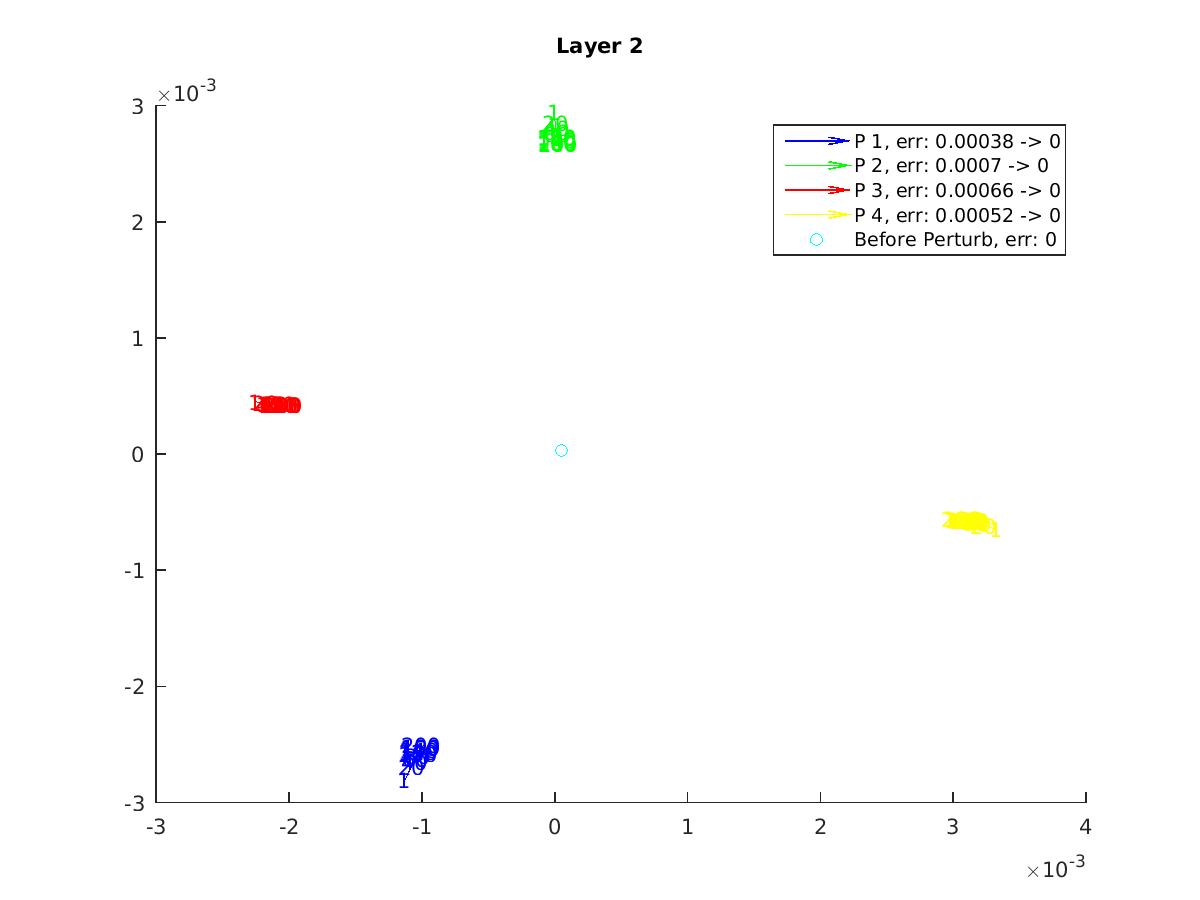}
\caption{Multidimensional scaling of the layer 2 weights throughout the retraining process. There are 4 runs indicated by 4 colors/trajectories, corresponding exactly to Figure \ref{appendix:fig:perturb_err_loss}. Each run corresponds to one perturbation of the model $M_{final}$ and subsequent 200 epochs' training. The number in the figure indicates the training epoch number after the initial perturbation.} 
\label{appendix:fig:perturb_layer_2A_from_epoch_sgd60_plus_gd400}
\end{figure*}

\begin{figure*}[!h]
\centering
\includegraphics[width=0.7\textwidth]{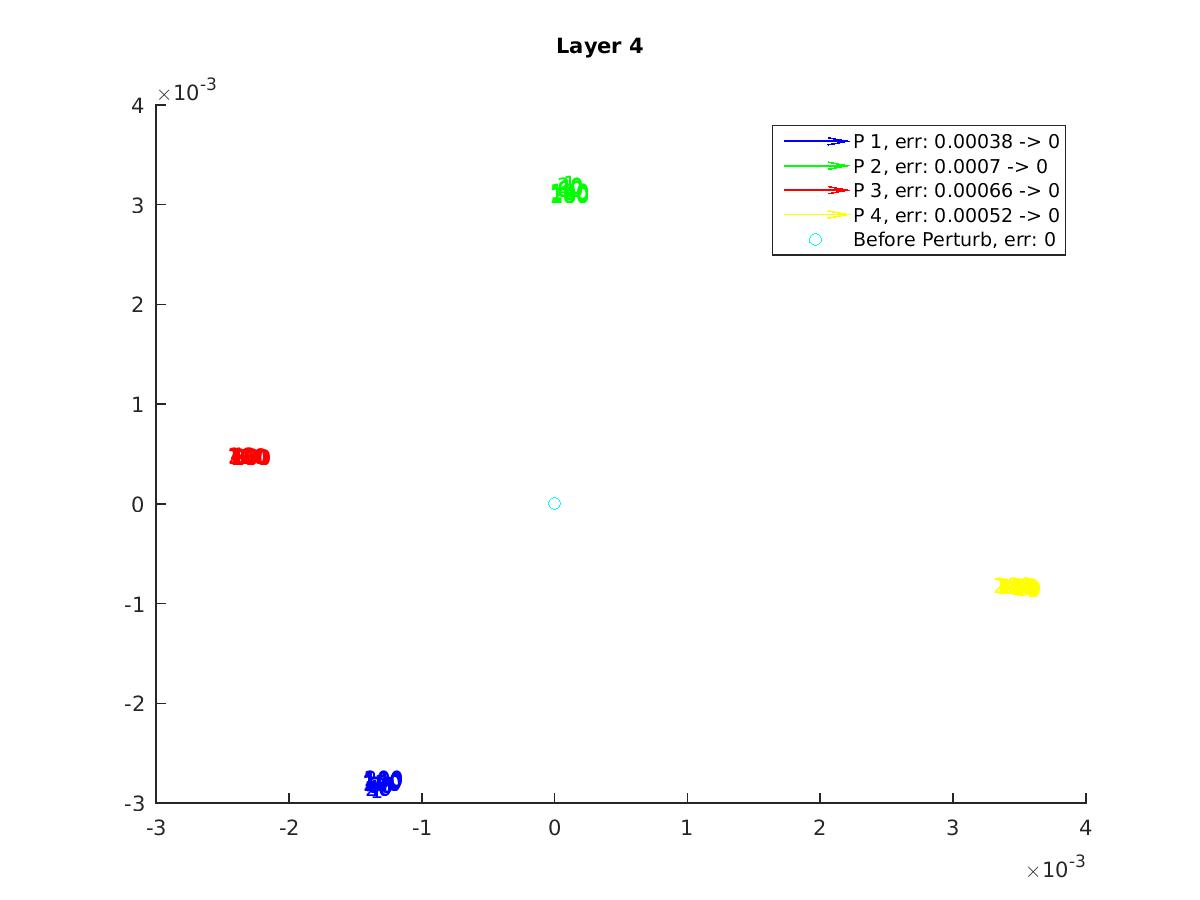}
\caption{Multidimensional scaling of the layer 4 weights throughout the training process. There are 4 runs indicated by 4 colors/trajectories, corresponding exactly to Figure \ref{appendix:fig:perturb_err_loss}. Each run corresponds to one perturbation of the model $M_{final}$ and subsequent 200 epochs' training. The number in the figure indicates the training epoch number after the initial perturbation.}
\label{appendix:fig:perturb_layer_4A_from_epoch_sgd60_plus_gd400}
\end{figure*}
 
\begin{figure*}[!h]
\centering
\includegraphics[width=0.7\textwidth]{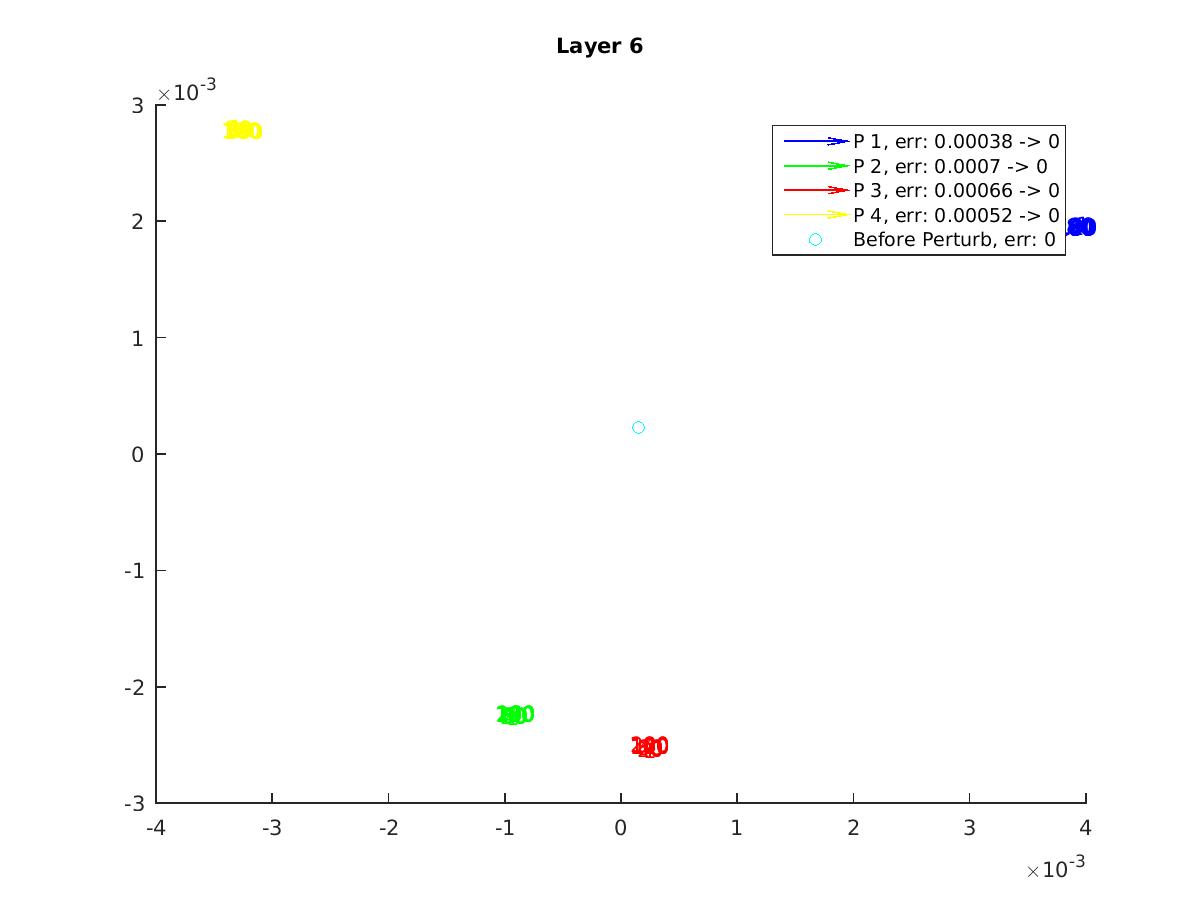}
\caption{Multidimensional scaling of the layer 6 weights throughout the training process. There are 4 runs indicated by 4 colors/trajectories, corresponding exactly to Figure \ref{appendix:fig:perturb_err_loss}. Each run corresponds to one perturbation of the model $M_{final}$ and subsequent 200 epochs' training. The number in the figure indicates the training epoch number after the initial perturbation.}  
\label{appendix:fig:perturb_layer_6A_from_epoch_sgd60_plus_gd400}
\end{figure*}

\begin{figure}  
\begin{tabular}{ccc}
  \subfloat[]{\includegraphics[width=0.31\textwidth]{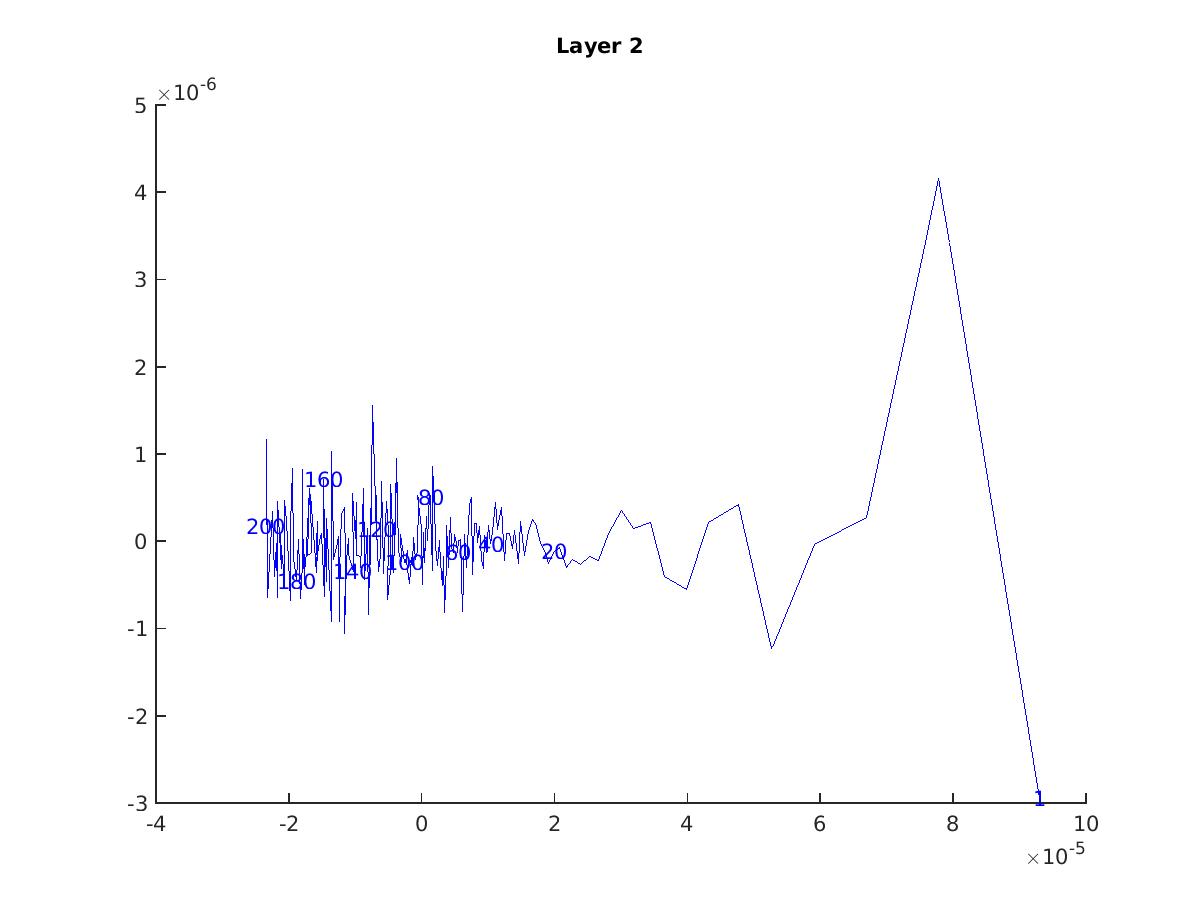}}   & \subfloat[]{\includegraphics[width=0.31\textwidth]{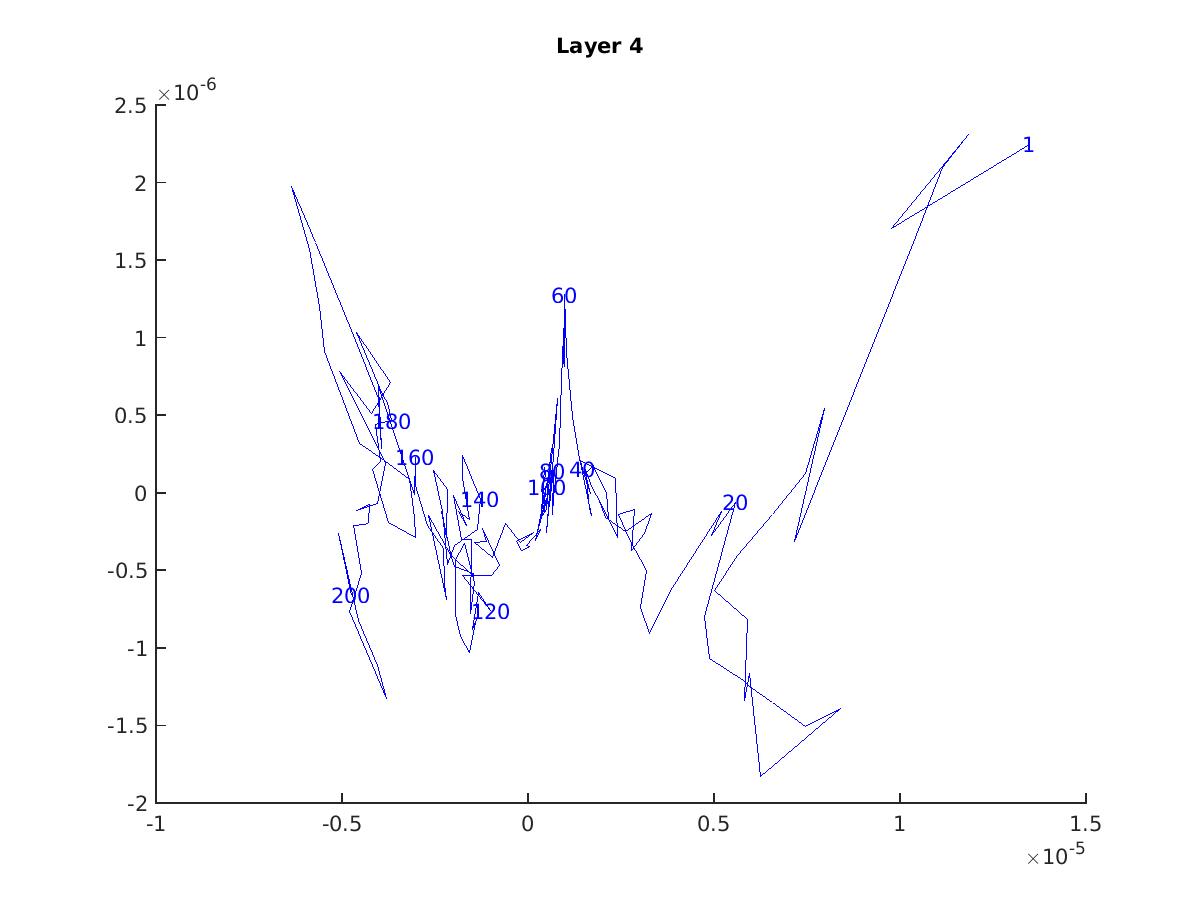}}   & \subfloat[]{\includegraphics[width=0.31\textwidth]{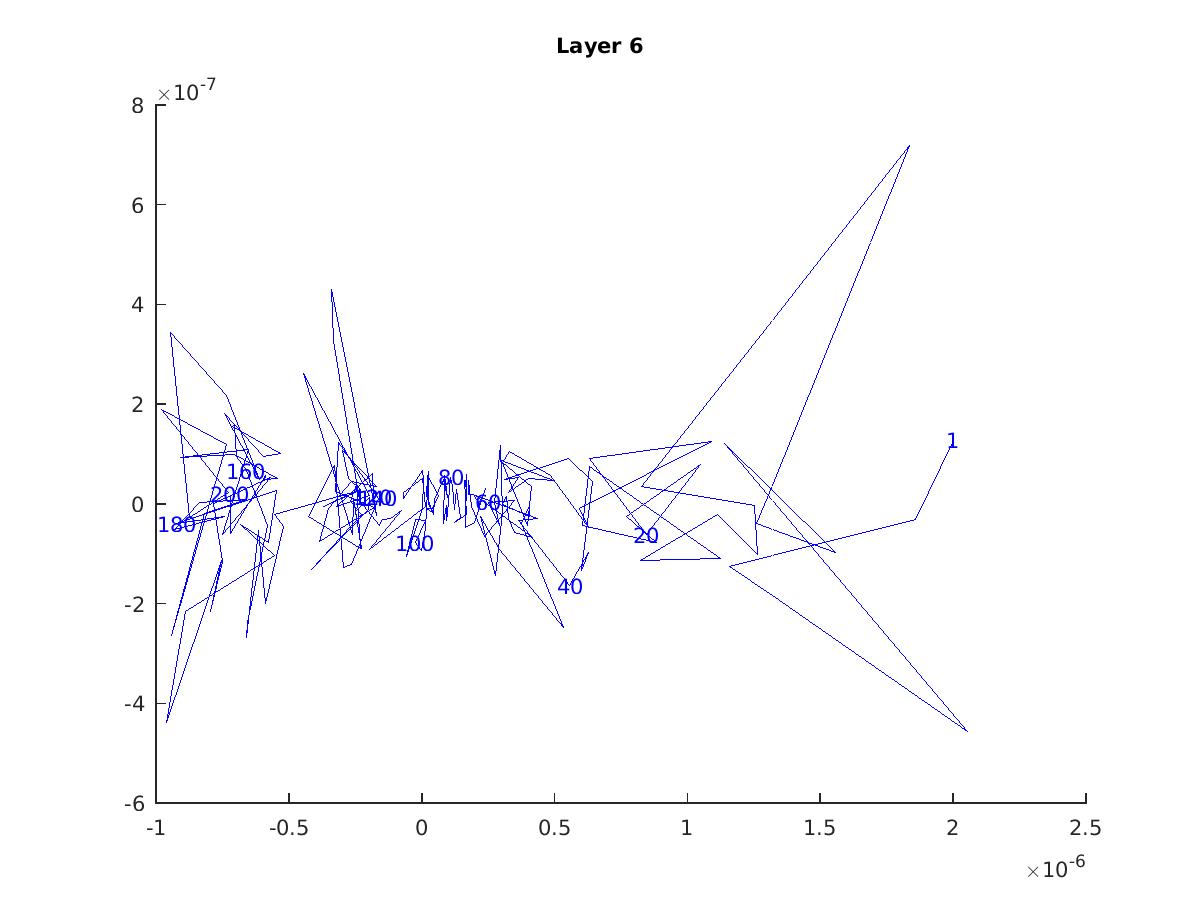}} \\  
\end{tabular}
\caption{The Multidimensional Scaling (MDS) of the layer 2/4/6 weights of model  $M_{final}$ throughout the retraining process.  To provide more resolution, only the weights of one run/trajectory (run 1 in Figure \ref{appendix:fig:perturb_err_loss}) are fed into MDS. The number in the figure indicates the training epoch number after the initial perturbation.} 
\end{figure}

\begin{figure*}[!h]
\centering
\includegraphics[width=0.7\textwidth]{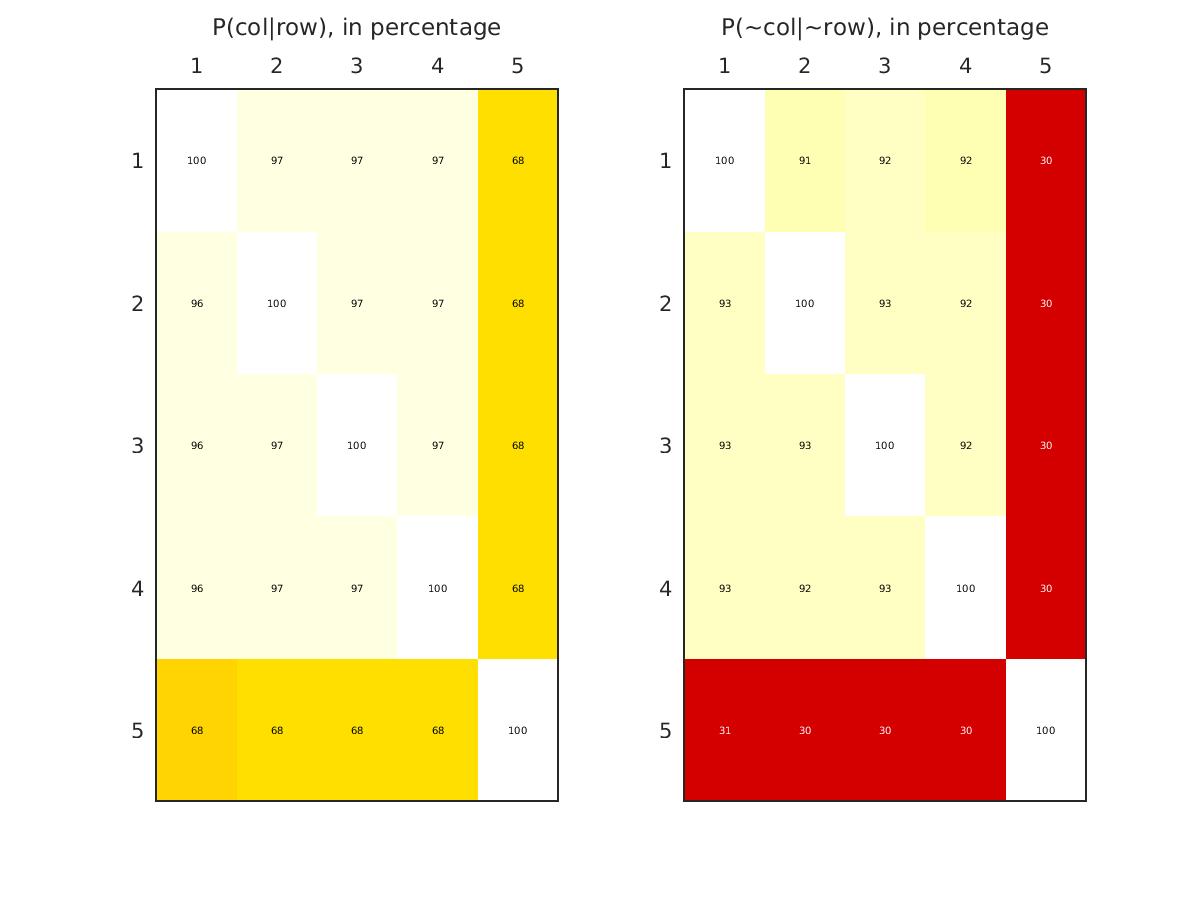}
\caption{The similarity between the final models of the 4 perturbations (row/column 1-4) of $M_{final}$ and a random reference model (the 5-th row/column). The left figure shows the probability of the column model being correct given the row model is correct.  The right figure shows the probability of the column model being incorrect given the row model is incorrect. These two figures show that the perturbed models really become different (although a little similar) models after continued training.}    
\label{appendix:fig:similarity_from_epoch_sgd60_plus_gd400} 
\end{figure*}

\begin{figure*}[!h]
\centering
\includegraphics[width=0.7\textwidth]{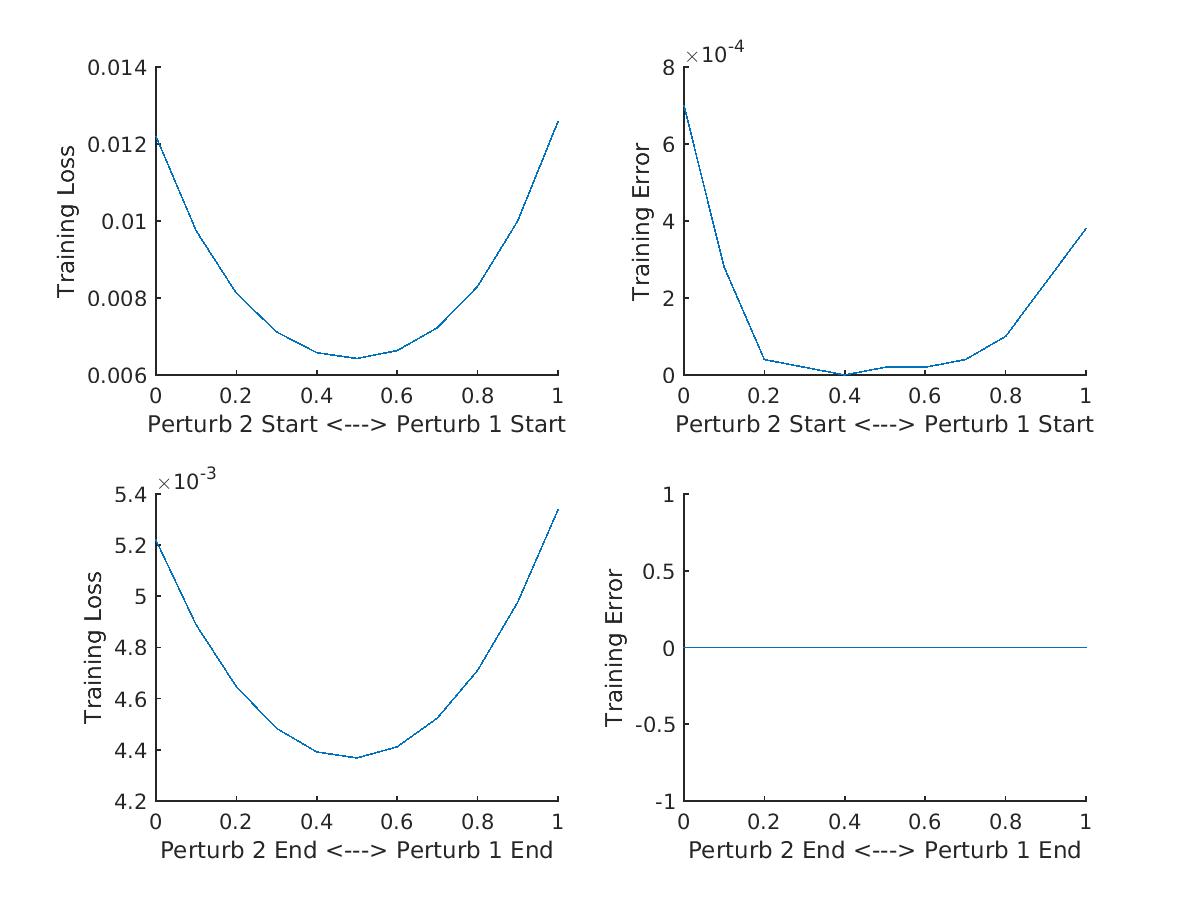}
\caption{Interpolating perturbation 1 and 2 of model  $M_{final}$. The top and bottom rows interpolate the perturbed models before and after retraining, respectively. The x axes are interpolating ratios --- at x=0 the model becomes perturbed model 1 and at x=1 the model becomes perturbed model 2. It is a little surprising to find that the interpolated models between perturbation 1 and 2 can even have slightly lower errors. Note that this is no longer the case if the perturbations are very large.}
\label{appendix:fig:interp_from_epoch_sgd60_plus_gd400} 
\end{figure*}

\newpage

\section{The Landscape of the Empirical Risk: Towards an Intuitive Baseline Model}
\label{appendix:sec:intuitive}

In this section, we propose a simple baseline model for the landscape of empirical risk that is consistent with all of our theoretical and experimental findings.
In the case of overparametrized DCNNs, here is a recapitulation of our main observations:
\begin{itemize}
\item Theoretically, we show that there are a large number of global minimizers with zero (or small) empirical error. The same minimizers are degenerate.
\item Regardless of Stochastic Gradient Descent (SGD) or Batch Gradient Descent (BGD), a small perturbation of the model almost always leads to a slightly different convergence path.  The earlier the perturbation is in the training process the more different the final model would be.
\item Interpolating two ``nearby'' convergence paths lead to another convergence path with similar errors every epoch.  Interpolating two ``distant'' models lead to raised errors.  
\item We do not observe local minima, even when training with BGD.
\end{itemize}

\begin{figure*}[h]
\centering
\includegraphics[width=0.7\textwidth]{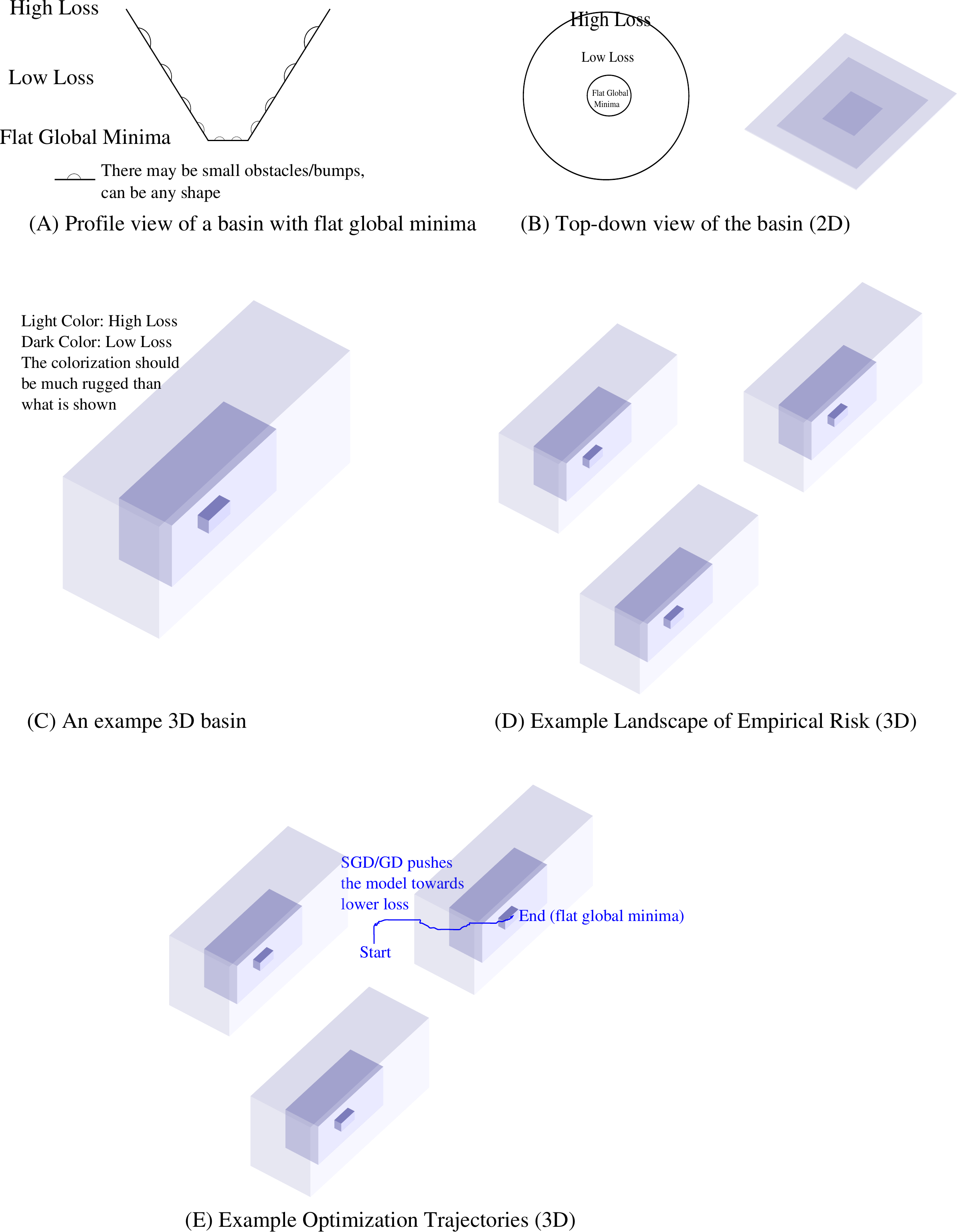}   
\caption{Illustrations of 3D or high-dimensional basins
  (hyper-basins). As shown in Figure \ref{appendix:fig:landscape_model}, in the
  case of overparametrized DCNNs, we believe that the landscape of
  empirical risk is simply a collection of basins that each has a flat
  global minima. We show some illustrations of 3D basins in (C), (D),
  (E). We use color to denote the loss value. One definition of a
  hyper-basin would be that as loss decreases, the hypervolume of the
  parameter space decreases. As we can see in (A), (B) and (C), with
  the same amount of scaling in each dimension, the volume shrinks
  much faster as the number of dimension increases --- with a linear
  decrease in each dimension, the hypervolume decreases as a
  exponential function of the number of dimensions. With the number of
  dimensions being the number of parameters, the volume shrinks
  incredibly fast. This leads to a phenomenon that we all observe
  experimentally: whenever one perturb a model by adding some
  significant noise, the loss almost always never go down. The larger
  the perturbation is, the more the error increases. The reasons are
  simple if the local landscape is a hyper-basin: the volume of a
  lower loss area is so small that by randomly perturbing the point,
  there is almost no chance getting there. The larger the perturbation
  is, the more likely it will get to a much higher loss area. }  
\label{appendix:fig:landscape_model_3d}
\end{figure*}

There is a simple model that is consistent with above observations. As
a first-order characterization, we believe that the landscape of
empirical risk is simply \textbf{a collection of (hyper) basins that each has
  a flat global minima}. Illustrations are provided in Figure
\ref{appendix:fig:landscape_model} and Figure \ref{appendix:fig:landscape_model_3d} (a
concrete 3D case).

As shown in Figure \ref{appendix:fig:landscape_model} and Figure
\ref{appendix:fig:landscape_model_3d}, the building block of the landscape is a
basin.\textbf{How does a basin look like in high dimension? Is there any evidence for this model?} One definition of a hyper-basin
would be that as loss decreases, the hypervolume of the parameter
space decreases: 1D (a slice of 2D), 2D and 3D examples are shown in
Figure \ref{appendix:fig:landscape_model_3d} (A), (B), (C), respectively. As we
can see, with the same amount of scaling in each dimension, the volume
shrinks much faster as the number of dimension increases --- with a
linear decrease in each dimension, the hypervolume decreases as a
exponential function of the number of dimensions. With the number of
dimensions being the number of parameters, the volume shrinks
incredibly fast. This leads to a phenomenon that we all observe
experimentally: whenever one perturb a model by adding some
significant noise, the loss almost always never go down. The larger
the perturbation is, the more the error increases. The reasons are
simple if the local landscape is a hyper-basin: the volume of a lower
loss area is so small that by randomly perturbing the point, there is
almost no chance getting there. The larger the perturbation is, the
more likely it will get to a much higher loss area.

\begin{figure*}[h]
\centering
\includegraphics[width=0.7\textwidth]{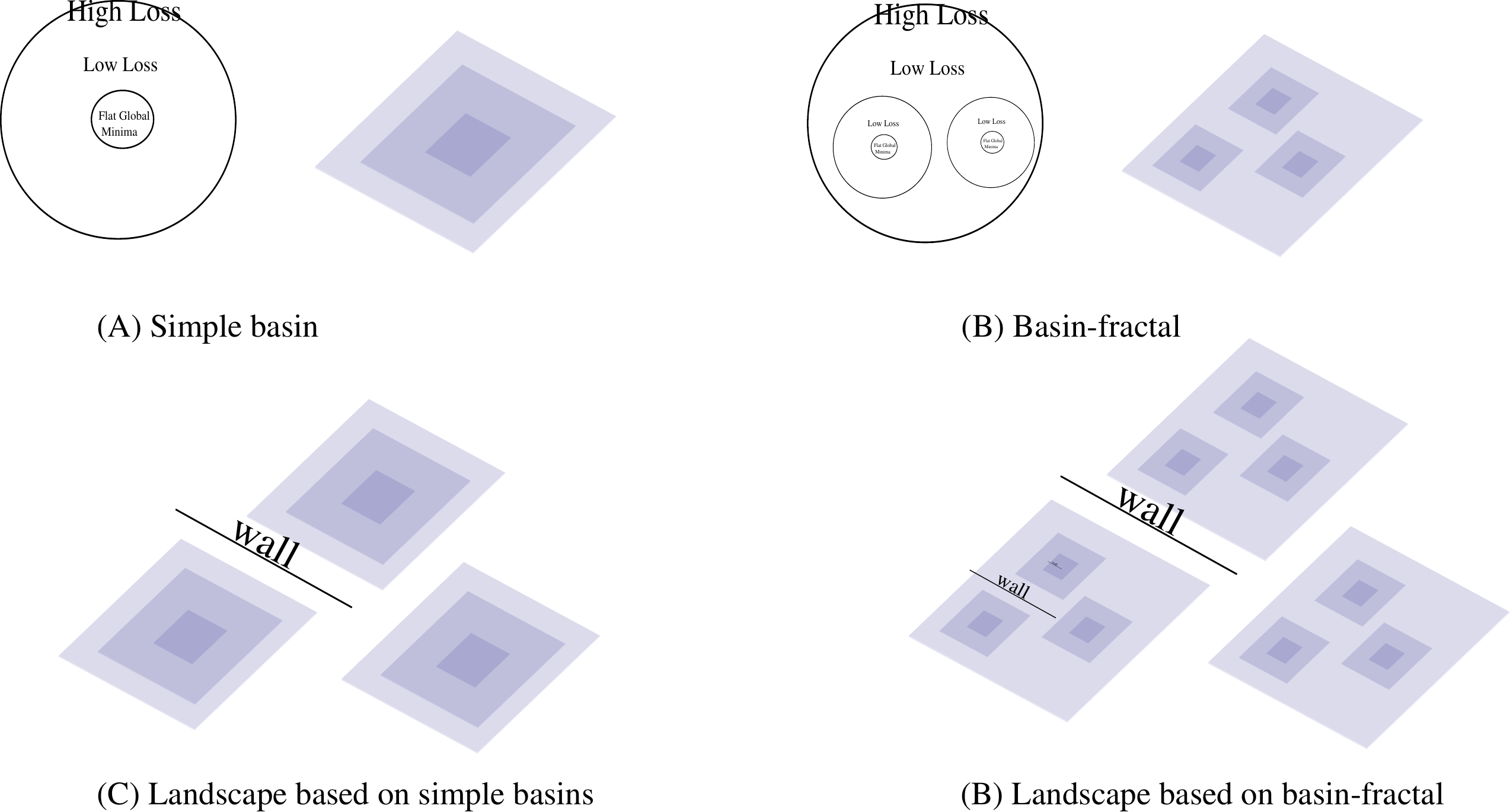}    
\caption{There are, nevertheless, other plausible variants of this model that
can explain our experimental and theoretical results. Here, we show one alternative model we call
``basin-fractal''. The key difference
between simple basins and ``basin-fractal'' is that in
``basin-fractal'', one should be able to find ``walls'' (raised
errors) between two models within the same basin. Since it is a
fractal, these ``walls'' should be present at all levels of errors.
For the moment, we only discovered ``walls'' between two models the
trajectories lead to which are very different (obtained either by
splitting very early in training, as shown in Figure
\ref{appendix:fig:branch_layer_2_sep_perturb_0.25} (a) and Figure
\ref{appendix:fig:branch_layer_3_sep_perturb_1} (a) or by a very significant
perturbation, as shown in Figure
\ref{appendix:fig:branch_layer_3_sep_perturb_1} (b)). We have not found other
significant ``walls'' in all other perturbation and interpolation
experiments. So a first order model of the landscape would be just a
collection of simple basins, as shown in (C). Nevertheless, we do find
``basin-fractal'' elegant, and perhaps the ``walls'' in the low loss
areas are just too flat to be noticed. }
\label{appendix:fig:basin_fractal}
\end{figure*}

There are, nevertheless, other plausible variants of this model that
can explain our experimental findings. In Figure
\ref{appendix:fig:basin_fractal}, we show one alternative model we call
``basin-fractal''. This model is more elegant while being also
consistent with most of the above observations. The key difference
between simple basins and ``basin-fractal'' is that in
``basin-fractal'', one should be able to find ``walls'' (raised
errors) between two models within the same basin. Since it is a
fractal, these ``walls'' should be present at all levels of errors.
For the moment, we only discovered ``walls'' between two models the
trajectories lead to which are very different (obtained either by
splitting very early in training, as shown in Figure
\ref{appendix:fig:branch_layer_2_sep_perturb_0.25} (a) and Figure
\ref{appendix:fig:branch_layer_3_sep_perturb_1} (a) or by a very significant
perturbation, as shown in Figure
\ref{appendix:fig:branch_layer_3_sep_perturb_1} (b)). We have not found other
significant ``walls'' in all other perturbation and interpolation
experiments. So a first order model of the landscape would be just a
collection of simple basins. Nevertheless, we do find
``basin-fractal'' elegant, and perhaps the ``walls'' in the low loss
areas are just too flat to be noticed.

Another surprising finding about the basins is that, they seem to be
so ``smooth'' such that there is no local minima. Even when training
with batch gradient descent, we do not encounter any local minima.
When trained long enough with small enough learning rates, one always
gets to 0 classification error and negligible cross entropy loss.

\section{Overparametrizing DCNN does not harm generalization}  
\begin{figure*}[!h]
\centering
\includegraphics[width=\textwidth]{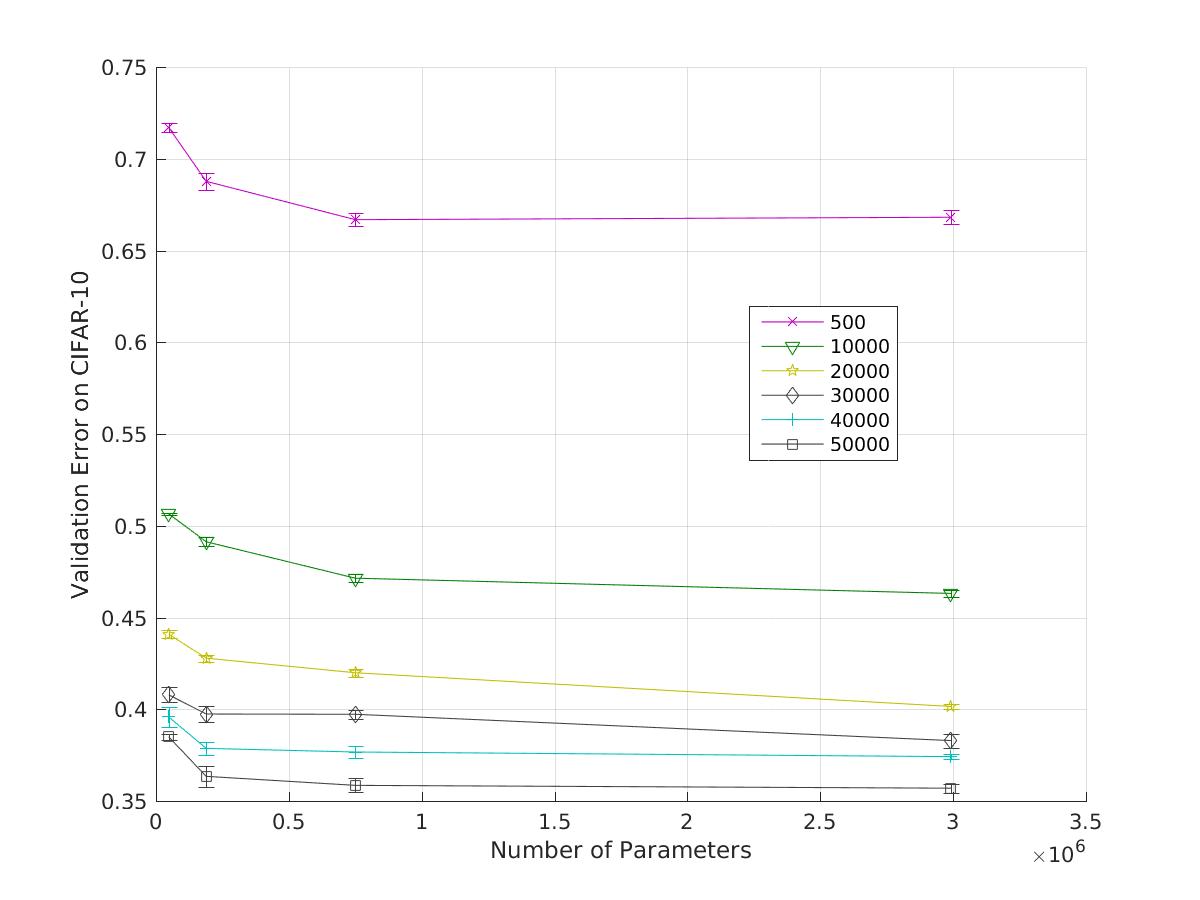}
\caption{We show that with different amount of training data, overparametrizing a DCNN never lower the validation error.}  
\label{appendix:fig:generalization}
\end{figure*}

See Figure \ref{appendix:fig:generalization}.
    
\section{Study the flat global minima by perturbations on CIFAR-10 (with smaller perturbations)}  
Zero-error model $M_{final}$: We first train a 6-layer (with the 1st layer 
being the input) convolutional network on CIFAR-10. The model reaches
0 training classification error after 60 epochs of stochastic gradient
descent (batch size = 100) and 372 epochs of gradient descent (batch
size = training set size). We call this model $M_{final}$. Next we perturb 
the weights of this zero-error model and continue training it. This
procedure was done multiple times to see whether the weights converge
to the same point.  Note that no data augmentation is performed, so
that the training set is fixed (size = 50,000).

The procedures are essentially the same as what described in main text
Section \ref{appendix:vis:sec3}. The main difference is that the perturbations
are smaller. The classification errors are even 0 after the perturbation
and throughout the entire following training process. 

\begin{figure*}[!h]
\centering
\includegraphics[width=0.8\textwidth]{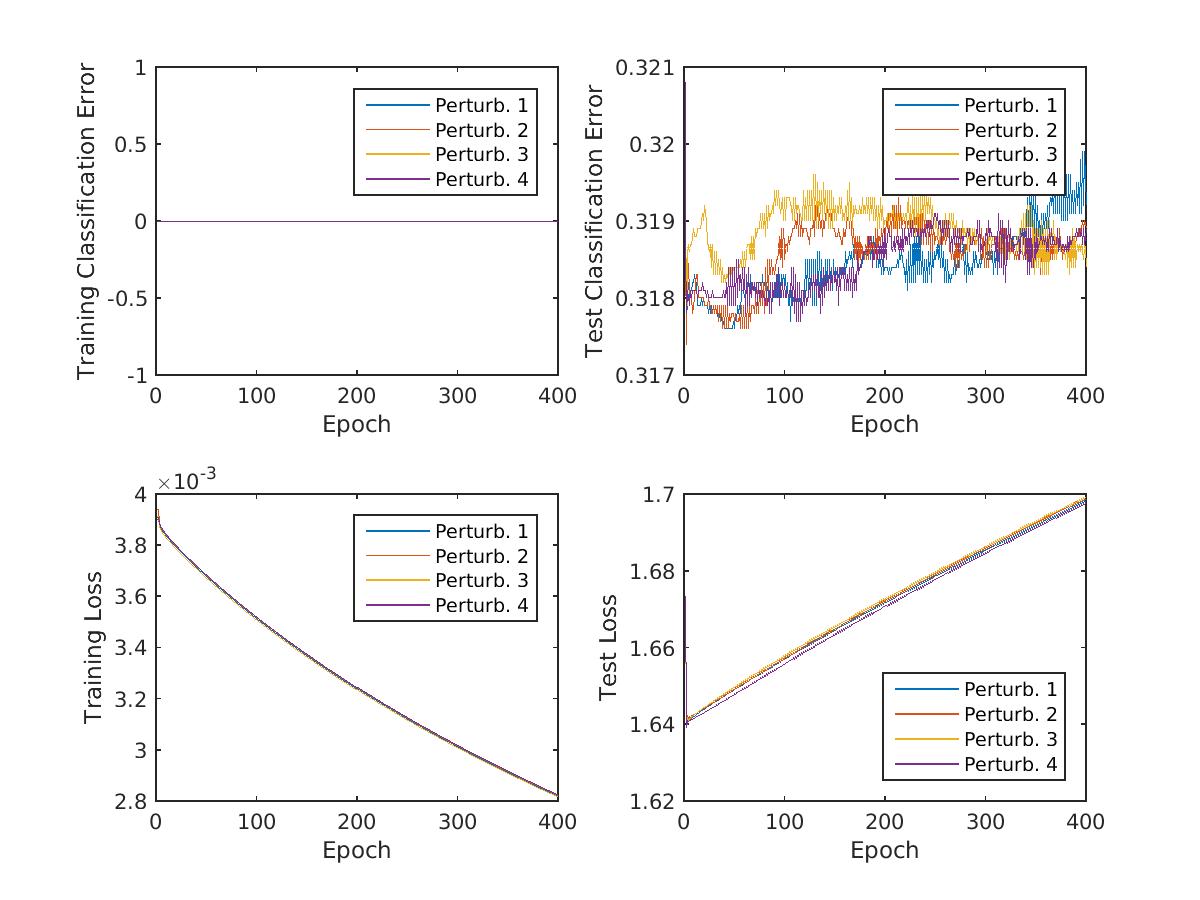}
\caption{We perturb the weights of model $M_{final}$ by adding a gaussian noise with standard deviation = 0.01 * m, where m is the average magnitude of the weights. After perturbation, we continue training the model with 400 epochs of gradient descent (i.e., batch size = training set size). The same procedure was performed 4 times, resulting in 4 curves shown in the figures. The training and test classification errors and losses are shown. The training classification errors are 0 after the perturbation and throughout the  entire following training process. } 
\label{appendix:fig:perturb_err_loss}
\end{figure*}

\begin{figure*}[!h]
\centering
\includegraphics[width=0.7\textwidth]{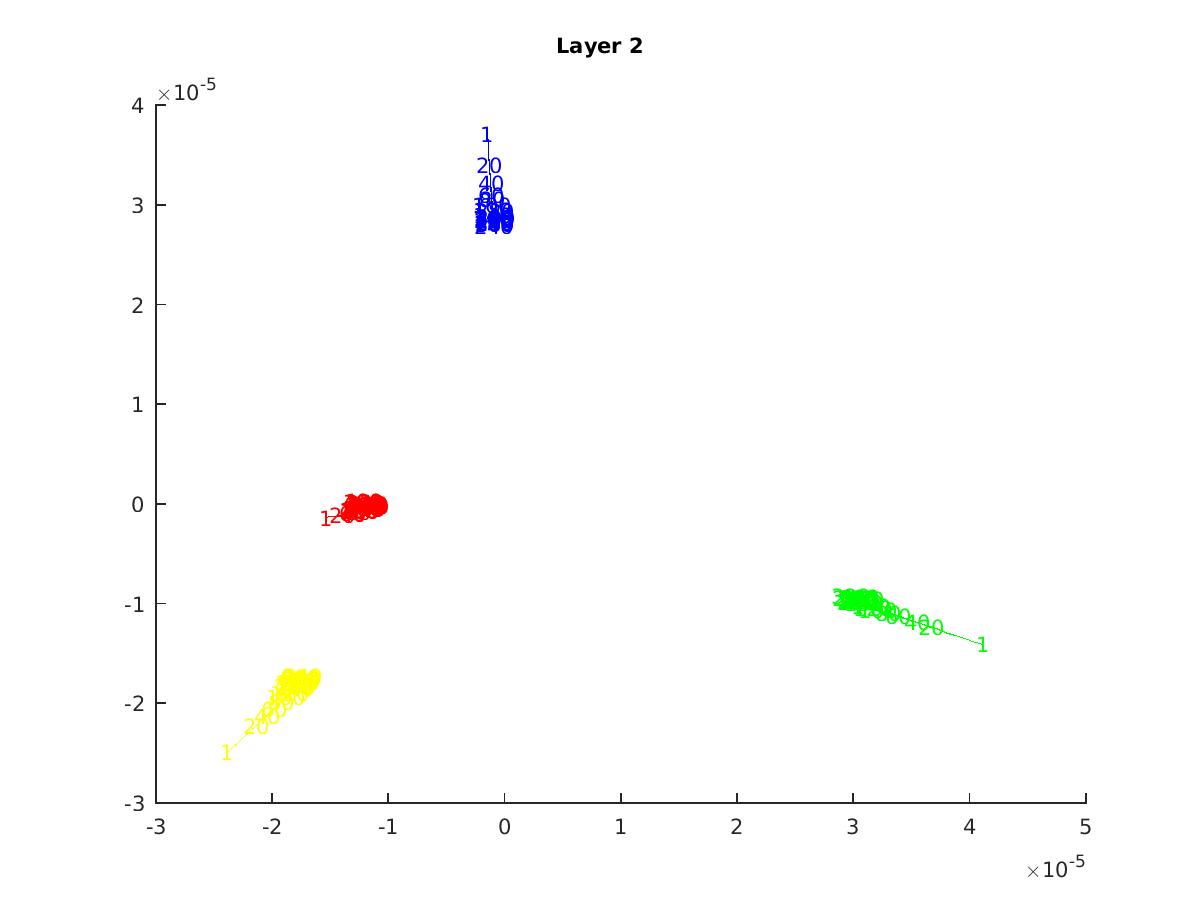}
\caption{Multidimensional scaling of the layer 2 weights throughout the training process. There are 4 runs indicated by 4 colors/trajectories, corresponding exactly to Figure \ref{appendix:fig:perturb_err_loss}. Each run corresponds to one perturbation of the zero-error model $M_{final}$ and subsequent 400 epochs' training. The number in the figure indicates the training epoch number after the initial perturbation. We see that even a small perturbation makes the weights significantly different and the 4 runs do not converge to the same weights. All points shown in this figure has 0 classification error on the entire training set.} 
\label{appendix:fig:perturb_layer_2A}
\end{figure*}

\begin{figure*}[!h]
\centering
\includegraphics[width=0.7\textwidth]{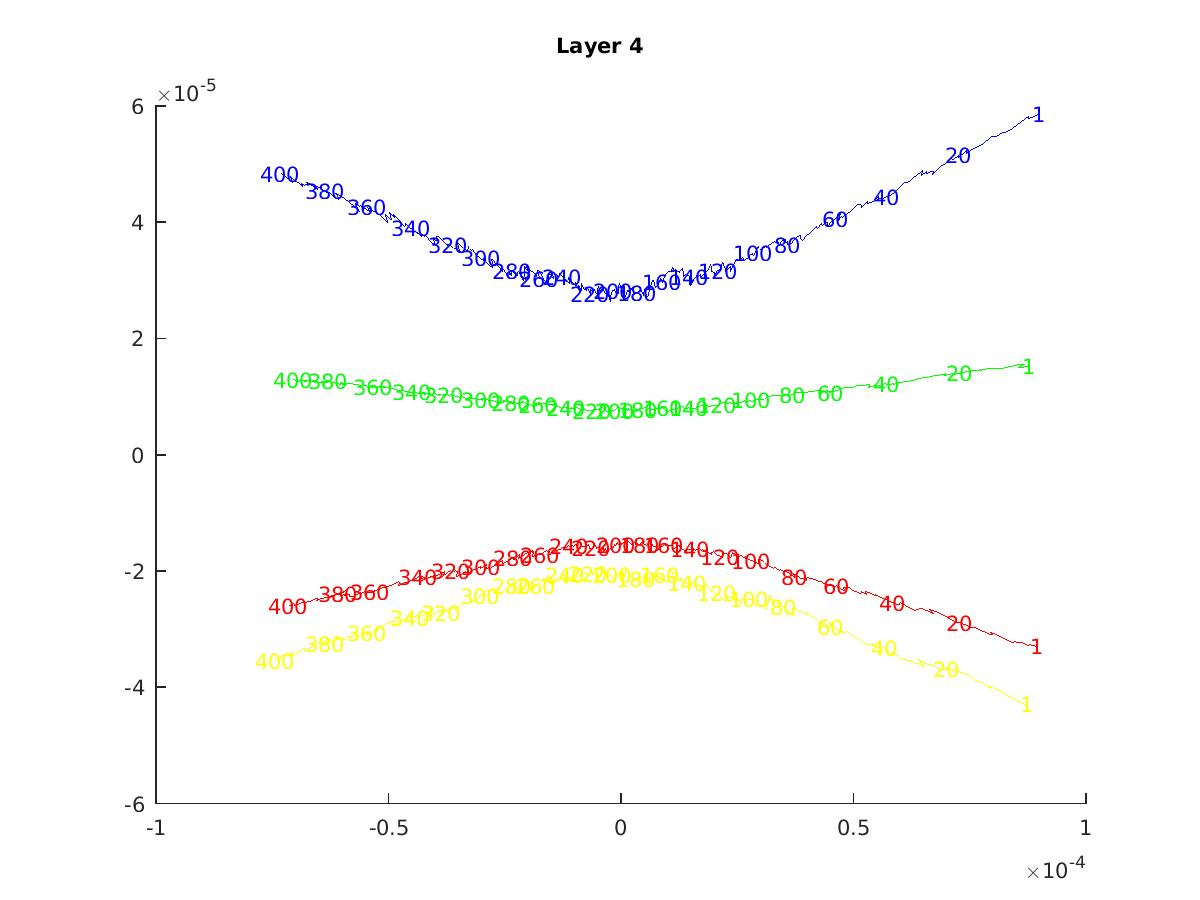}
\caption{Multidimensional scaling of the layer 4 weights throughout the training process. There are 4 runs indicated by 4 colors/trajectories, corresponding exactly to Figure \ref{appendix:fig:perturb_err_loss}. Each run corresponds to one perturbation of the zero-error model $M_{final}$ and subsequent 400 epochs' training. The number in the figure indicates the training epoch number after the initial perturbation.  All points shown in this figure has 0 classification error on the entire training set.}
\label{appendix:fig:perturb_layer_4A}
\end{figure*}
 
\begin{figure*}[!h]
\centering
\includegraphics[width=0.7\textwidth]{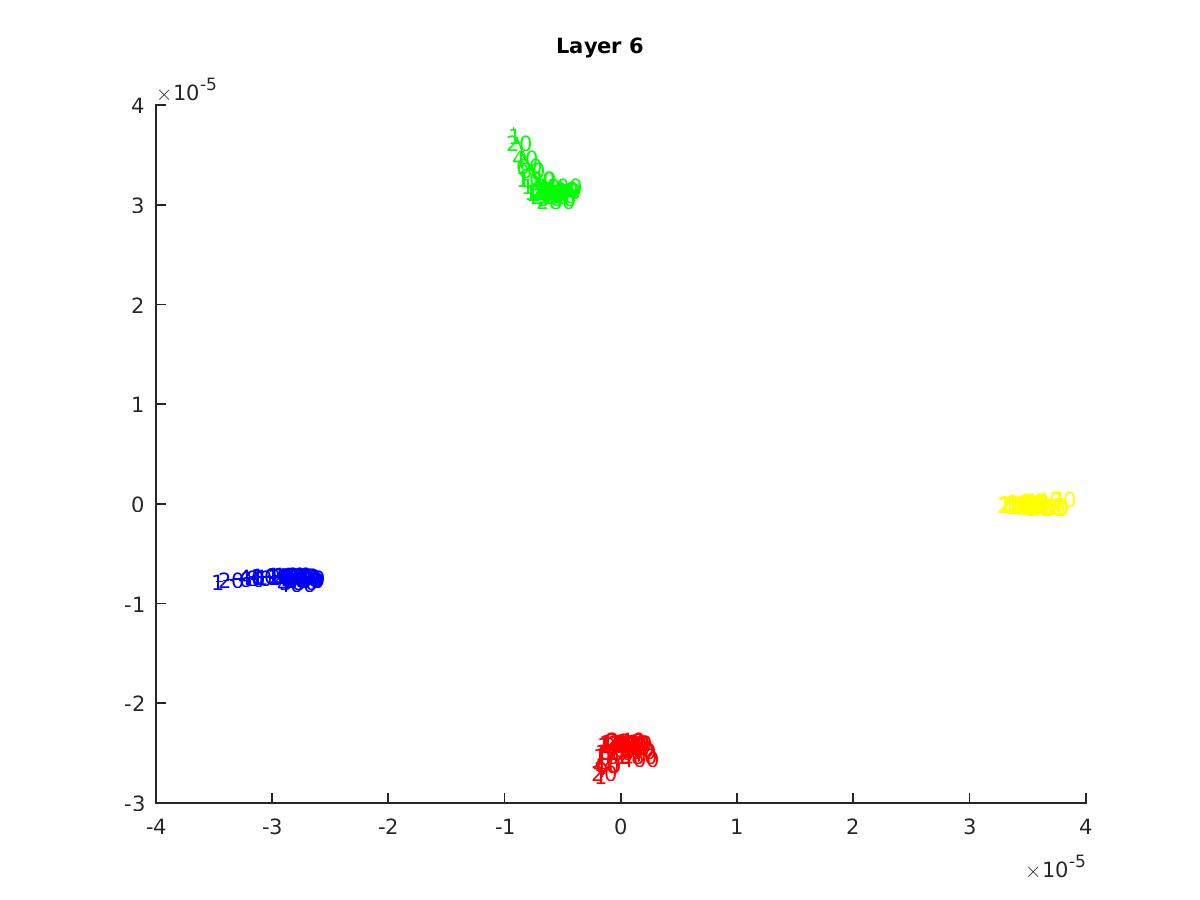}
\caption{Multidimensional scaling of the layer 4 weights throughout the training process. There are 4 runs indicated by 4 colors/trajectories, corresponding exactly to Figure \ref{appendix:fig:perturb_err_loss}. Each run corresponds to one perturbation of the zero-error model $M_{final}$ and subsequent 400 epochs' training. The number in the figure indicates the training epoch number after the initial perturbation. All points shown in this figure has 0 classification error on the entire training set.}
\label{appendix:fig:perturb_layer_6A} 
\end{figure*}

\begin{figure}  
\begin{tabular}{ccc}
  \subfloat[]{\includegraphics[width=0.31\textwidth]{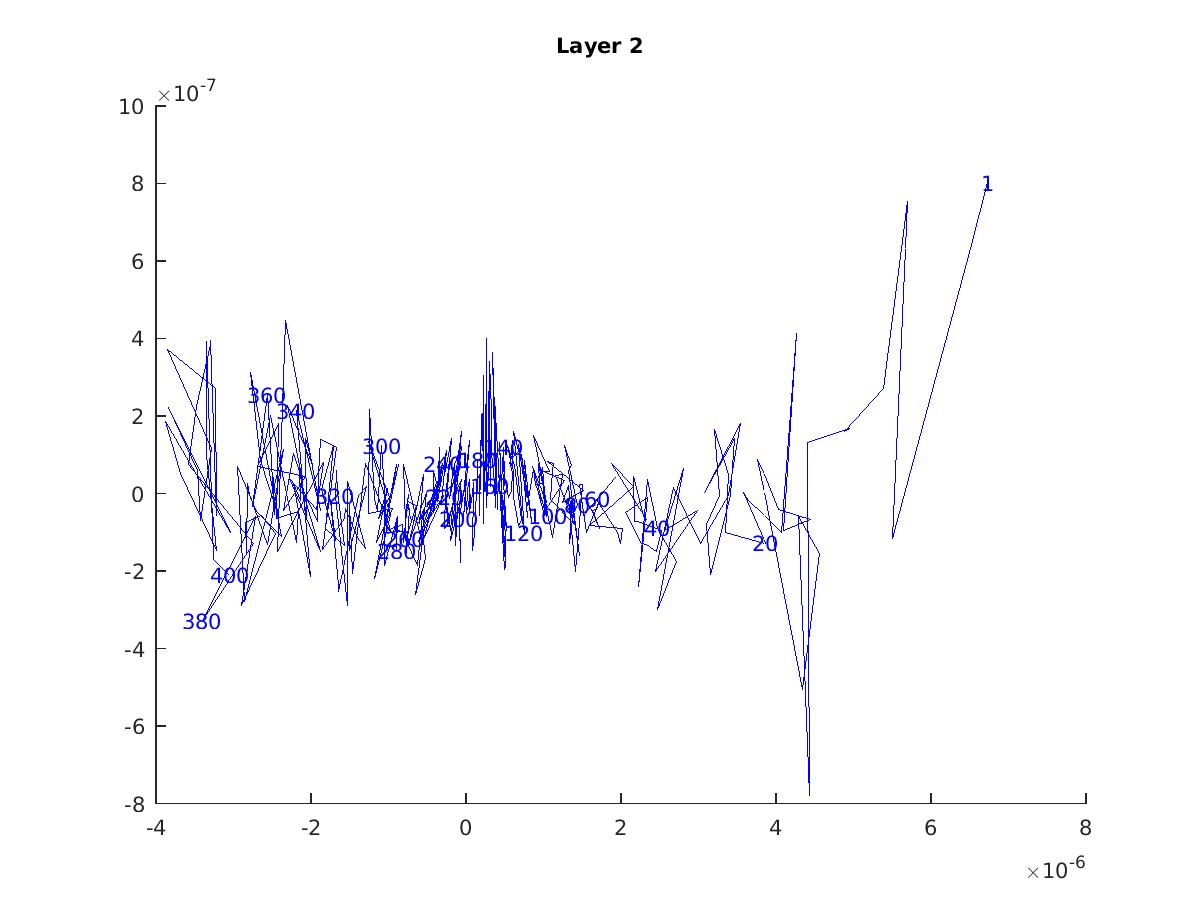}}   & \subfloat[]{\includegraphics[width=0.31\textwidth]{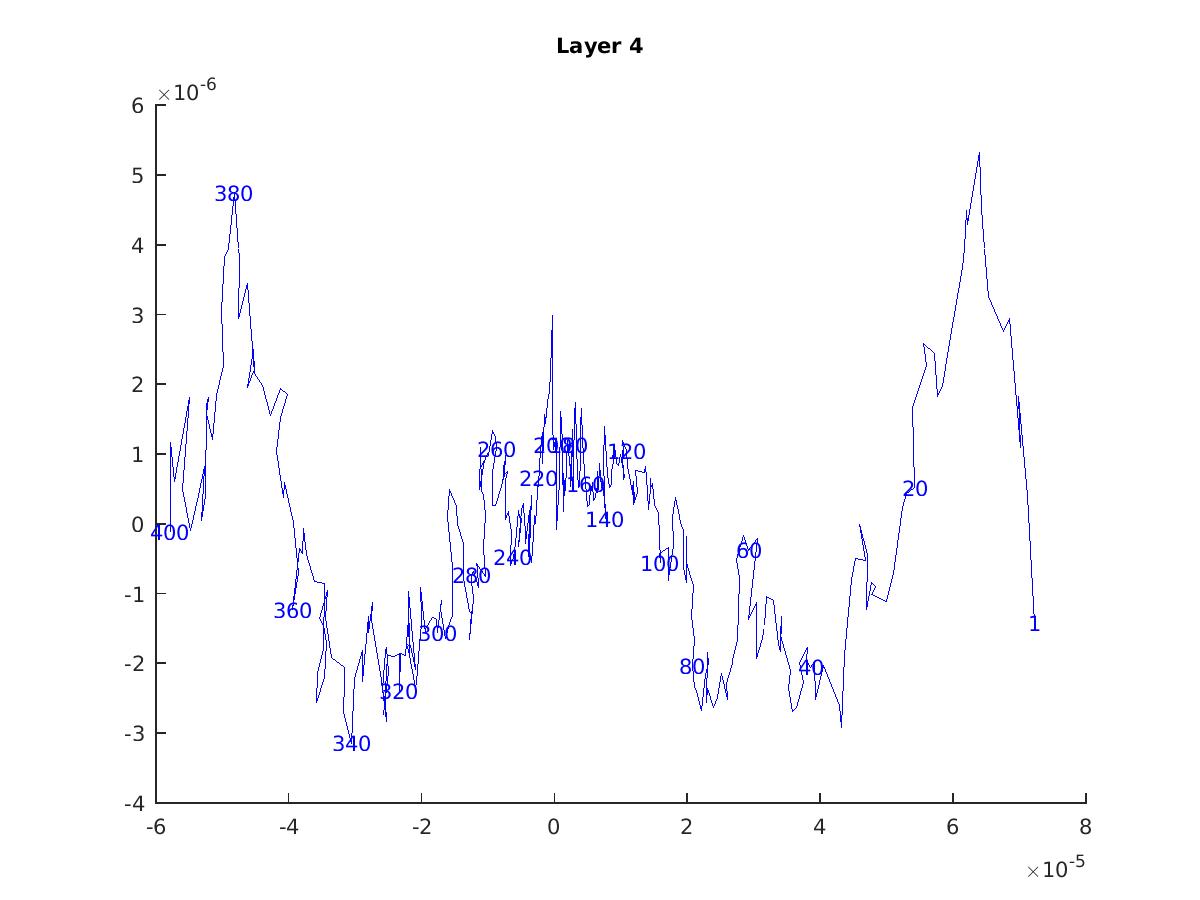}}   & \subfloat[]{\includegraphics[width=0.31\textwidth]{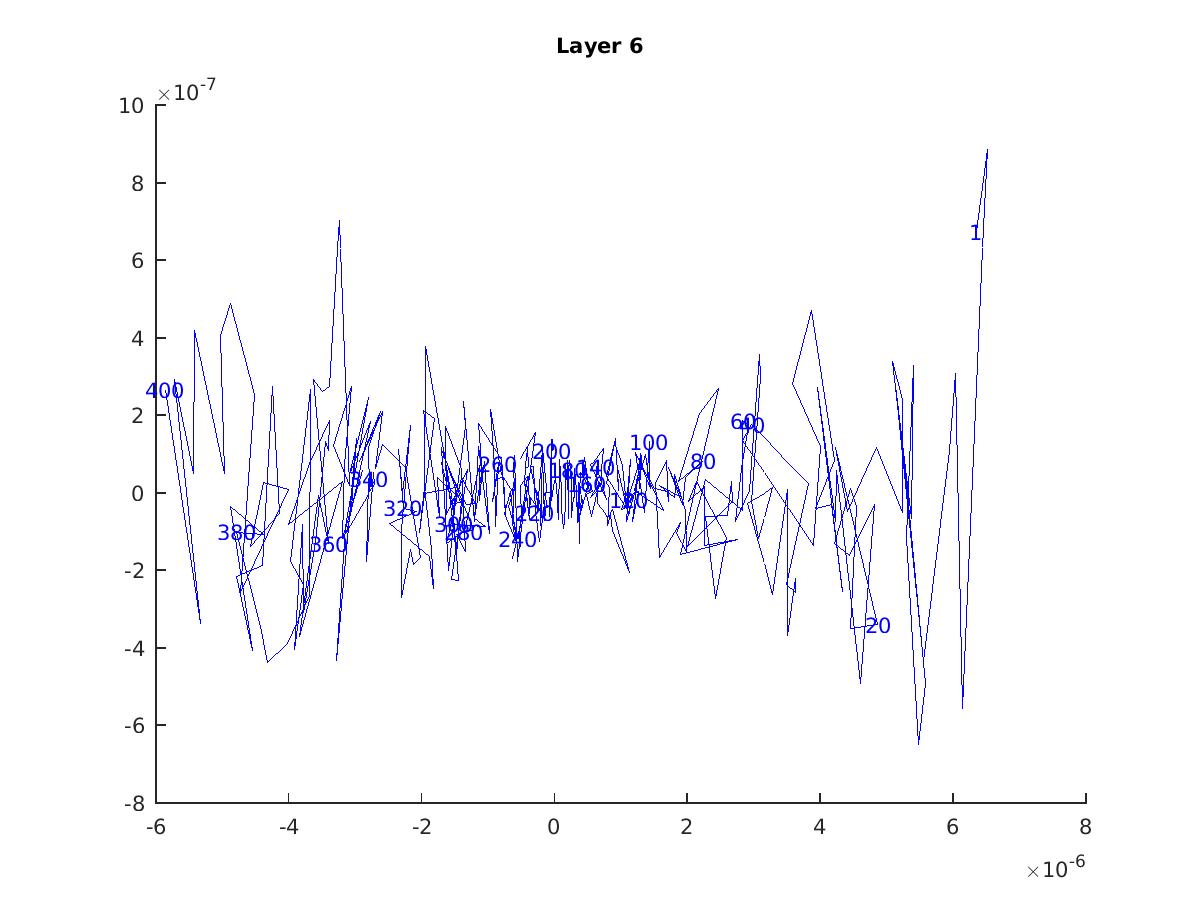}} \\  
\end{tabular}
\caption{The Multidimensional Scaling (MDS) of the layer 2 weights throughout the training process. Only the weights of one run (run 1 in Figure \ref{appendix:fig:perturb_err_loss}) are fed into MDS to provide more resolution. The number in the figure indicates the training epoch number after the initial perturbation. All points shown in this figure has 0 classification error on the entire training set.} 
\end{figure}


\end{document}